\documentclass[10pt,journal,cspaper,compsoc]{IEEEtran}
\pdfoutput=1
\usepackage{graphicx}
\graphicspath{{./}}
\DeclareGraphicsExtensions{.jpg,.pdf} 
\usepackage{dsfont}
\usepackage[cmex10]{amsmath}
\usepackage{multirow}
\usepackage{url}

\begin{document}
%
\title{Single View Depth Estimation from Examples}

\author{Tal~Hassner,
        Ronen~Basri,
\IEEEcompsocitemizethanks{\IEEEcompsocthanksitem T. Hassner is with the Department
of Mathematics and Computer Science, The Open University of Israel, Israel.\protect\\
E-mail: see http://www.openu.ac.il/home/hassner/
\IEEEcompsocthanksitem R. Basri is with The Dept. of Computer Science and Applied Math., Weizmann Institute, Israel.}
\thanks{}}

\markboth{}%
{Hassner and Basri: Single View Depth Estimation from Examples}


\IEEEcompsoctitleabstractindextext{%
\begin{abstract}
We describe a non-parametric, ``example-based'' method for estimating the depth of an object, viewed in a single photo. Our method consults a database of example 3D geometries, searching for those which look similar to the object in the photo. The known depths of the selected database objects act as shape priors which constrain the process of estimating the object's depth. We show how this process can be performed by optimizing a well defined target likelihood function, via a hard-EM procedure. We address the problem of representing the (possibly infinite) variability of viewing conditions with a finite (and often very small) example set, by proposing an on-the-fly example update scheme. We further demonstrate the importance of non-stationarity in avoiding misleading examples when estimating structured shapes. We evaluate our method and present both qualitative as well as quantitative results for challenging object classes. Finally, we show how this same technique may be readily applied to a number of related problems. These include the novel task of estimating the occluded depth of an object's backside and the task of tailoring custom fitting image-maps for input depths.
\end{abstract}}

\maketitle


%

\section{Introduction}
\IEEEPARstart{T}{he} human visual system is remarkably adapt at estimating the shapes of objects from just a single view, despite this being an ill posed problem; many different shapes can appear the same in an image and any one of them is as plausible as the next. To overcome this difficulty, existing computational methods routinely make a-priori assumptions on the lighting properties, the object's surface properties, the structure of the scene, and more. 

Here, we make the following alternative assumption: We assume that the object viewed is roughly similar in shape (but by no means identical) to the shapes of a set of related objects. The obvious example here is of course faces. As we will later show, examples of typical face shapes can be used to estimate even highly unusual faces. We claim that the same is true for other object classes and indeed demonstrate results for images of challenging objects, including hands, full body figures (non-rigid objects), and fish (highly textured objects). 

Specifically, we assume that we have at our disposal a database of relevant example 3D geometries. We can easily obtain the appearances of these objects, viewed under any desired viewing condition, by using standard rendering techniques. To estimate the shape of a novel object from a single image we search through the appearances of these objects, possibly rendering new appearances as we go, looking for ones appearing similar to the input image. Once found, the known depths of these selected objects serve as priors for the object's shape. We perform this task at the patch level, thus obtaining depth estimates very different from those in the database. This process is performed via a Hard-EM procedure, optimizing a well defined target likelihood function representing the likelihood of the estimated depth given the input image and the set of examples.

This approach to depth estimation has a number of advantages: (1) Our method is non-parametric, and as such, requires no a-priori model selection or design. Consequently, (2) it is {\em versatile}. As we will show, the same method is used to estimate the shapes of objects belonging to very different object classes and even to solve additional related tasks. Finally, (3) a data-driven approach requires making no assumptions on the properties of the object in the image nor the viewing conditions. Our chief requirement is the existence of a suitable set of 3D examples. We believe this to be a reasonable requirement given the growing availability of such databases. 


Obviously, in taking an example-based approach to depth estimation, we have no guarantee that the example data sets we use contain objects sufficiently similar to the one in the input image. We therefore follow the example of methods such as~\cite{assa2007,Hoiem:popup,Zhang:svm} in seeking to produce {\em plausible} depth estimates and not necessarily the true depths. Here, however, the concept of a plausible depth is formally defined by our target function. Moreover, we present quantitative results suggesting that our method is indeed capable of producing accurate estimates even for challenging objects, given an adequate example set.

To summarize, this report reviews the following topics.
\begin{itemize}
\item {\bf Example-based approach to depth estimation.} We describe an approach to single-view depth estimation and present both qualitative and quantitative results on a number of challenging object classes. We have tested out method on large sets of objects, including real and synthetic images of objects with arbitrary texture, pose, and genus, viewed under unconstrained viewing conditions.  
\item{\bf On-the-fly example update scheme.} We augment existing example-based methods by arguing that examples {\em need not be selected a-priori}. To handle the possible infinite viewing conditions and postures of the objects being reconstructed we produce better suited examples while removing less adequate ones {\em on-the-fly}, as part of the reconstruction process. 
\item{\bf Non-stationarity for structured shape reconstruction.} We emphasize the importance of non-stationarity in avoiding depth ambiguities and making better example selections. 
\item{\bf Additional applications.} We show how the same method used for depth estimation may also be used for the additional tasks of estimating the depths of the {\em occluded} backside of objects viewed in an image as well as estimating the colors of objects from their shape.
\end{itemize}
The rest of this report is organized as follows. In the next section we review related work. Our depth estimation framework is described in Sec.~\ref{sec:estimate}. Our example update scheme is presented in Sec.~\ref{sec:variability}  followed by a discussion on non-stationarity in Sec.~\ref{sec:structural}. We propose additional applications, based on our method in Sec.~\ref{sec:app}. Implementation and results are presented in Sec.~\ref{sec:results}. Finally, we conclude in Sec.~\ref{sec:future}.

\section{Related work}\label{sec:related}
{\bf Depth estimation.} There is an immense volume of literature on the problem of estimating the shapes of objects or scenes from a single image. Indeed, this problem is considered to be one of the classical challenges of Computer Vision. Methods for single image reconstruction very often rely on different cues such as shading, silhouette shapes, texture, and vanishing points (e.g.,~\cite{assa2007,Cipolla:cusps,Criminisi00a,Delage:Manhatten3D,Horn75,Witkin:shapetexture}). These methods restrict the allowable reconstructions by placing constraints on the properties of reconstructed objects (e.g., reflectance properties, viewing conditions, and symmetry).

There has been a growing interest in producing depth estimates for large scale outdoor scenes from single images. One approach~\cite{Hoiem:popup,Hoiem:geocontext} reconstructs outdoor scenes assuming they can be labeled as
``ground,'' ``sky,'' and ``vertical'' billboards. Other approaches include the Diorama construction method of~\cite{assa2007} and the Make3D system of~\cite{Saxena:08:make3d}. Although both visually pleasing and quantitatively accurate estimates have been demonstrated, it is unclear how to extend these methods to classes other than outdoor scenes.

Recently, there is a growing number of methods explicitly using examples to guide the reconstruction process. One notable approach makes the assumption that all 3D objects in the class being modeled lie in a linear space spanned using a few basis objects (e.g.,~\cite{Atick:Statistical,Blanz:morphable,Dovgard:Symmetric,Romdhani:Efficient}). This approach is applicable to faces, but it is less clear how to extend it to more variable classes because it requires dense correspondences between surface points across examples. Another approach~\cite{Kemelmacher:moldingfaces} uses a single example to produce accurate, shape-from-shading estimates of face shapes. This approach too is tailored for the particular problem of estimating face shapes. By contrast, our chief assumption is that the object viewed in the query image has similar looking counterparts in our example set, and so can be applied to produce depth estimates for a range of different object classes.\\




\noindent{\bf Synthesis ``by-example''.} A fully data-driven method was first proposed in~\cite{Hassner06reconstruct}, inspired by methods for constructing 3D models by combining parts~\cite{Hassner03Mincut}. It operates by assuming a collection of example, reference images, along with known 3D shapes (depths). For a given query image, it seeks to match the appearance of the query to the appearance of these references, and produces a depth estimate by combining the known reference depth values, associated with the matching appearances. This report elaborates on the original method described in~\cite{Hassner06reconstruct} and provides additional information and results compared to that paper. \\

\begin{figure*}[!t]
\begin{center}
\begin{tabular}{ccc}
\includegraphics[height=3.4cm, clip, trim=0.8cm 0cm 0cm 0.5cm]{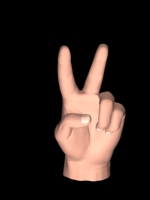} &%
\includegraphics[height=5.5cm, clip, trim=0cm 1.7cm 1.3cm 2cm]{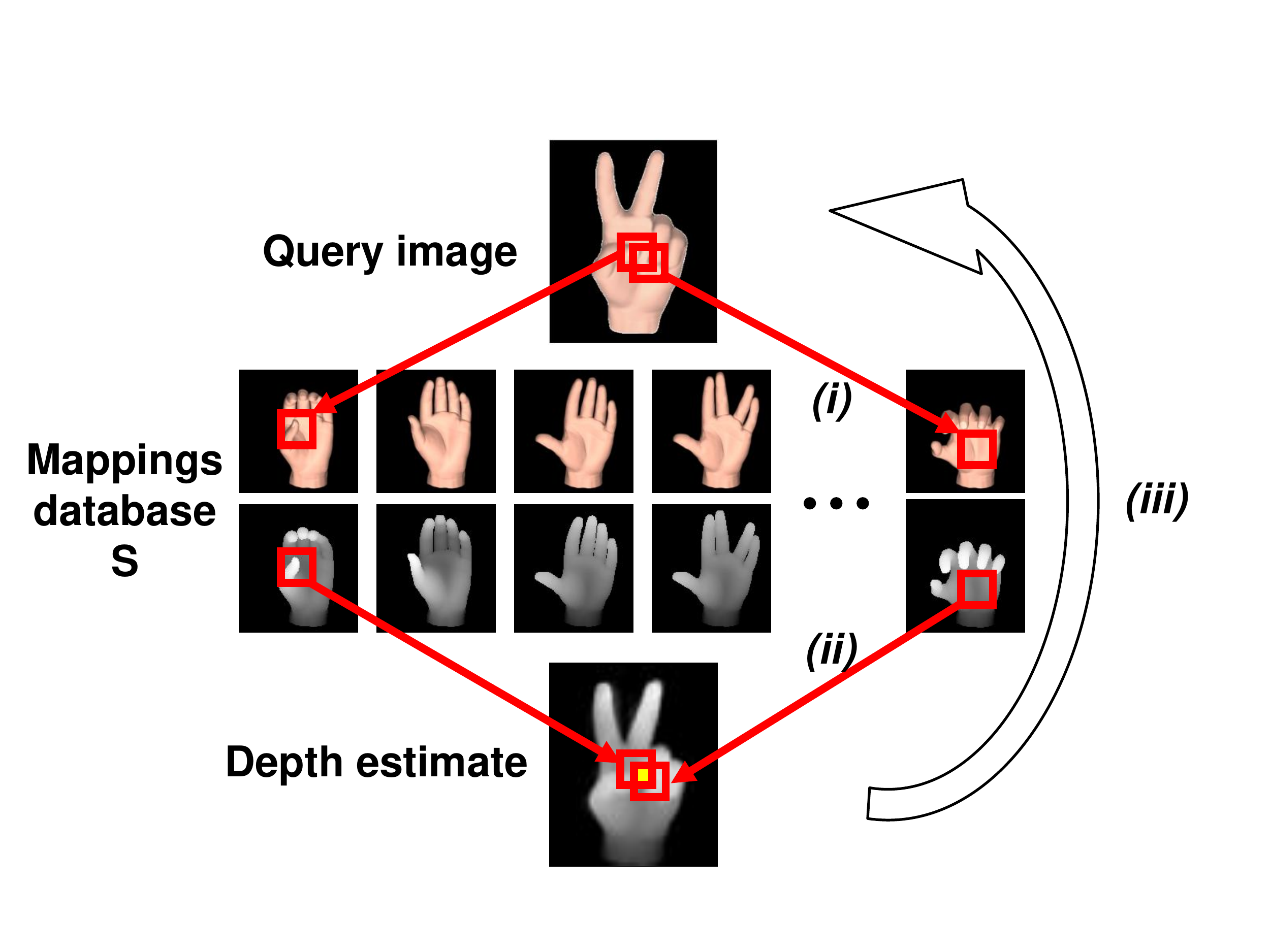} &%
\hspace{-0.5cm}\includegraphics[height=4cm, clip, trim=1cm 0cm 0cm 0cm]{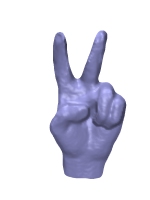}
\\
(a) & (b) & (c)
\end{tabular}
\end{center}

\caption{{\bf Visualization of our process.} (a) The input image. (b)
Step~(i) finds for every query patch a similar patch in the
database. Each patch provides depth estimates for the pixels it
covers. Thus, overlapping patches provide several depth estimates
for each pixel. These values are combined at each pixel to produce
a new estimate for that pixel's depth in Step~(ii). This process
is then repeated until convergence (Step~(iii)) by returning to
Step~(i), now searching for patches matching in both intensity and
depth, using the current depth estimate for the comparison. (c)
Our final depth estimate.}\label{fig:optimize}
\end{figure*} 

\noindent{\bf Shape decoration.} Sec.~\ref{sec:app} demonstrate how our framework can be applied to solve 
additional problems beyond shape reconstruction. In particular, we demonstrate the use of our method for automatically colorizing depth-maps, as a quick means for decorating 3D shapes (Sec~\ref{sec:app_colorization}). Existing automatic methods for decorating 3D models have mostly focused on covering the surface
of 3D models with texture examples (e.g.,~\cite{Turk:3Dtexturesynth,wei01texture,ying:texture}). We
note in particular a work concurrent to our own~\cite{Han:SurfaceTextureSynthesis}, which uses an optimization procedure similar to the one used here. Their goal, however, is to cover 3D surfaces with 2D texture examples. Finally, recent methods have attempted to allow the modeler semi-automatic means of producing non-textural image-maps (e.g.,~\cite{Zhou05texturemontage}). These methods rely on the user forming explicit correspondences between parts of the 3D surface, and different texture examples, which are then merged together to produce the complete image-map for an input 3D model. Our work, on the other hand, is fully automatic.

Finally, there have been a number of publications presenting methods for 3D model (e.g., triangulated mesh) correspondences and cross parameterization (e.g.,~\cite{Kraevoy:CrossParameterization,Praun:ConsistentParameterization}). These methods establish correspondences across two or more 3D surfaces. Once these correspondences are computed, surface properties such as texture, can be transferred from one corresponding 3D object to another, thus providing a novel model with a custom image-map. These methods, however, often require a human modeler to input a seed set of correspondences across the models, or else assume the models are similar in general form. In our colorization method, no prior
correspondences are required, the process is fully automatic, and the models need only be locally similar.

\section{Estimating depth from examples}\label{sec:estimate} 
Given a query image $I$ of some object of a certain class, our goal is to estimate a depth map $D$ for the object. To this end we use examples of {\em feasible mappings} from intensities (appearances) to depths for the class. These mappings are given in a database $S=\{M_i\}_{i=1}^n=\{(I_i, D_i)\}_{i=1}^n$, where $I_i$ and $D_i$ respectively are the image and the depth-map of an object from the class. These image-depth pairs are produced by applying standard rendering techniques to a set of example, textured, 3D geometries. For simplicity we assume first that all the image-depth pairs in the database were produced by rendering the geometries from a single viewing direction, shared also with the input image. Later in Sec.~\ref{sec:variability} we relax this assumption by demonstrating how an estimate of the camera pose may be recovered along with the depth.

Our goal is to produce a depth $D$ such that every $k\times k$ patch of mappings in $M=(I,D)$ will have a similar counterpart in $S$ (i.e., will be {\em feasible}). Specifically, we seek a depth $D$ satisfying the following two criteria:
\begin{enumerate}[i]
\item For every $k\times k$ patch of mappings in $M$, there is a similar patch in $S$, and 
\item if two patches overlap a pixel $p$, then the two database patches selected as their matches must agree of the depth at $p$. 
\end{enumerate}
We next describe how we produce depth estimates satisfying these criteria.

\subsection{Optimization scheme}\label{sec:optimization}
Given an input image $I$, we produce a depth estimate $D$ meeting the two criteria mentioned above, by building on the following simple, two-step procedure (see also Fig.~\ref{fig:optimize}): (i) At every location $p$ in $I$ we consider a $k\times k$ window around $p$ and seek a matching window in the database with a similar intensity pattern in the least squares sense (Fig.~\ref{fig:optimize}.($i$)). (ii) Finding such a window, we extract its corresponding $k\times k$ depths. We do this for all pixels in $I$, matching overlapping intensity patterns and obtaining $k^2$ depth estimates for every pixel coordinate. The depth value at every $p$ is then determined by taking an Gaussian weighted mean of these $k^2$ estimates (Fig.~\ref{fig:optimize}.($ii$)). Here, the Gaussian weights weigh in favor of estimates from patches centered closer to $p$.

Of course, there is nothing to guarantee that the depth estimate obtained by executing these two steps just once will meet our criteria. In order to produce a suitable estimate, we therefore take the current depth to be an initial guess which we then refine iteratively. We repeat the following process until convergence (see also Fig.~\ref{fig:code}): At every step we seek for every patch in $M$, a database patch similar in both intensity as well as depth, using $D$ from the previous iteration for the comparison. Thus, unlike the initial step, we now look for similar {\em mappings}. Having found new matches, we compute a new depth estimate for each pixel as before, by taking the Gaussian weighted mean of its $k^2$ estimates. In Section~\ref{sec:optimizationproof} we prove that this two-step procedure is a hard-EM optimization of a well defined target function. As such, it is guaranteed to converge to a local optimum of the target function.

Fig.~\ref{fig:code} summarizes this process. The function $getSimilarPatches$ searches $S$ for patches of mappings which match those of $M$, in the least squares sense. The set of all such matching patches is denoted ${\cal V}$. The function $updateDepths$ then updates the depth estimate $D$ at every pixel $p$ by taking the weighted mean over all depth values for $p$ in ${\cal V}$.
\begin{figure}[!ht]
\rule{\linewidth}{1pt}
\begin{tabbing}
D = \= estimateDepth($I$, $S$)\\
\> $M=(I,?)$ \\
\> repeat \= until no change in $M$\\
\> (i)\> ${\cal V}$ = getSimilarPatches($M$, $S$) \\
\> (ii)\> $D$ = updateDepths($M$, ${\cal V}$) \\
\> \> $M=(I,D)$
\end{tabbing}
\caption{{\bf Summary of the basic steps of our algorithm.}}
\label{fig:code}
\rule{\linewidth}{1pt}
\end{figure}

\subsection{Plausibility as a likelihood function}\label{sec:optimizationproof} 
We now analyze our iterative process and show that it is in fact a hard-EM optimization~\cite{KearnsMN97} of the following target function (which in turn, satisfies our criteria of Sec.~\ref{sec:estimate}). Denote by $W_p$ a $k\times k$ window from the query $M$ centered at $p$, containing both intensity values and (unknown) depth values, and denote by $V$ a similar window in some $M_i\in S$. Our target function can now be defined as 
\begin{equation}
Plaus(D|I,S) = \prod_{p\in I}\max_{V\in S}Sim(W_p,V),
\label{eq:plaus}
\end{equation}
with the similarity measure $Sim(W_p,V)$ being:
\begin{equation}
Sim(W_p, V) = \exp\left(-\frac{1}{2}(W_p-V)^T \Sigma^{-1} (W_p -
V)\right),\label{eq:sim}
\end{equation}
where $\Sigma$ is a constant diagonal matrix, its components representing the individual variances of the intensity and depth components of patches in the class. These are provided by the user as weights (see also Sec.~\ref{sec:represent}). To make this norm robust to illumination changes we normalize the intensities in
each window to have zero mean and unit variance, similarly to the normalization often applied to patches in detection and recognition methods (e.g.~\cite{Fergus:sparse}).



\begin{figure}[h!]
\begin{center}
\includegraphics[height=2.7cm, clip, trim=3.8cm 6cm 4.2cm 4.2cm] {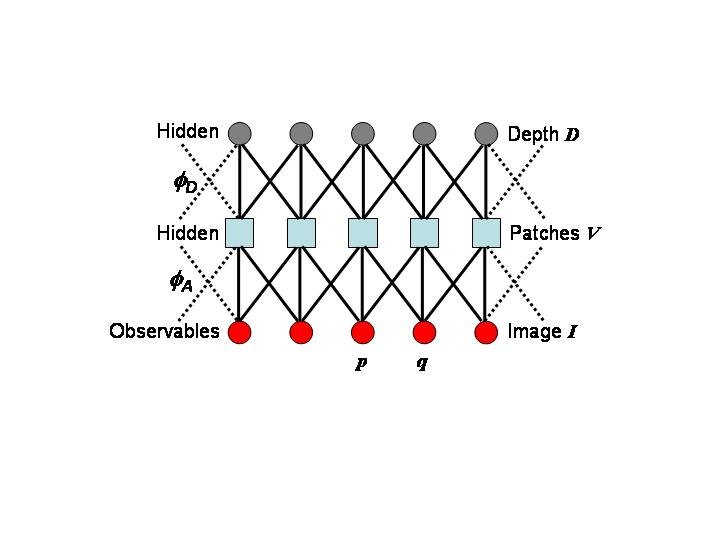}
\end{center}
\caption{{\bf Graphical model representation.} Please see text for more details.}\label{fig:graphical}
\end{figure}

In Fig.~\ref{fig:graphical} we represent the intensities of the query image $I$ as observables and the matching database patches ${\cal V}$ and the sought depth values $D$ as hidden variables. The joint probability of the observed and hidden variables can be formulated through the edge potentials by 
$$
f(I,{\cal V};D)=\prod_{p\in I} \prod_{q\in W_p} \phi_I(V_p(q),
I(q)) \cdot \phi_D(V_p(q), D(q))\label{eq:potentials}
$$

where $V_p$ is the database patch matched with an image patch $W_p$ centered at $p$ by the global assignment~${\cal V}$. Taking $\phi_I$ and $\phi_D$ to be Gaussians with different covariances over the appearance and depth respectively, implies
$$
f(I,{\cal V};D) = \prod_{p\in I} Sim(W_p, V_p).
$$
Where $Sim$ is defined in~\eqref{eq:sim}. Integrating over all possible assignments of ${\cal V}$ we obtain the likelihood function
$$
L=f(I;D)=\sum_{{\cal V}} f(I,{\cal V};D)=\sum_{{\cal V}}
\prod_{p\in I} Sim(W_p, V_p).
$$
We approximate the sum with a maximum operator. Note that this is common practice for EM algorithms, often referred to as hard-EM (e.g.,~\cite{KearnsMN97}). Since similarities can be computed independently, we can interchange the product and maximum operators, obtaining the following maximum likelihood:
$$
\max L \approx \prod_{p\in I} \max_{V\in S} Sim(W_p, V) = Plaus(D
| I, S),
$$
which is our cost function~\eqref{eq:plaus}.

The function $estimateDepth$ (Fig.~\ref{fig:code}) maximizes this measure by implementing a hard-EM optimization. The function $getSimilarPatches$ performs a hard E-step by selecting the set of assignments ${\cal V}^{t+1}$ for time $t+1$ which maximizes the posterior:
$$
f({\cal V}^{t+1}|I;D^t) \propto \prod_{p\in I}Sim(W_p, V_p).
$$
Here, $D^t$ is the depth estimate at time $t$. Due to the independence of patch similarities, this can be maximized by finding for each patch in $M$ the most similar patch in the database, in the least squares sense.

The function $updateDepths$ approximates the M-step by finding the most likely depth assignment at each pixel:
$$
D^{t+1}(p)=\arg \max_{D(p)}(-\sum_{q\in
W_p}(D(p)-depth(V_q^{t+1}(p))^2)).
$$
This is maximized by taking the mean depth value over all $k^2$ estimates $depth(V_q^{t+1}(p))$, for all neighboring pixels~$q$.

We note that optimization with Hard-EM, well known to converge to a local optimum of the target function~\cite{KearnsMN97}.

\section{Finding the right examples}\label{sec:variability} By-example, patch based approaches have become quite popular and are successfully employed for solving problems ranging from texture synthesis to recognition. The underlying assumption behind these methods is that class variability can be captured by a finite, preferably small, set of examples. Many applications can typically guarantee these conditions (notably texture synthesis). However, when the examples include non-rigid objects, objects varying in texture, or when viewing conditions are allowed to change, it becomes increasingly harder to apply these methods: Adding more examples to allow more variability (e.g., rotations of the input image in~\cite{Drori:completion}), implies larger storage requirements, longer running times, and higher risk of false matches. 

Our goal here is to handle objects viewed from any direction, non-rigid objects (e.g. hands), and objects which vary in texture (e.g. fish). Ideally, we would like to use few examples whose shape (depth) is similar to that of the object in the input image, viewed under similar conditions. This, however, implies a chicken-and-egg problem: Depth estimation requires choosing similar example objects, but knowing which objects are similar first requires a depth estimate.

Our optimization scheme provides a convenient means of solving this problem. Instead of committing beforehand to a fixed set of examples we update the set of examples, {\em on-the-fly}, alongside the optimization process. We start with an initial seed database of examples. In subsequent iterations of our optimization we drop the least used examples $M_i$ from our database, replacing them with ones deemed better suited for the depth estimation process. These are produced by on-the-fly rendering of more suitable 3D models, with viewing conditions closer to the one used in the query. In our experiments, we applied this idea to search for more similar example objects and better viewing angles. We believe that other parameters such as lighting conditions can also be similarly resolved. We next describe the details of our implementation.


\subsection{Searching for the best views} 
\begin{figure}[t]
\begin{center}
\begin{tabular}{c@{}c@{}c}
& Input image & \\
&
\includegraphics[height=1.43cm, clip, trim=1.7cm 0.4cm 1.5cm 0.5cm] {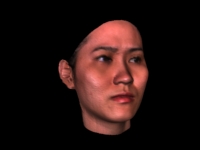}&
\\
(a)&(b)&(c)
\\
\includegraphics[height=1.8cm, clip, trim=10mm 0mm 0mm 10mm] {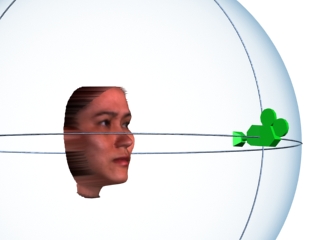}&%
\hspace{-0.5cm}\includegraphics[height=1.8cm, clip, trim=10mm 0mm 0mm 10mm] {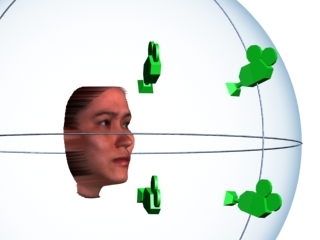}&%
\hspace{-0.5cm}\includegraphics[height=1.8cm, clip, trim=10mm 0mm 0mm 10mm] {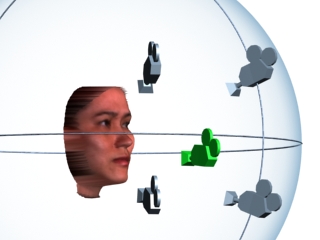}%
\\
\includegraphics[height=3cm, clip, trim=2cm 5mm 2cm 5mm] {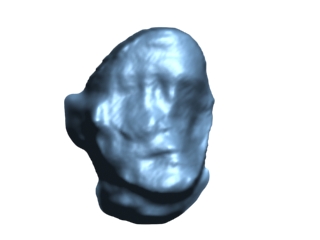}&%
\includegraphics[height=3cm, clip, trim=2cm 5mm 2cm 5mm] {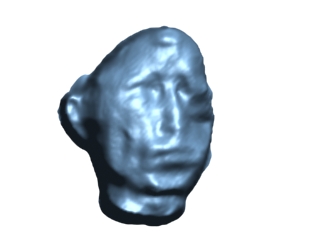}&%
\includegraphics[height=3cm, clip, trim=2cm 5mm 2cm 5mm] {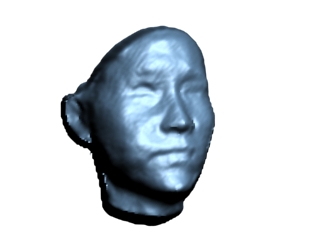}%

\end{tabular}
\caption{{\bf Depth estimate with unknown viewing angle.} A woman's face viewed from camera angle $(\alpha,\beta)=(0^\circ,-22^\circ)$. (a) Database mappings $S$ rendered with the camera at angle $(0^\circ,0^\circ)$. (b) Database generated with cameras positioned in angles $(-20^\circ,0^\circ)$, $(20^\circ,0^\circ)$, $(-20^\circ,-40^\circ)$, and $(20,-40)$, without updating the database viewing position. (c) Estimating depth while updating the database camera view on-the-fly. Starting from the angles in (b), now updating angles until convergence to
$(-6^\circ,-24^\circ)$.}\label{fig:angles}
\end{center}
\end{figure}

Fig.~\ref{fig:angles} demonstrates a depth estimation result produced by using example images generated from a single incorrect viewing angle (Fig.~\ref{fig:angles}.a) and four fixed, widely spaced viewing angles (Fig.~\ref{fig:angles}.b). Both results are inadequate.

It stands to reason that mappings from viewing angles closer to the true one, will contribute more patches to the process than those further away. We thus adopt the following scheme. We start with a small number of pre-selected views, sparsely covering parts of the viewing sphere (the gray cameras in Fig.~\ref{fig:angles}.c). The seed database $S$ is produced by taking the mappings $M_i$ of our objects, rendered from these views, and is used to obtain an initial depth estimate. In subsequent iterations, we re-estimate our views by taking the mean of the currently used angles, linearly weighted by the relative number of patches selected from each angle. We then drop from $S$ mappings originating from the least used angle and replace them with ones from the new view. If the new view is sufficiently close to one of the remaining angles (e.g., its distance to an existing view falls below a predefined threshold), we instead increase the number of objects to maintain the size of $S$. Fig.~\ref{fig:angles}.c presents a result obtained with our angle update scheme.

Although methods exist which accurately estimate the viewing angle~\cite{Osadchy:Synergistic,Romdhani:identification}, we preferred embedding this estimation in our optimization. To understand why, consider non-rigid classes such as the human body where posture cannot be captured with only a few parameters. Our approach uses information from several viewing angles simultaneously, without pre-committing to any single view.

\subsection{Searching for the best example objects} Although we have collected at least 40 objects in each database, we can use no more than 12 objects at a time in the optimization, as it becomes increasingly difficult to handle larger sets. We select the ones used in practice as follows. Starting from a set of arbitrarily selected seed objects, at every update step we drop those leased referenced. We then scan the remainder of our objects for those who's depth, $D_i$, best matches the current depth estimate $D$ (i.e., $(D-D_i)^2$ is smallest, $D$ and $D_i$ center aligned) adding them to the database instead of those dropped. In practice, a fourth of our objects were replaced after the first iteration of our process.

\section{Preserving global structure}\label{sec:structural} 
The scheme described in Sec.~\ref{sec:optimization}, makes an implicit stationarity assumption~\cite{Wei:fastsynthesis}: Put simply, the probability for the depth at any pixel, given those of
its neighbors, is the same throughout the output image. This is generally untrue for structured objects, where depth often depends on position. For example, the probability of a pixel's depth being ``tip-of-the-nose high'' is different at different locations of a face. To overcome this problem, we suggest enforcing {\em non}-stationarity by adding additional constraints to the patch matching process. Specifically, we encourage selection of patches from similar semantic parts by favoring patches which match not only in intensities and depth, but also in position relative to the centroid of the  input depth-map. This is achieved by adding relative position values to each patch of mappings in both the database and the query image.

\begin{figure}[!t]
\begin{center}
\begin{tabular}{ccc}
\includegraphics[height=3.3cm, clip, trim=0.4cm 0.1cm 0.4cm 0.1cm] {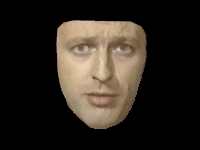}&%
\hspace{2mm}
\includegraphics[height=3.5cm, clip, trim=2cm 5mm 2cm 5mm] {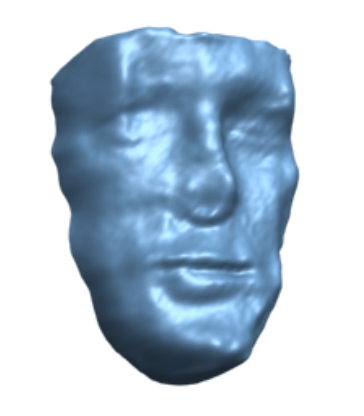}&%
\
\includegraphics[height=3.5cm, clip, trim=2cm 5mm 2cm 5mm] {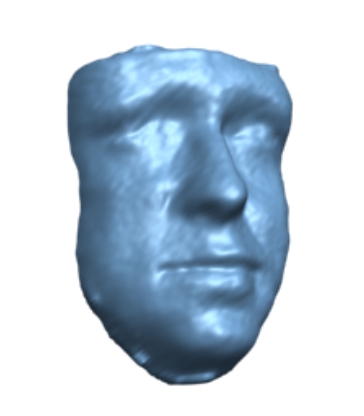}\\%
(a) & (b) & (c)
\end{tabular}
\end{center}

\caption{{\bf Preserving relative position.} (a) Input image. (b) Depth estimate without position preservation constraints and (c) with them.} \label{fig:coherence} 
\end{figure}

Let $p=(x,y)$ be the coordinates of a pixel in $I$ (the center of the patch $W_p$) and let $(x_c, y_c)$ be the coordinates of the center of mass of the area occupied by non background depths in the current depth estimate $D$. We add the values $(\delta x, \delta y) = (x-x_c, y-y_c)$, to each such patch $W_p$ and similar values to all database patches (i.e., by using the center of each depth image $D_i$ for $(x_c, y_c)$). These values now force the matching process to find patches similar in both mapping and global position. Fig.~\ref{fig:coherence} demonstrates a reconstruction result with and without these constraints.

If the query object is segmented from the background, an initial estimate for the query's centroid can be obtained from the foreground pixels. Alternatively, this constraint can be applied only after an initial depth estimate has been computed (i.e., Sec.~\ref{sec:estimate}).

\section{Additional applications}\label{sec:app}
One of the appealing aspects of this example-based approach is that other problems besides depth estimation may be similarly solved with little or no change to the method we described. Specifically, we have thus far considered example mappings from appearances to depths. This has allowed us to estimate depths for input appearances. We next show how by defining alternative mappings we obtain solutions to additional problems within the very same framework.

\subsection{Backside reconstruction}\label{sec:backside}
What can be said about the shape of a surface which does not appear in the image? Methods for depth estimation have predominantly focused on estimating the shapes of surfaces visible in images. Here we suggest that an input image may contain sufficient cues which, coupled with additional examples, may allow us to guess the shapes of surfaces even when they are occluded in the image. Specifically, we next demonstrate how the shape of the occluded backside of an object may be estimated by using the same process described thus far, from a single image of the object's front. 

We consider a database of mappings containing not appearances and their corresponding depths, but rather depths and a corresponding second depth layers (or in general, multiple depth layers). This additional depth layer is taken here to be the depth at the back of the objects viewed. Once again, we generate this database by applying standard rendering techniques to our example 3D geometries. We thus obtain a database $S'=\{M_i'\}_{i=1}^n=\{(D_i, D_i')\}_{i=1}^n$, where where $D_i'$ is the  second depth layer.

Having recovered the visible depth of an object (its depth map, $D$), we define the mapping from visible to occluded depth as $M'(p)=(D(p), D'(p))$, where $D'$ is its second depth layer. Synthesizing $D'$ can now proceed similarly to the synthesis of the visible depth layers. We note that this second depth layer may indeed have little in common with the true depth at the object's back. It is, however, reminiscent of the image hole-filling problem in attempting to produce a plausible estimate of this information, where none other exists.

\subsection{Automatic depth-map colorization}\label{sec:app_colorization}
An additional problem may be solved by reversing our original mappings. Here we propose an application similar in spirit to the problem faced by a sculptor when applying paint to enhance the appearance of statues. Given an input depth-map, our goal is to fabricate a tailor made image-map for the depth. The motivation for doing this comes from the graphics community, where considerable effort is put into developing automatic 3D colorization techniques. Here we achieve this goal by simply switching the roles of the intensities and depths in the example mappings: We now use a database $S=\{M_i'\}_{i=1}^n=\{(D_i, I_i)\}_{i=1}^n$. Given an input depth map $D$, our goal is now to produce an image map $I$ such that $M'=(D, I)$ consists of feasible mappings from shape to intensities.

We have found that for this particular application, on-the-fly database update is unnecessary, as our input is a 3D shape, allowing us to easily select similar shapes from the database before commencing with the optimization. We thus choose for synthesis a small number (often as small as one or two) of models who's depths best match the input depth in the least squares sense. These are used throughout the synthesis process. We note that when only one database object is used, our method effectively morphs its image-map to fit the 3D features of the novel input depth (See Sec.~\ref{sec:experiments}).

\section{Implementation and results}\label{sec:results}
\subsection{Representing mappings} \label{sec:represent}
\begin{figure}[!t]
\begin{center}

\includegraphics[height=2.6cm, clip, trim=0cm 0cm 0cm 0cm] {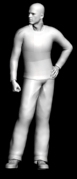}~%
\includegraphics[height=2.6cm, clip, trim=0cm 0cm 0cm 0cm] {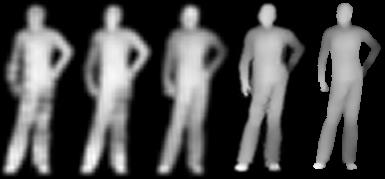}~%
\includegraphics[height=2.6cm, clip, trim=0cm 0cm 0cm 0cm] {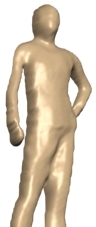}%
\caption{{\bf Depth estimates at multiple resolutions.} From left to right, input image, five intermediate depth-map estimates from different resolutions, and a zoomed in view of our output reconstruction.}\label{fig:usf1042}
\end{center}
\end{figure}

\begin{figure}[!t]
\begin{center}
\begin{tabular}[b]{c@{~~~~~~}c}
\includegraphics[height=2.8cm, clip, trim=0mm 0mm 10mm 0mm] {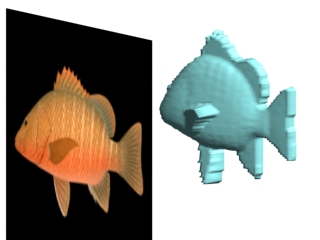}&%
\includegraphics[height=2.8cm, clip, trim=0mm 0mm 10mm 0mm] {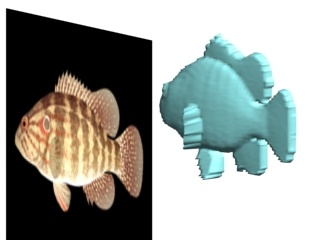}
\\
\includegraphics[height=2.8cm, clip, trim=0mm 0mm 15mm 0mm] {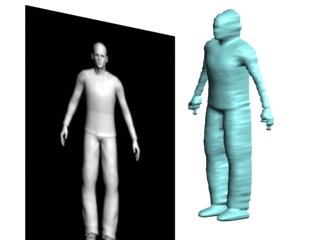}&%
\includegraphics[height=2.8cm, clip, trim=0mm 0mm 15mm 0mm] {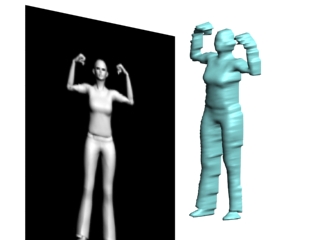}%
\end{tabular}
\caption{{\bf Database mappings used as examples.} In the top row, two
appearance-depth images, out of the 67 in the Fish database.
Bottom row, two of 50 pairs from our Human-posture
database.}\label{fig:DB}
\end{center}
\end{figure}
For the purpose of depth reconstruction, the mapping at each pixel in $M=(I,D)$, and similarly every $M_i=(I_i,D_i)$, encodes both appearance and depth (See examples in Fig.~\ref{fig:DB}). In practice, the appearance component of each pixel is its intensity and high frequency values, as encoded in the Gaussian and Laplacian pyramids of $I$~\cite{Burt:pyramids}. We have found direct synthesis of depths to result in low frequency noise (e.g.,
``lumpy'' surfaces). We thus estimate a Laplacian pyramid of the depth, producing the final depth by collapsing the depth high-frequencies estimates from all scales. In this fashion, low frequency depths are synthesized in the course scale of the pyramid and only sharpened at finer scales (See example in Fig.~\ref{fig:usf1042}). For depth colorization we used mappings from depths and depth high frequencies to Y, Cb, Cr components, also computed at different scales of the Gaussian and Laplacian pyramids. 

\begin{figure*}[th]
\begin{center}
\begin{tabular}[b]{c@{\extracolsep{3mm}}c@{}c@{\extracolsep{1cm}}c@{\extracolsep{3mm}}c@{}c@{}c}
\includegraphics[height=3cm, clip, trim=2cm 7.5cm 1.5cm 0] {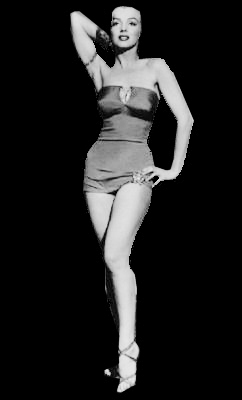}&
\includegraphics[height=3cm, clip, trim=3cm 0cm 3.2cm 0cm] {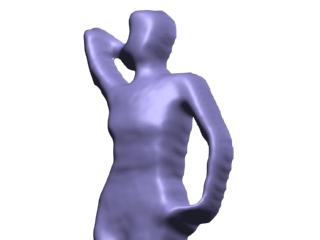}&%
\includegraphics[height=3cm, clip, trim=5mm 2mm 5mm 2mm] {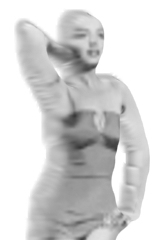}&%
\includegraphics[height=3cm, clip, trim=0cm 0cm 0cm 0] {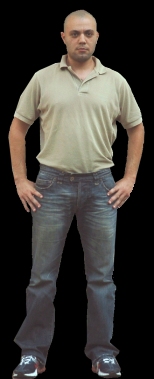}&%
\includegraphics[height=3cm, clip, trim=3cm 0cm 3.2cm 0cm] {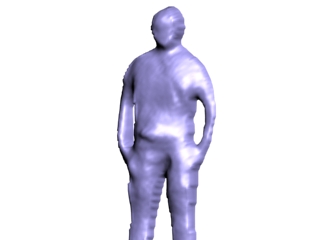}&%
\includegraphics[height=3cm, clip, trim=5mm 2cm 2mm 0mm] {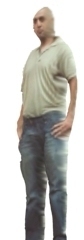}&
\includegraphics[height=3cm, clip, trim=3.5cm 0cm 3.5cm 0.5cm] {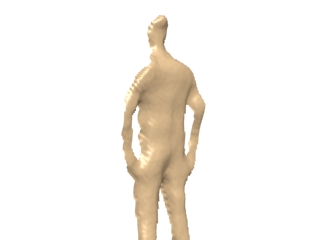}%
\end{tabular}
\caption{{\bf Full body depth results.} Left to right: Input image,
the output depth without and with texture, input image of a man,
output depth, textured view of the output, output estimate of the
depth at the {\bf back}. Man results shown
zoomed-in.}\label{fig:MarilynAndAvi}
\end{center}
\end{figure*}

\begin{figure*}[th]
\begin{center}
\begin{tabular}[b]{cc@{}cc@{}c@{}c}
\includegraphics[height=2.3cm, clip, trim=1mm 0mm 1mm 0mm] {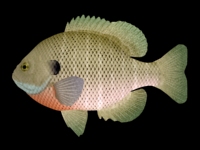}&
\includegraphics[height=2.3cm, clip, trim=3.5cm 1cm 4cm 1cm] {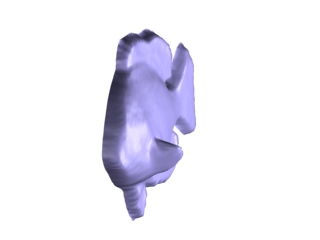}&%
\includegraphics[height=2.3cm, clip, trim=0mm 0cm 0mm 0mm] {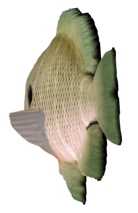}%
\includegraphics[height=2.3cm, clip, trim=0mm 0mm 0mm 0mm] {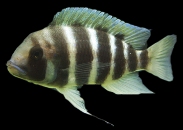}&
\hspace{-0.5cm}\includegraphics[height=2.3cm, clip, trim=2.3cm 1cm 1.8cm 0.5cm] {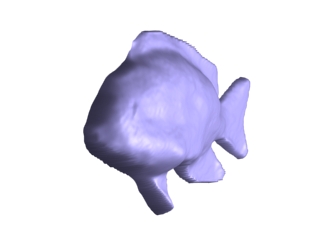}&%
\hspace{-0.2cm}\includegraphics[height=2.3cm, clip, trim=0mm 0mm 0mm 0mm] {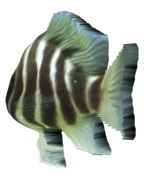}%
\end{tabular}
\end{center}
\caption{{\bf Two fish depth results.} Left to right: Input image (removed from the example database); estimated depth and a textured view of the output; Input image; estimated depth and a textured view of the output.}\label{fig:fish}
\end{figure*}

Different patch components, including relative positions, contribute different amounts of information in different classes, as reflected by their different variance (i.e., $\Sigma$ in the definition of $Sim$, Eq.~\ref{eq:sim}). For example, faces are highly structured, thus, position plays an important role in their
reconstruction. On the other hand, due to the variability of human postures, relative position is less reliable for that class. We therefore amplify different components of each patch, $W_p$, of mappings for different classes, by weighting them differently. Section~\ref{sec:experiments} present weights computed automatically for different object classes and quantitative results obtained with these weights.


Finally, we note that, in principle, database objects may come in any coordinate system, and in particular their depth values can be shifted (i.e., $z^{\prime} = z + z_0$). This may pose a problem when combining depths from different objects to form a single estimate. A possible solution would be to synthesize surface normals instead of depths (as in, e.g.,~\cite{Huang:ExemplarReconstruction}). Doing so, however, raises the problem of dealing with depth discontinuities. Here we chose instead to produce our examples in a common frame of reference by setting $z=0$ at the centroid of the 3D object and performing the reconstruction in this common frame of reference.



\subsection{Experiments}\label{sec:experiments}
In our reconstruction and colorization experiments, we used the following data sets. 52 Hand and 45 Human-posture objects, produced by exporting built-in models from the Poser software, 77 busts from the USF head database~\cite{Sarkar:usfdb}, and a fish database~\cite{Toucan:fishdb} containing 41 models. In addition, for the colorization
experiments we used a database of five human figures courtesy of Cyberware~\cite{Cyberware}. Our objects are stored as textured 3D triangulated meshes. We can thus render them to produce example mappings using any standard rendering engine. Example mappings from the fish and human posture data-sets are displayed in Fig.~\ref{fig:DB}. We used 3D Studio Max for rendering the images and depth-maps. We preferred pre-rendering the images and depth maps instead of rendering different objects from different angles on-the-fly. Thus, we trade rendering times with disk access times and large storage. Note that this is an implementation decision; at any one time we load only a small number of images to memory. The angle update step (Sec.~\ref{sec:variability}) therefore selects the
existing pre-rendered angle closest to the mean angle.

\begin{figure}[th]
\begin{center}
\begin{tabular}[b]{c@{}c@{}c@{}c}
\includegraphics[height=3cm, clip, trim=4mm 0mm 4mm 0mm] {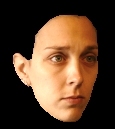}&
\begin{tabular}[b]{c@{}c}
\includegraphics[height=1.43cm, clip, trim=17mm 0 17mm 0]{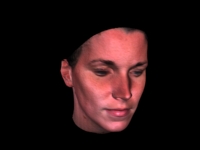}&
\includegraphics[height=1.43cm, clip, trim=17mm 0 17mm 0]{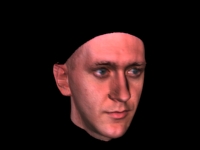}\\
\includegraphics[height=1.43cm, clip, trim=17mm 0 17mm 0]{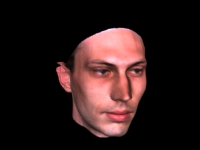}&
\includegraphics[height=1.43cm, clip, trim=17mm 0 17mm 0]{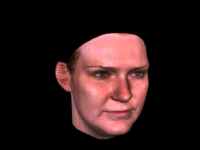}
\end{tabular}
&
\includegraphics[height=2.6cm, clip, trim=188mm 8cm 16cm 8cm] {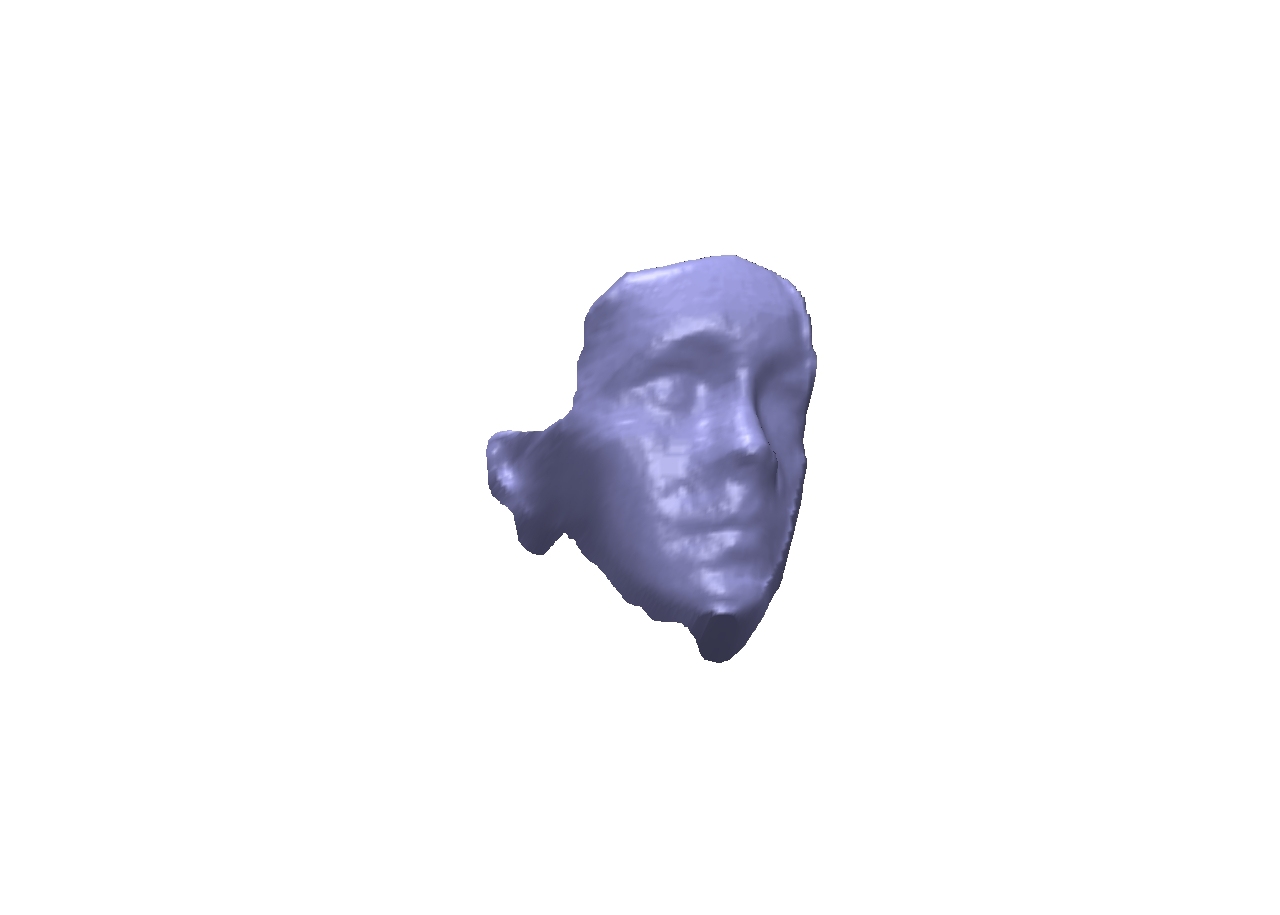}&
\includegraphics[height=2.6cm, clip, trim=188mm 8cm 155mm 8cm] {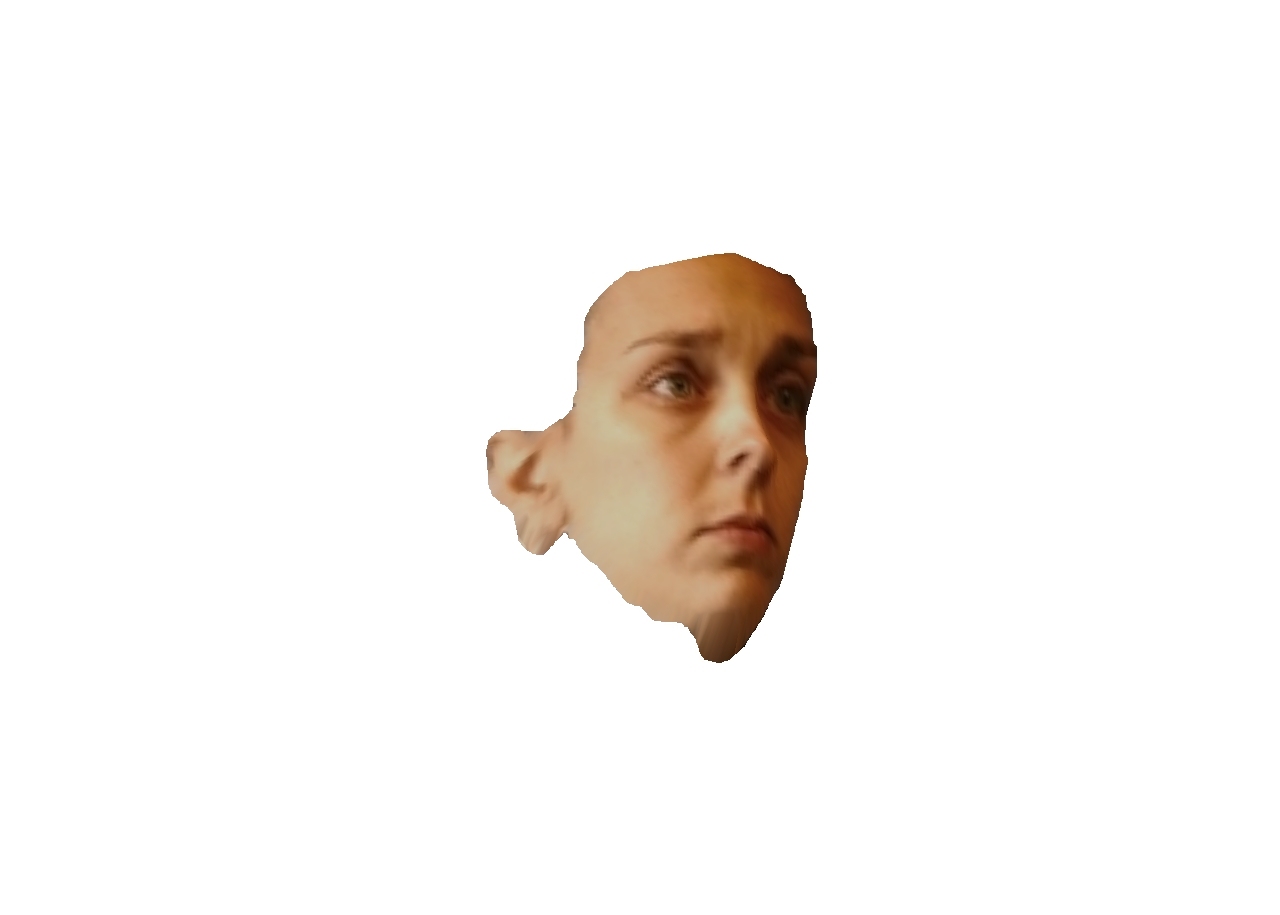}%
\\
\includegraphics[height=3cm, clip, trim=0mm 0mm 0mm 0mm] {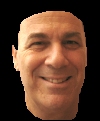}&
\begin{tabular}[b]{c@{}c}
\includegraphics[height=1.43cm, clip, trim=17mm 0 17mm 0]{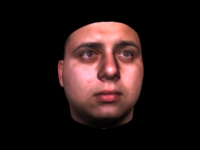}&
\includegraphics[height=1.43cm, clip, trim=17mm 0 15mm 0]{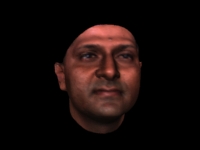}\\
\includegraphics[height=1.43cm, clip, trim=17mm 0 17mm 0]{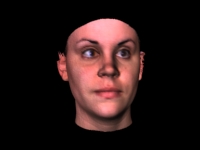}&
\includegraphics[height=1.43cm, clip, trim=17mm 0 15mm 0]{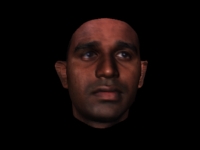}
\end{tabular}
&
\hspace{-2mm}\includegraphics[height=2.6cm, clip, trim=175mm 8cm 16cm 8cm] {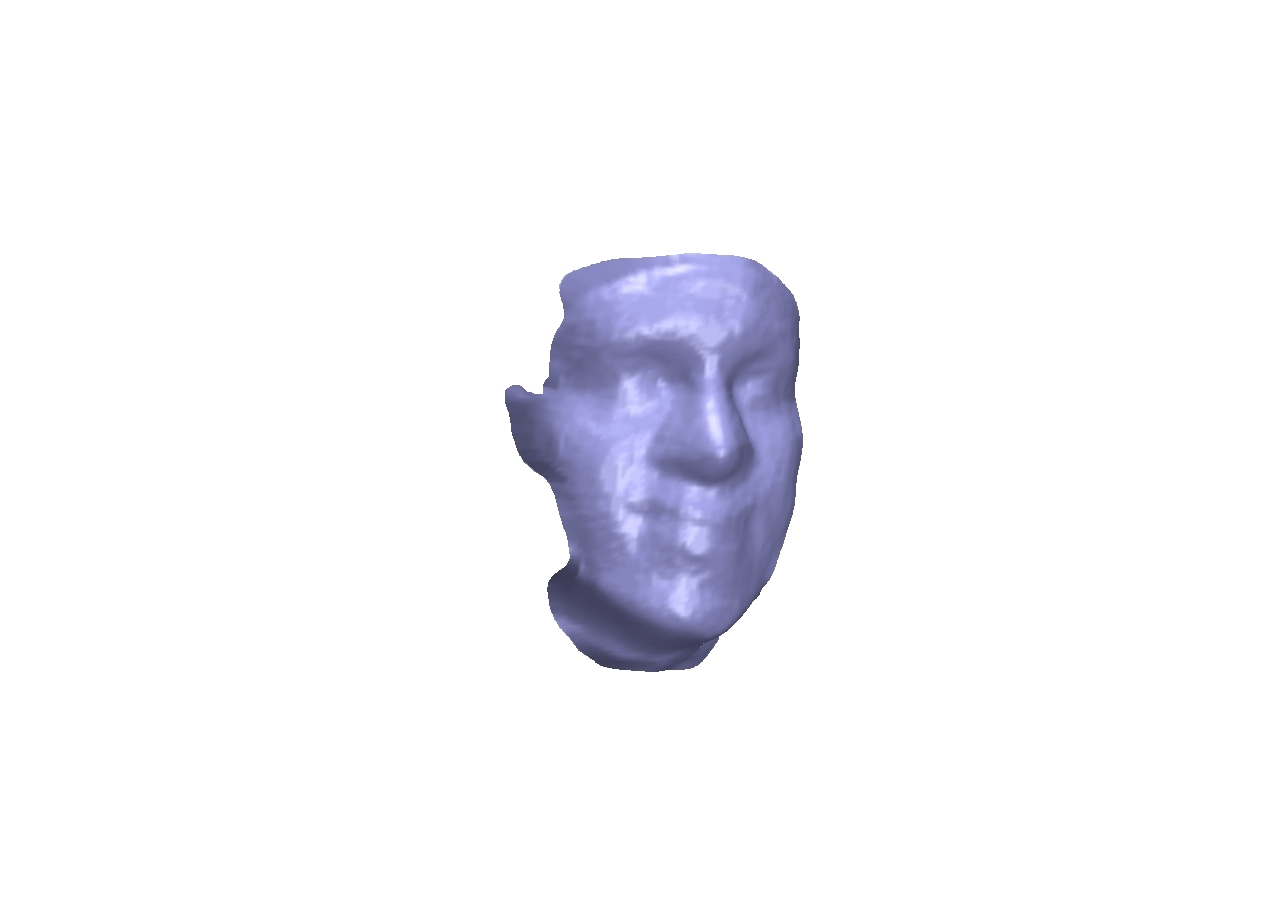}&%
\includegraphics[height=2.6cm, clip, trim=188mm 8cm 165mm 8cm] {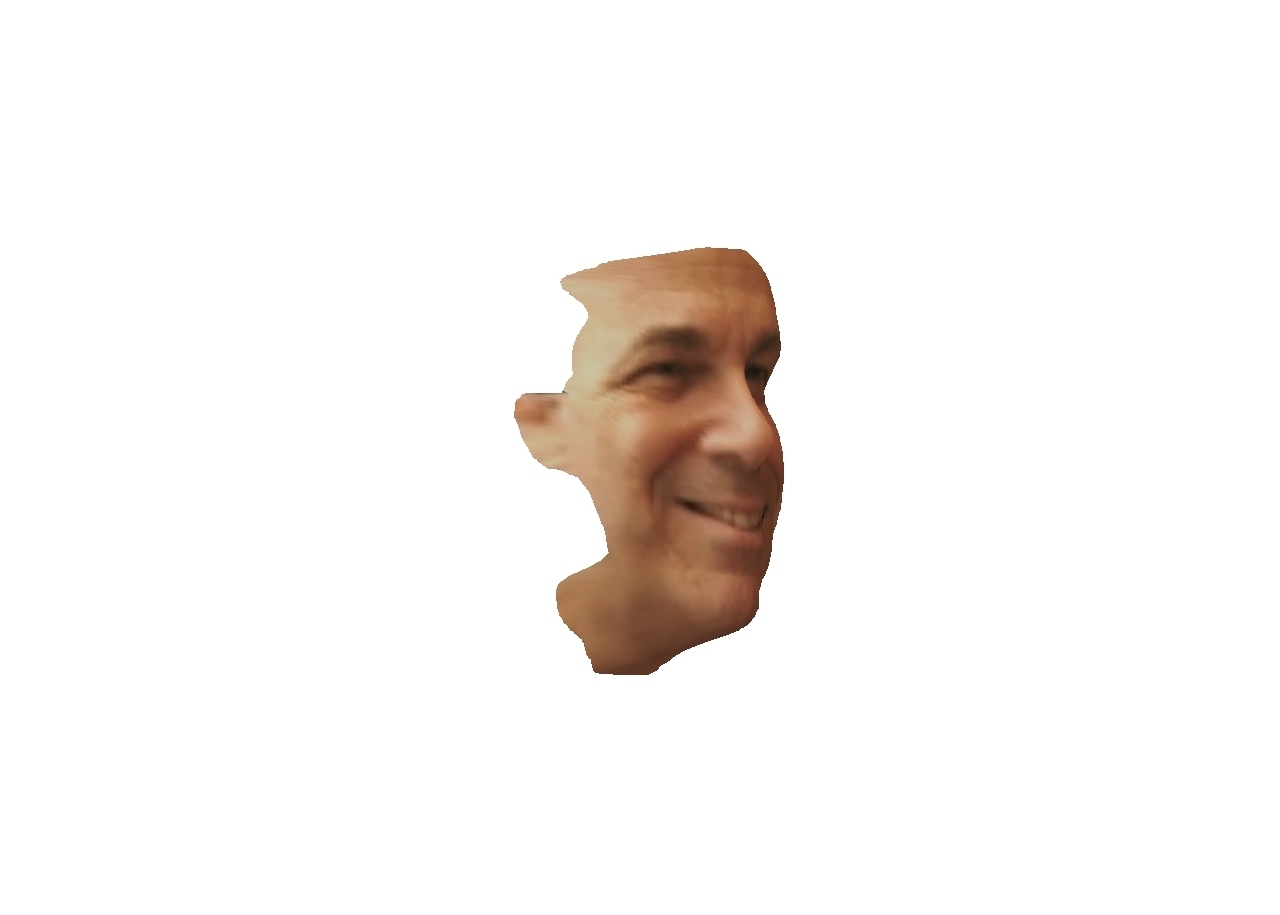}%
\end{tabular}
\caption{{\bf Two face depth results.} Left to right: Input image,
four most referenced database images in the last iteration, our
output depth without and with texture, input image, four most referenced
database images in the last iteration, our output depth without and with
texture.}\label{fig:faces}
\end{center}
\end{figure}

\begin{figure}[!t]
\begin{center}
\begin{tabular}[b]{ccc}

\includegraphics[height=3.3cm, clip, trim=0mm 0mm 0mm 0mm] {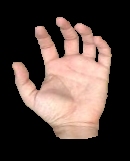}&%
\includegraphics[height=3.15cm, clip, trim=2.7cm 0cm 2.7cm 0.5cm] {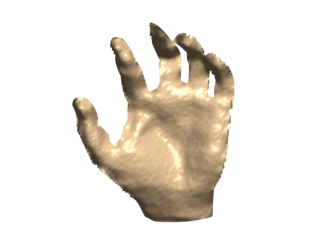}&%
\includegraphics[height=3.15cm, clip, trim=3.2cm 1cm 3.5cm 1cm] {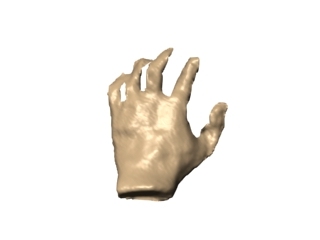}%
\\
(a)&(b)&(c)
\end{tabular}
\caption{{\bf Hand depth result.} (a) Input image. (b) Our output. (c) Output estimate for the {\bf back} of the
hand.}\label{fig:hand}
\end{center}
\end{figure}

\noindent{\bf Depth Reconstruction.} Our results include depth estimates from single images for structured objects such as faces (Fig.~\ref{fig:angles},~\ref{fig:coherence},~\ref{fig:faces}) as well as highly non-rigid objects such as hands (Fig.~\ref{fig:optimize} and~\ref{fig:hand}) and full bodies (Fig.~\ref{fig:usf1042} and~\ref{fig:MarilynAndAvi}) in various postures. These results include in particular objects with higher than zero genus (e.g.,~Fig.~\ref{fig:MarilynAndAvi}) and objects with depth discontinuities such as the fingers of the hand in Fig.~\ref{fig:optimize} and the fin of the left fish in Fig.~\ref{fig:fish}. Additionally, we show that our method can produce estimates even when the objects in the image are highly textured as in the fish examples in Fig.~\ref{fig:fish}. Similarly to shape-from-shading methods, we assume here that the query objects were pre-segmented from their background and then aligned with a single preselected database image to solve for scale and image-plane rotation differences (see, e.g.,~\cite{Kemelmacher:moldingfaces}).

\begin{figure}[!ht]
\begin{center}
\begin{tabular}{cc@{}c}
\multirow{2}{*}[1.3cm]{\includegraphics[height=3.2cm, clip, trim=1.5cm 0.1cm 1.5cm 0.5cm] {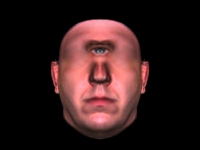}}&%
\includegraphics[height=3.2cm, clip, trim=2.5cm 0.7cm 2.5cm 0.7cm] {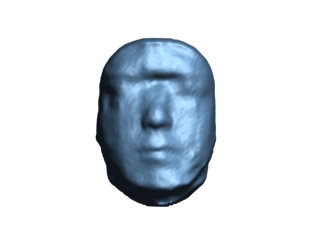}&%
\includegraphics[height=3.2cm, clip, trim=3cm 0mm 3cm 0mm] {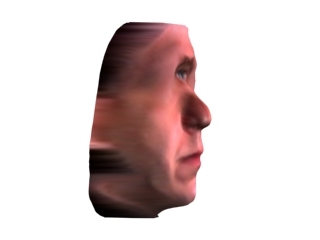}\\%
 &
\includegraphics[height=2.6cm, clip, trim=0cm 0cm 0cm 0cm] {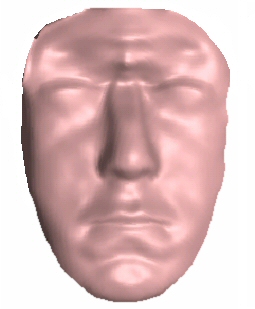}&%
\includegraphics[height=2.6cm, clip, trim=0cm 0mm 0cm 0mm] {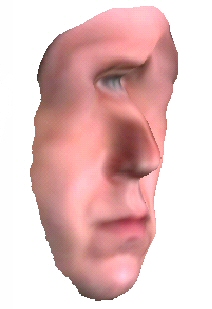}\\ 
(a) & (b) & (c)
\end{tabular}
\end{center}
\caption{{\bf Cyclops depth result.} (a) Input image of a Cyclopean face. (b-c) {\bf Top row}: Our depth estimate rendered in 3D and a textured view; {\bf Bottom row}: Depth estimate produced using the method of~\cite{Kemelmacher:moldingfaces} rendered in 3D and a textured view. Both methods use example shapes of {\em binocular faces}. Although~\cite{Kemelmacher:moldingfaces} produce more detailed estimates, their strong prior on a face shape results in a face with three separate eyes.}\label{fig:cyclops}
\end{figure}

It is interesting to compare our method with a method tailored to reconstructing face depths~\cite{Kemelmacher:moldingfaces} (see Fig.~\ref{fig:cyclops}). For a non-standard face (a cyclops) our patch based method appropriately produces a shape estimate with only one eye socket, using examples of typical, binocular faces. Although~\cite{Kemelmacher:moldingfaces} produce a finer detailed estimate, their strong global face shape prior results in an estimate erroneously containing three separate eyes.
%

Occluded back estimation results (Sec.~\ref{sec:backside}) are presented in Fig.~\ref{fig:MarilynAndAvi} and~\ref{fig:hand}. Two non-structured objects with non-trivial backs were selected for these tests.

\begin{figure}[!ht]
\begin{center}
\begin{tabular}{c@{~~~~~~~~~~}c}
\includegraphics[height=3.5cm, clip, trim=0.4cm 0.1cm 0.4cm 0.1cm] {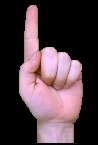}%
\includegraphics[height=3.5cm, clip, trim=3.1cm 5mm 3.8cm 5mm] {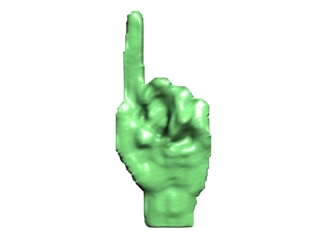} &%
%
%
\includegraphics[height=3.5cm, clip, trim=0.2cm 0.4cm 0.4cm 0.1cm] {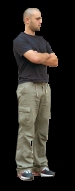}%
\includegraphics[height=3.5cm, clip, trim=3.1cm 5mm 4cm 5mm] {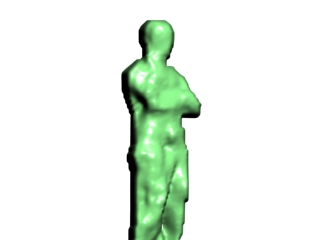}\\%
(a) & (b)
\end{tabular}
\end{center}
\caption{{\bf Failures.} (a) Hand reconstructions are particularly
challenging, as they are largely untextured, and can vary greatly
in posture. (b) The uniform black shirt differed greatly from the
ones worn by our database objects (see Fig.~\ref{fig:DB}). No
reliable matches could thus be found, resulting in a lumpy
surface. Resulting surface presented from a zoomed-in
view.}\label{fig:failure}
\end{figure}

In general, the quality of our depth estimate depends on the database used, the input image, and how the two match. Fig.~\ref{fig:failure} presents some failed results. In Fig.~\ref{fig:failure}(a) our method's lack of a global prior resulted in the middle finger which both points forward and downwards. In Fig.~\ref{fig:failure}(b) the subject was waring a dark shirt and bright trousers, very different from the uniformly colored objects in the database.

\begin{table}[ht]\centering
\caption{{\bf Depth estimation database parameters.} m - Number of mappings (objects) $M_i$ used for synthesis. Weights for intensities, depth, and relative position components. Patch sizes were $7\times 7$ pixels in all tests.}\label{tbl:depth_params}
\begin{tabular}{|l|c|c|} \hline
\emph{DB Name} & \emph{m} & \emph{Weights}\\
\hline \hline
Human-posture& 4 & 0.2140, 0.1116, 0.0092\\
Hands& 5 & 0.1294, 0.1120, 0.0103\\
Fish& 5 & 0.1361, 0.1063, 0.0116\\
\hline
\end{tabular}
\end{table}

\begin{table}[ht]\centering
\caption{{\bf Depth estimation quantitative results.} Mean and STD results of L1 distances between estimated depths and ground truth.}\label{tbl:depth_res}
\begin{tabular}{|l|c|c|c|} \hline
\emph{DB Name} & \emph{Baseline} & \emph{Make3D~\cite{Saxena:08:make3d}} & \emph{Our method}\\
\hline \hline
Posture& .040 $\pm$ .01 & .248 $\pm$ .09 & .023 $\pm$ .00\\
Hands& .039 $\pm$ .02& .228 $\pm$ .05 & .026 $\pm$ .01\\
Fish& .044 $\pm$ .02& .277 $\pm$ .12 & .036 $\pm$ .02\\
\hline
\end{tabular}
\end{table}

\noindent{\bf Quantitative depth estimation results.} To evaluate the performance of our algorithm, we ran leave-one-out, depth estimation tests on the human-posture, hands, and fish data sets. We used five randomly selected objects from each object class as training. Their images and ground truth depths were used to {\em automatically} search for optimal weights for the three components of the mappings (appearance, depth, and relative position). We used a direct simplex search method to search for these three parameters separately for each class minimizing the error between our depth estimate and the known ground truth. The parameters thus obtained are presented in Table~\ref{tbl:depth_params}. The search for parameters was performed only once for each class, and the parameters obtained were applied to all input images. We did not screen our results for any failures and all depth estimates were included when computing the global result.

We next estimated the depths of the objects which were not included in the training. The quality of our estimates was compared against a naive selection of the depths belonging to the database objects most similar in appearance to the test images. We have included the accuracy obtained by applying the method of~\cite{Saxena:08:make3d}, using their own code\footnote{Make3D code available from \url{http://make3d.cs.cornell.edu}}. We note, however, that this method was developed and optimized for outdoor scenes, and so it is not surprising that it should under-perform when applied to images of objects. Table~\ref{tbl:depth_res} summarizes our results, presenting a comparison of the mean and STD L1 distances between the ground truth depths and the estimated depths produced by our method, Make3D~\cite{Saxena:08:make3d} and the naive selection as baseline.

\begin{figure*}
\begin{center}
\begin{tabular}[b]{c@{}c@{\hspace{-0.2cm}}c@{\hspace{-0.2cm}}c@{\hspace{0.5cm}}c@{}c@{\hspace{-0.2cm}}c@{\hspace{-0.2cm}}c@{}}

Input&Ground&DB&Our&Input&Ground&DB&Our\\
image&truth&examples&result&image&truth&examples&result\\
\includegraphics[height=2.2cm, clip, trim=0cm 0cm 0cm 0cm] {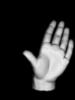}&%
\includegraphics[height=2.2cm, clip, trim=0cm 0cm 0cm 0cm] {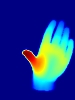}&%
\begin{tabular}[b]{c@{}c}
\includegraphics[height=1.02cm, clip, trim=0cm 0cm 0cm 0cm] {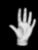}&%
\includegraphics[height=1.02cm, clip, trim=0cm 0cm 0cm 0cm] {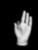}\\%
\includegraphics[height=1.02cm, clip, trim=0cm 0cm 0cm 0cm] {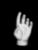}&%
\includegraphics[height=1.02cm, clip, trim=0cm 0cm 0cm 0cm] {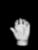}%
\end{tabular}&
\includegraphics[height=2.2cm, clip, trim=0cm 0cm 0cm 0cm] {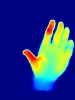}&%

\includegraphics[height=2.2cm, clip, trim=0cm 0cm 0cm 0cm] {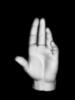}&%
\includegraphics[height=2.2cm, clip, trim=0cm 0cm 0cm 0cm] {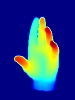}&%
\begin{tabular}[b]{c@{}c}
\includegraphics[height=1.02cm, clip, trim=0cm 0cm 0cm 0cm] {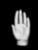}&%
\includegraphics[height=1.02cm, clip, trim=0cm 0cm 0cm 0cm] {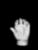}\\%
\includegraphics[height=1.02cm, clip, trim=0cm 0cm 0cm 0cm] {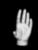}&%
\includegraphics[height=1.02cm, clip, trim=0cm 0cm 0cm 0cm] {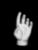}%
\end{tabular}&
\includegraphics[height=2.2cm, clip, trim=0cm 0cm 0cm 0cm] {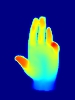}\\%

\includegraphics[height=2.2cm, clip, trim=0cm 0cm 0cm 0cm] {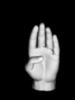}&%
\includegraphics[height=2.2cm, clip, trim=0cm 0cm 0cm 0cm] {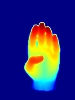}&%
\begin{tabular}[b]{c@{}c}
\includegraphics[height=1.02cm, clip, trim=0cm 0cm 0cm 0cm] {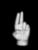}&%
\includegraphics[height=1.02cm, clip, trim=0cm 0cm 0cm 0cm] {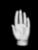}\\%
\includegraphics[height=1.02cm, clip, trim=0cm 0cm 0cm 0cm] {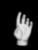}&%
\includegraphics[height=1.02cm, clip, trim=0cm 0cm 0cm 0cm] {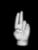}%
\end{tabular}&
\includegraphics[height=2.2cm, clip, trim=0cm 0cm 0cm 0cm] {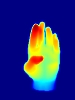}&%

\includegraphics[height=2.2cm, clip, trim=0cm 0cm 0cm 0cm] {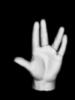}&%
\includegraphics[height=2.2cm, clip, trim=0cm 0cm 0cm 0cm] {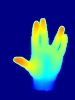}&%
\begin{tabular}[b]{c@{}c}
\includegraphics[height=1.02cm, clip, trim=0cm 0cm 0cm 0cm] {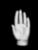}&%
\includegraphics[height=1.02cm, clip, trim=0cm 0cm 0cm 0cm] {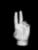}\\%
\includegraphics[height=1.02cm, clip, trim=0cm 0cm 0cm 0cm] {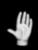}&%
\includegraphics[height=1.02cm, clip, trim=0cm 0cm 0cm 0cm] {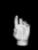}%
\end{tabular}&
\includegraphics[height=2.2cm, clip, trim=0cm 0cm 0cm 0cm] {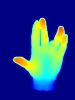}\\%

\end{tabular}
\caption{{\bf Hand depth estimates.} Four out of the 52 hand-object depth estimates computed using automatically obtained weights (see Table~\ref{tbl:depth_params} for parameter values). In both columns from left to right: input image, its ground truth, four of the five {\em automatically} selected database examples used for the reconstructions, and our output estimate.}\label{fig:batch_hands}
\end{center}
\end{figure*}

\begin{figure*}
\begin{center}
\begin{tabular}[b]{c@{}c@{\hspace{-0.2cm}}c@{\hspace{-0.2cm}}c@{\hspace{0.5cm}}c@{}c@{\hspace{-0.2cm}}c@{\hspace{-0.2cm}}c@{}}

Input&Ground&DB&Our&Input&Ground&DB&Our\\
image&truth&examples&result&image&truth&examples&result\\
\includegraphics[height=2.2cm, clip, trim=0cm 0cm 0cm 0cm] {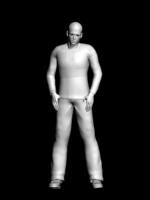}&%
\includegraphics[height=2.2cm, clip, trim=0cm 0cm 0cm 0cm] {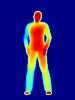}&%
\begin{tabular}[b]{c@{}c}
\includegraphics[height=1.02cm, clip, trim=0cm 0cm 0cm 0cm] {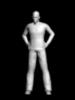}&%
\includegraphics[height=1.02cm, clip, trim=0cm 0cm 0cm 0cm] {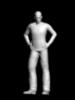}\\%
\includegraphics[height=1.02cm, clip, trim=0cm 0cm 0cm 0cm] {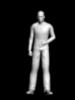}&%
\includegraphics[height=1.02cm, clip, trim=0cm 0cm 0cm 0cm] {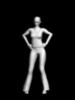}%
\end{tabular}&
\includegraphics[height=2.2cm, clip, trim=0cm 0cm 0cm 0cm] {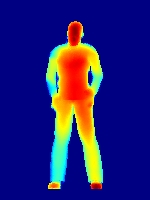}&%

\includegraphics[height=2.2cm, clip, trim=0cm 0cm 0cm 0cm] {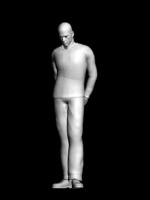}&%
\includegraphics[height=2.2cm, clip, trim=0cm 0cm 0cm 0cm] {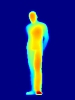}&%
\begin{tabular}[b]{c@{}c}
\includegraphics[height=1.02cm, clip, trim=0cm 0cm 0cm 0cm] {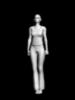}&%
\includegraphics[height=1.02cm, clip, trim=0cm 0cm 0cm 0cm] {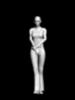}\\%
\includegraphics[height=1.02cm, clip, trim=0cm 0cm 0cm 0cm] {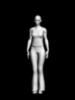}&%
\includegraphics[height=1.02cm, clip, trim=0cm 0cm 0cm 0cm] {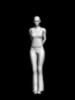}%
\end{tabular}&
\includegraphics[height=2.2cm, clip, trim=0cm 0cm 0cm 0cm] {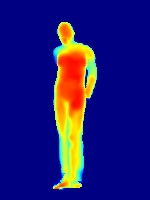}\\%

\includegraphics[height=2.2cm, clip, trim=0cm 0cm 0cm 0cm] {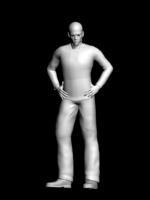}&%
\includegraphics[height=2.2cm, clip, trim=0cm 0cm 0cm 0cm] {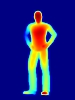}&%
\begin{tabular}[b]{c@{}c}
\includegraphics[height=1.02cm, clip, trim=0cm 0cm 0cm 0cm] {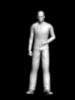}&%
\includegraphics[height=1.02cm, clip, trim=0cm 0cm 0cm 0cm] {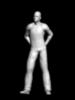}\\%
\includegraphics[height=1.02cm, clip, trim=0cm 0cm 0cm 0cm] {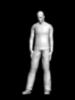}&%
\includegraphics[height=1.02cm, clip, trim=0cm 0cm 0cm 0cm] {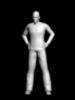}%
\end{tabular}&
\includegraphics[height=2.2cm, clip, trim=0cm 0cm 0cm 0cm] {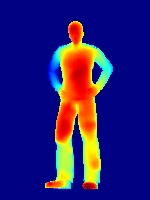}&%

\includegraphics[height=2.2cm, clip, trim=0cm 0cm 0cm 0cm] {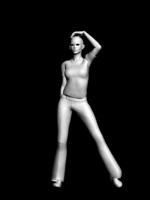}&%
\includegraphics[height=2.2cm, clip, trim=0cm 0cm 0cm 0cm] {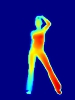}&%
\begin{tabular}[b]{c@{}c}
\includegraphics[height=1.02cm, clip, trim=0cm 0cm 0cm 0cm] {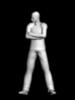}&%
\includegraphics[height=1.02cm, clip, trim=0cm 0cm 0cm 0cm] {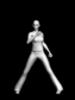}\\%
\includegraphics[height=1.02cm, clip, trim=0cm 0cm 0cm 0cm] {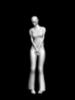}&%
\includegraphics[height=1.02cm, clip, trim=0cm 0cm 0cm 0cm] {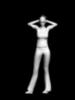}%
\end{tabular}&
\includegraphics[height=2.2cm, clip, trim=0cm 0cm 0cm 0cm] {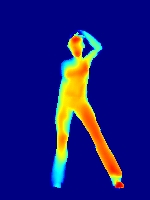}\\%

\end{tabular}
\caption{{\bf Human-posture depth estimates.} Four out of the 45 human-posture depth estimates computed using automatically obtained weights (see Table~\ref{tbl:depth_params} for parameter values). In both columns from left to right: input image, its ground truth, the four {\em automatically} selected database examples used for the reconstructions, and our output estimate.}\label{fig:batch_bodies}
\end{center}
\end{figure*}

\begin{figure*}
\begin{center}
\begin{tabular}[b]{c@{}c@{}c@{}c@{}c}

Input&Ground&DB&Our\\
image&truth&examples&result\\
\includegraphics[height=2.2cm, clip, trim=0cm 0cm 0cm 0cm] {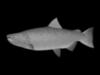}&%
\includegraphics[height=2.2cm, clip, trim=0cm 0cm 0cm 0cm] {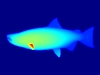}&%
\begin{tabular}[b]{c@{}c}
\includegraphics[height=1cm, clip, trim=0cm 0cm 0cm 0cm] {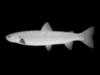}&%
\includegraphics[height=1cm, clip, trim=0cm 0cm 0cm 0cm] {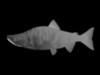}\\%
\includegraphics[height=1cm, clip, trim=0cm 0cm 0cm 0cm] {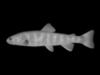}&%
\includegraphics[height=1cm, clip, trim=0cm 0cm 0cm 0cm] {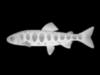}%
\end{tabular}&
\includegraphics[height=2.2cm, clip, trim=0cm 0cm 0cm 0cm] {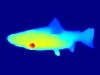}\\%

\includegraphics[height=2.2cm, clip, trim=0cm 0cm 0cm 0cm] {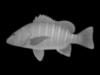}&%
\includegraphics[height=2.2cm, clip, trim=0cm 0cm 0cm 0cm] {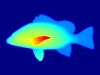}&%
\begin{tabular}[b]{c@{}c}
\includegraphics[height=1cm, clip, trim=0cm 0cm 0cm 0cm] {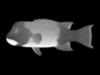}&%
\includegraphics[height=1cm, clip, trim=0cm 0cm 0cm 0cm] {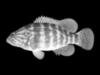}\\%
\includegraphics[height=1cm, clip, trim=0cm 0cm 0cm 0cm] {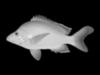}&%
\includegraphics[height=1cm, clip, trim=0cm 0cm 0cm 0cm] {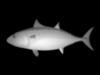}%
\end{tabular}&
\includegraphics[height=2.2cm, clip, trim=0cm 0cm 0cm 0cm] {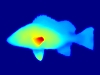}\\%

\includegraphics[height=2.2cm, clip, trim=0cm 0cm 0cm 0cm] {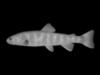}&%
\includegraphics[height=2.2cm, clip, trim=0cm 0cm 0cm 0cm] {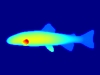}&%
\begin{tabular}[b]{c@{}c}
\includegraphics[height=1cm, clip, trim=0cm 0cm 0cm 0cm] {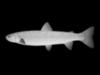}&%
\includegraphics[height=1cm, clip, trim=0cm 0cm 0cm 0cm] {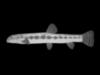}\\%
\includegraphics[height=1cm, clip, trim=0cm 0cm 0cm 0cm] {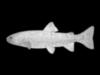}&%
\includegraphics[height=1cm, clip, trim=0cm 0cm 0cm 0cm] {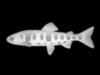}%
\end{tabular}&
\includegraphics[height=2.2cm, clip, trim=0cm 0cm 0cm 0cm] {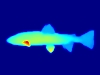}\\%

\includegraphics[height=2.2cm, clip, trim=0cm 0cm 0cm 0cm] {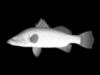}&%
\includegraphics[height=2.2cm, clip, trim=0cm 0cm 0cm 0cm] {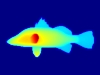}&%
\begin{tabular}[b]{c@{}c}
\includegraphics[height=1cm, clip, trim=0cm 0cm 0cm 0cm] {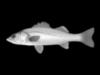}&%
\includegraphics[height=1cm, clip, trim=0cm 0cm 0cm 0cm] {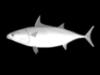}\\%
\includegraphics[height=1cm, clip, trim=0cm 0cm 0cm 0cm] {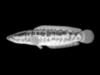}&%
\includegraphics[height=1cm, clip, trim=0cm 0cm 0cm 0cm] {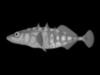}%
\end{tabular}&
\includegraphics[height=2.2cm, clip, trim=0cm 0cm 0cm 0cm] {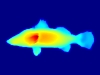}\\%

\end{tabular}
\caption{{\bf Fish depth estimates.} Four out of the 41 fish depth estimates computed using automatically obtained weights (see Table~\ref{tbl:depth_params} for parameter values). From left to right: input image, its ground truth, four of the five {\em automatically} selected database examples used for the reconstructions, and our output estimate.}\label{fig:batch_fish}
\end{center}
\end{figure*}

Fig.~\ref{fig:batch_hands},~\ref{fig:batch_bodies}, and~\ref{fig:batch_fish} present depth estimates obtained by own own method in these batch tests. The wide standard deviation of both the baseline and our method, as reported in Table~\ref{tbl:depth_res}, suggests that these three data sets do not fully capture the range of shapes and appearances of objects in the classes; some objects do not have sufficiently similar counterparts in the database and consequently their estimates (obtained with our method as well as the baseline) were poor. This is not surprising considering the nature of the objects included in these sets (i.e., non-rigid and textured objects). Paired t-tests comparing our algorithm to the baseline method show the improved performance of our method to be significant for all three data sets, with $p<10^9$ for the human-postures, $p<10^4$ for the hands and $p<10^2$ for the fish data sets.\\

\begin{figure*}[ht]
\begin{center}
\begin{tabular}[b]{c@{}c@{}c@{}c@{}c}
\includegraphics[height=1.8cm, clip, trim=16cm 13cm 14.3cm 11cm] {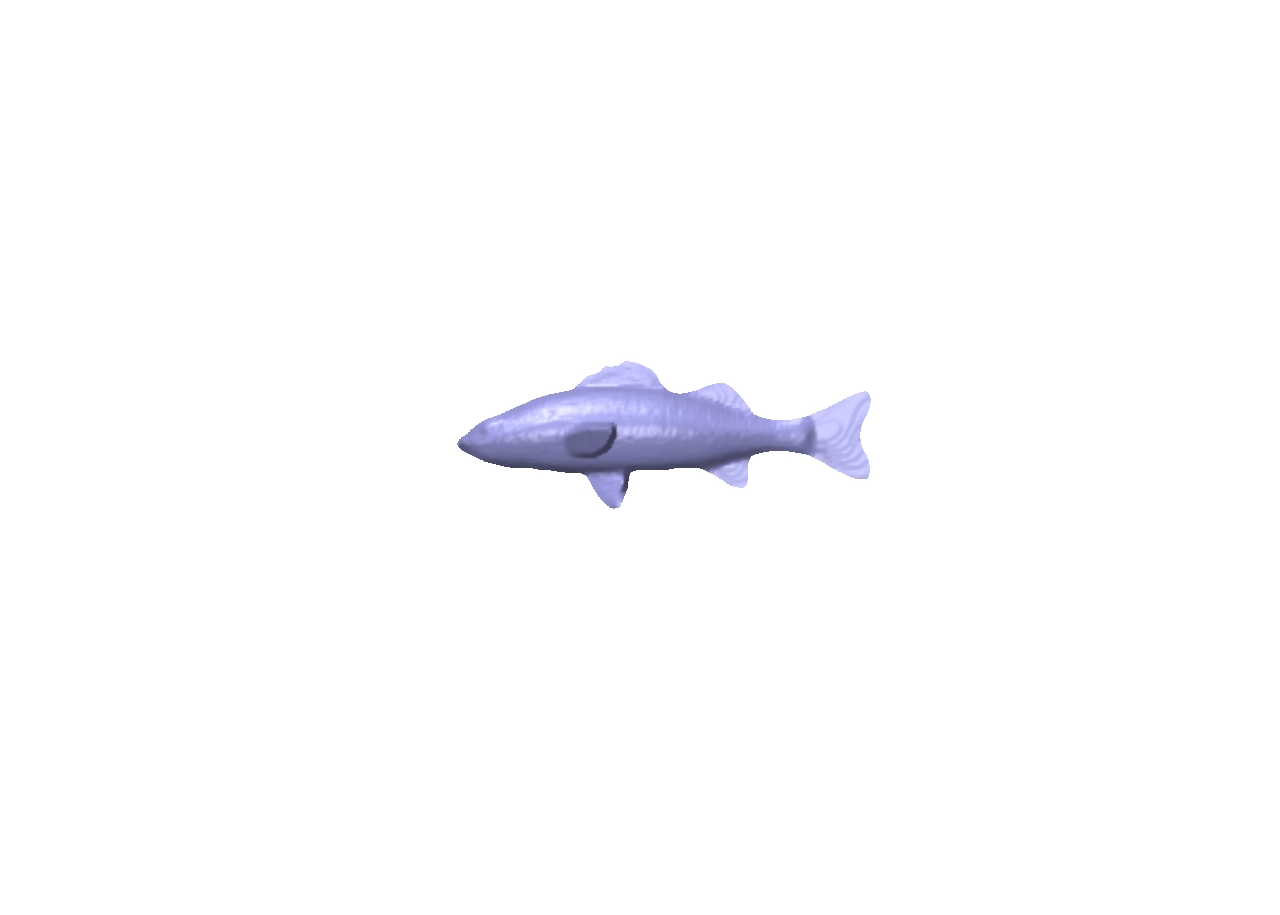}&%
%
\begin{tabular}[b]{c}
\includegraphics[height=0.9cm, clip, trim=0cm 1cm 0cm 1cm] {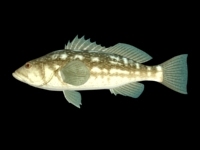}\\%
\includegraphics[height=0.9cm, clip, trim=0cm 1cm 0cm 1cm] {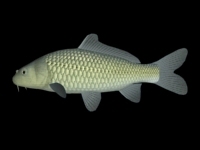}%
\end{tabular}&
%
%

\includegraphics[height=1.8cm, clip, trim=16cm 13cm 14.3cm 11cm] {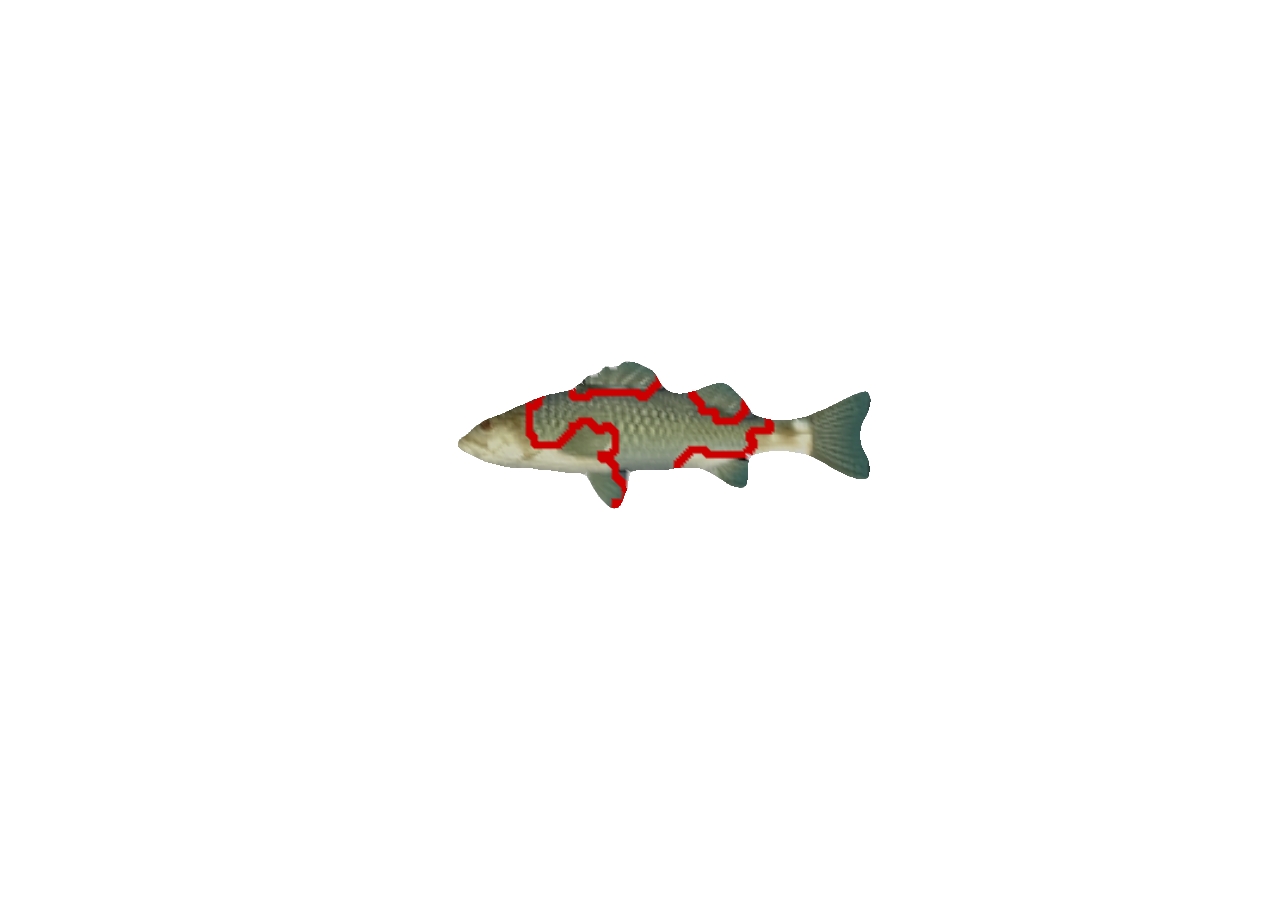}&
%
\begin{tabular}[b]{c}
\includegraphics[height=0.62cm, clip, trim=16cm 14.5cm 14.4cm 13.5cm] {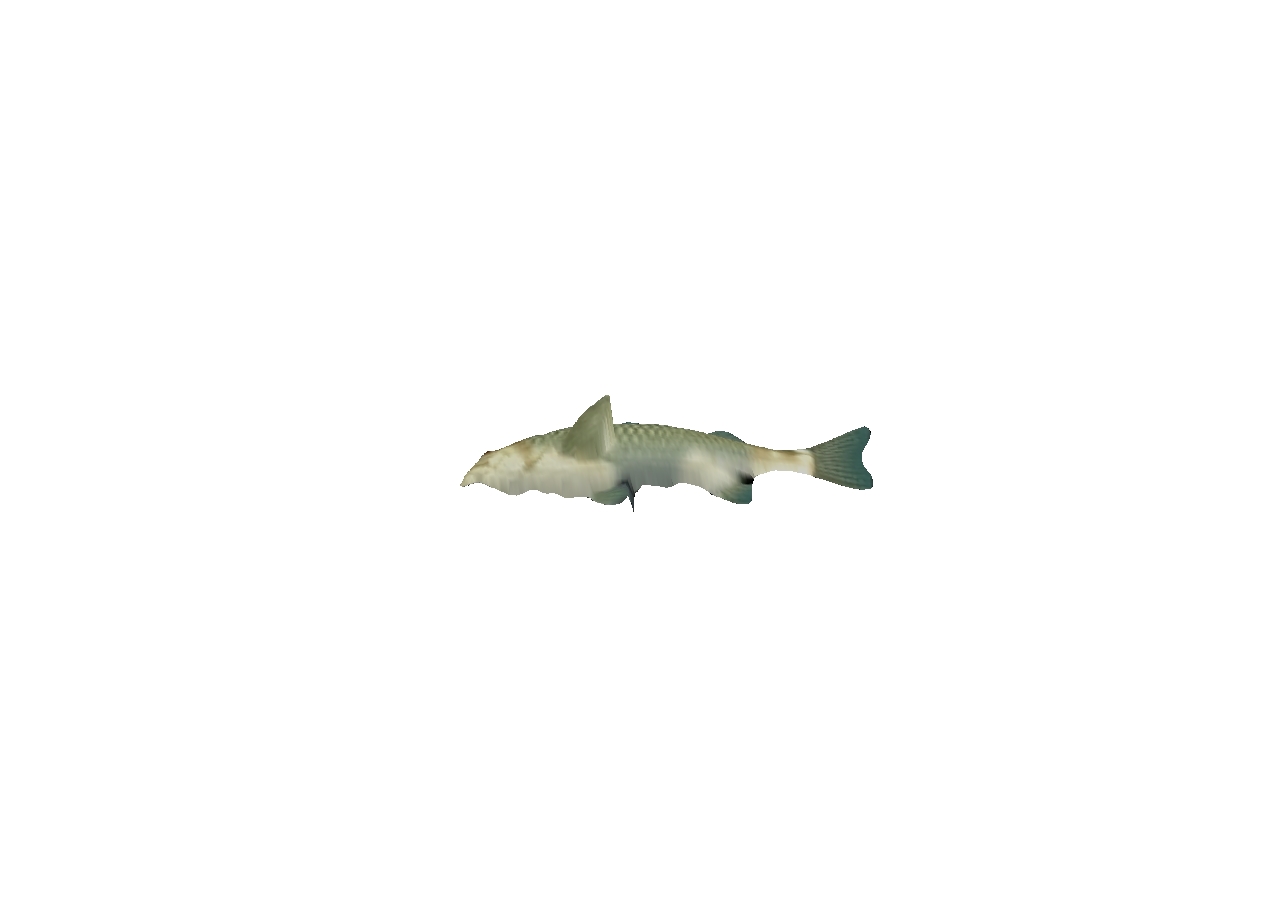}\\%
\includegraphics[height=0.62cm, clip, trim=16cm 14.5cm 14.4cm 13.5cm] {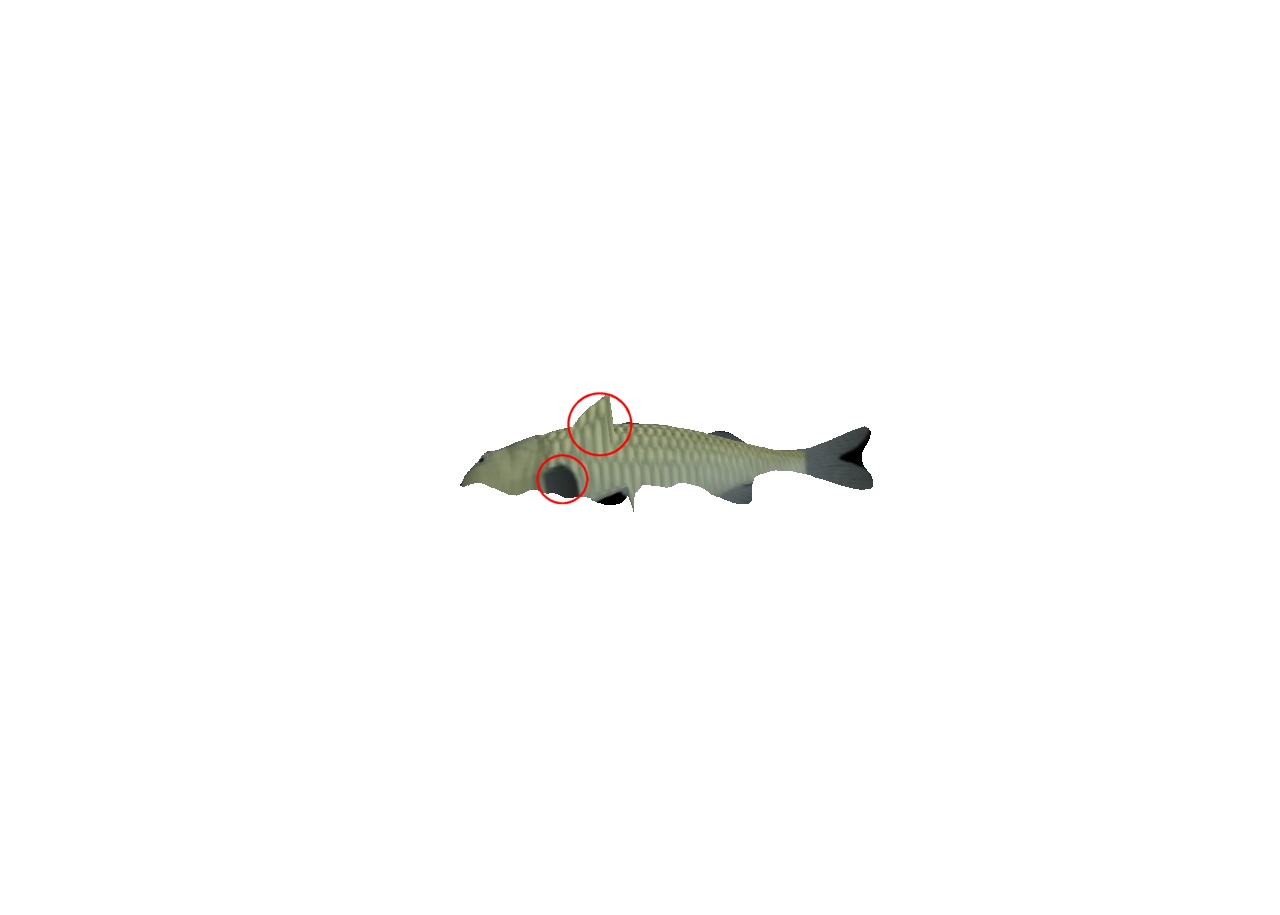}\\%
\includegraphics[height=0.62cm, clip, trim=16cm 14.5cm 14.4cm 13.5cm] {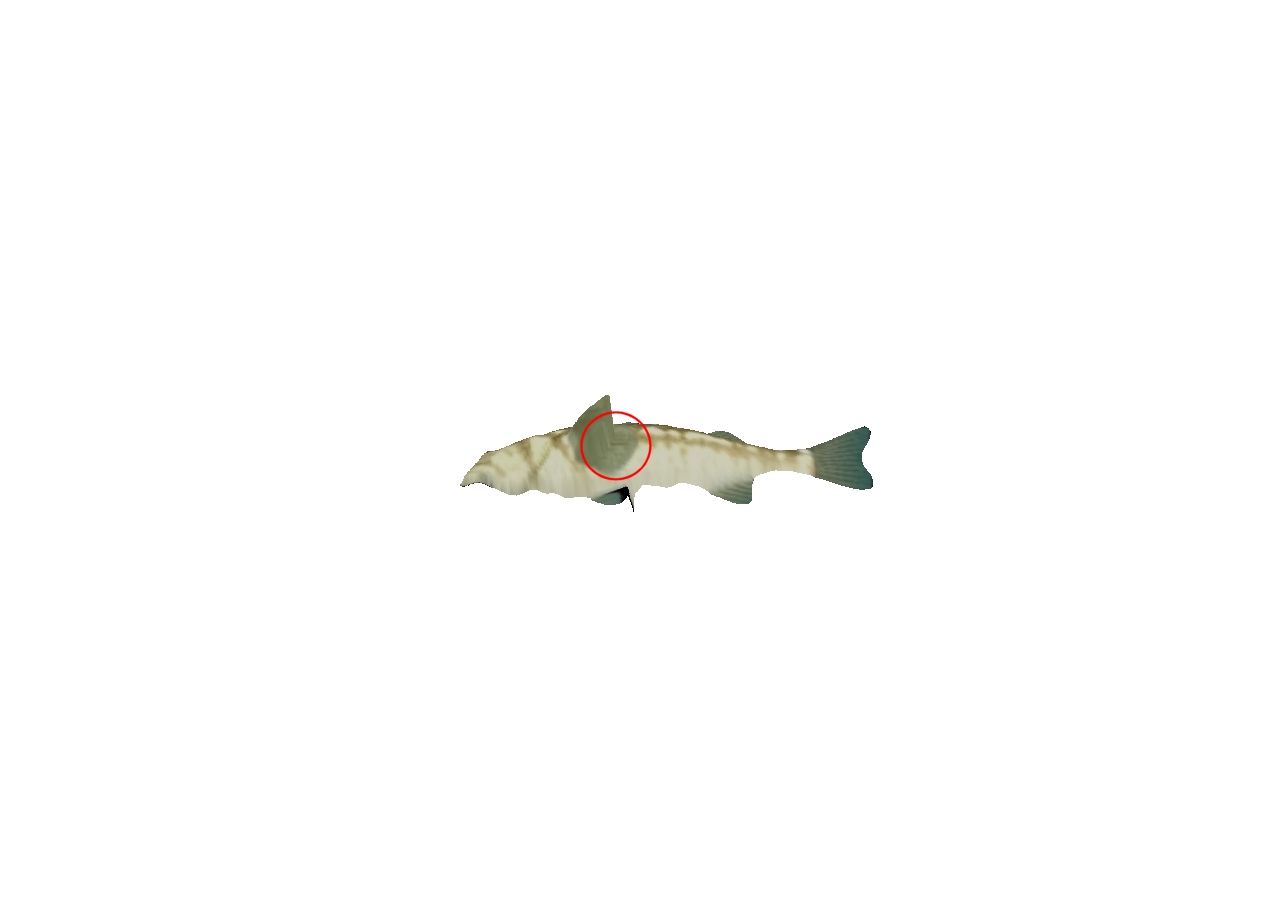}%
\end{tabular}
&
%
%
\includegraphics[height=1.8cm, clip, trim=9.1cm 7cm 8cm 7cm] {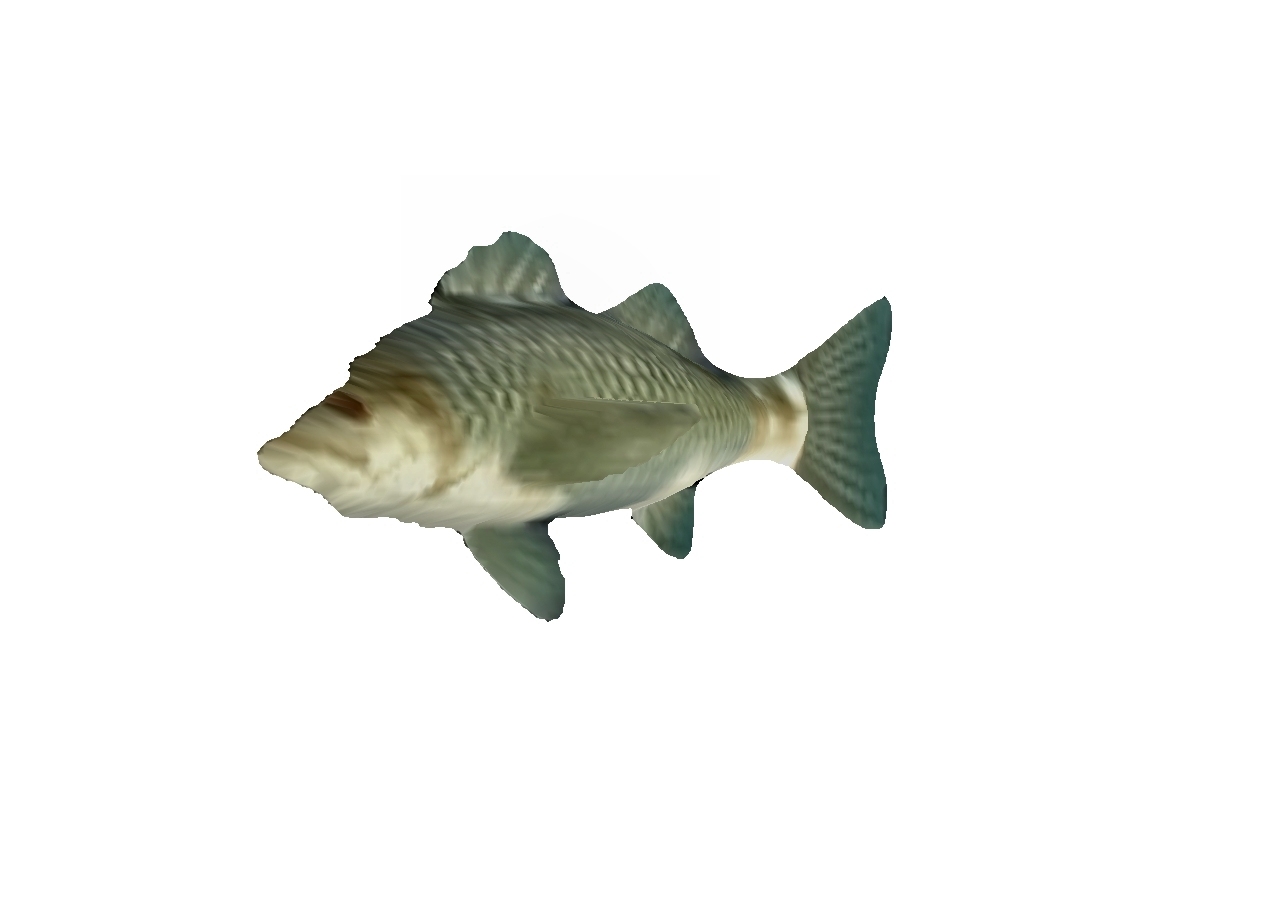}%
\\
(a)&(b)&(c)&(d)&(e)
\end{tabular}
\end{center}

\caption{{\bf Fish image-maps.} (a) Input depth-map. (b)
Automatically selected database objects (image-maps displayed). (c)
Output image marked with the areas taken from each database image. (d)
Input depth rendered with, from top to bottom, result image and database
image-maps. Note the mismatching features when using the database
images. (e) Textured 3D view of our output.}
\label{fig:fish_detailed} 
\end{figure*}

\begin{figure*}[!ht]
\begin{center}
\begin{tabular}[b]{c@{}c@{}c@{}c@{}c@{}c}
\includegraphics[height=1.4cm, clip, trim=16cm 11cm 14.3cm 10cm] {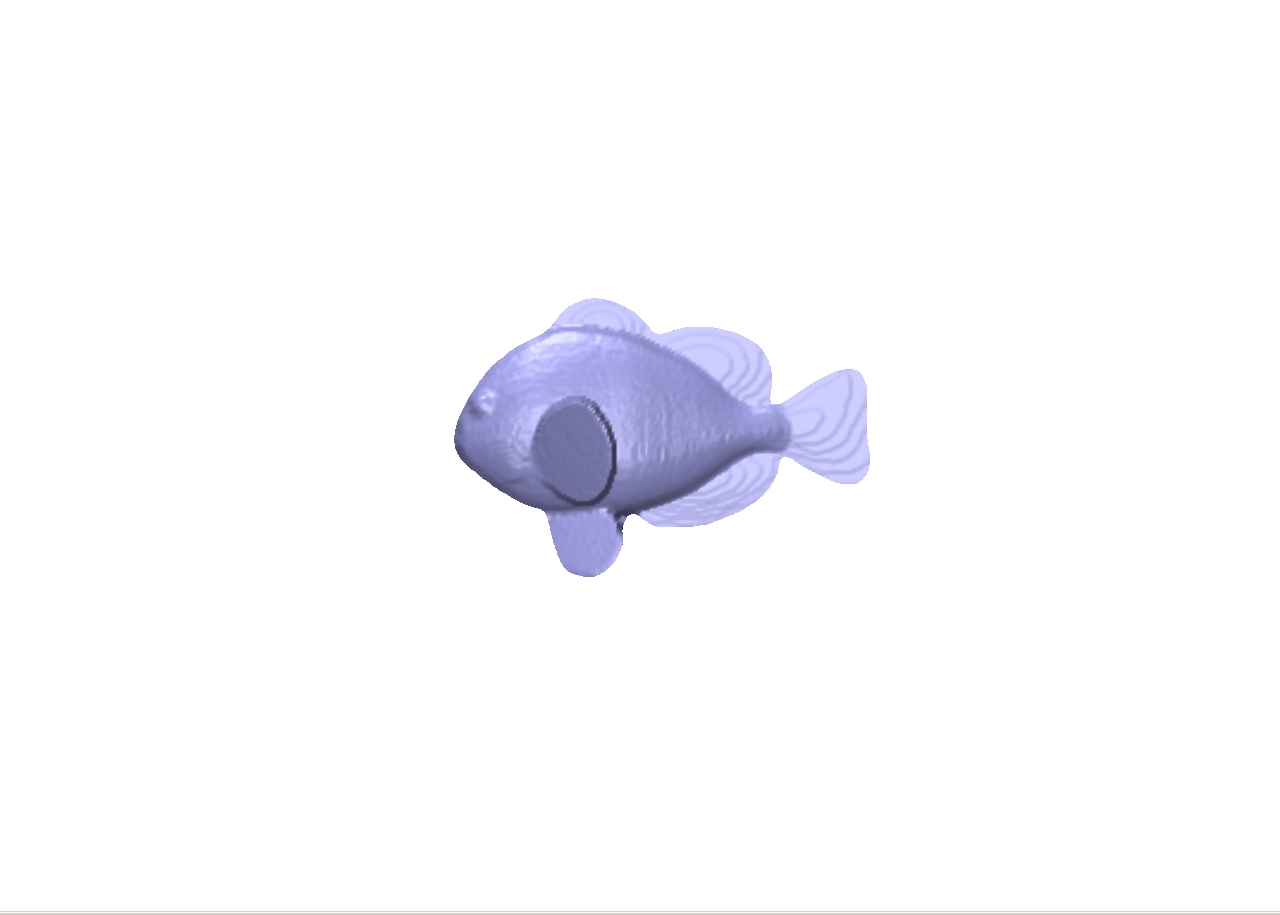}&%
\includegraphics[height=1.4cm, clip, trim=16cm 13cm 14.3cm 11cm] {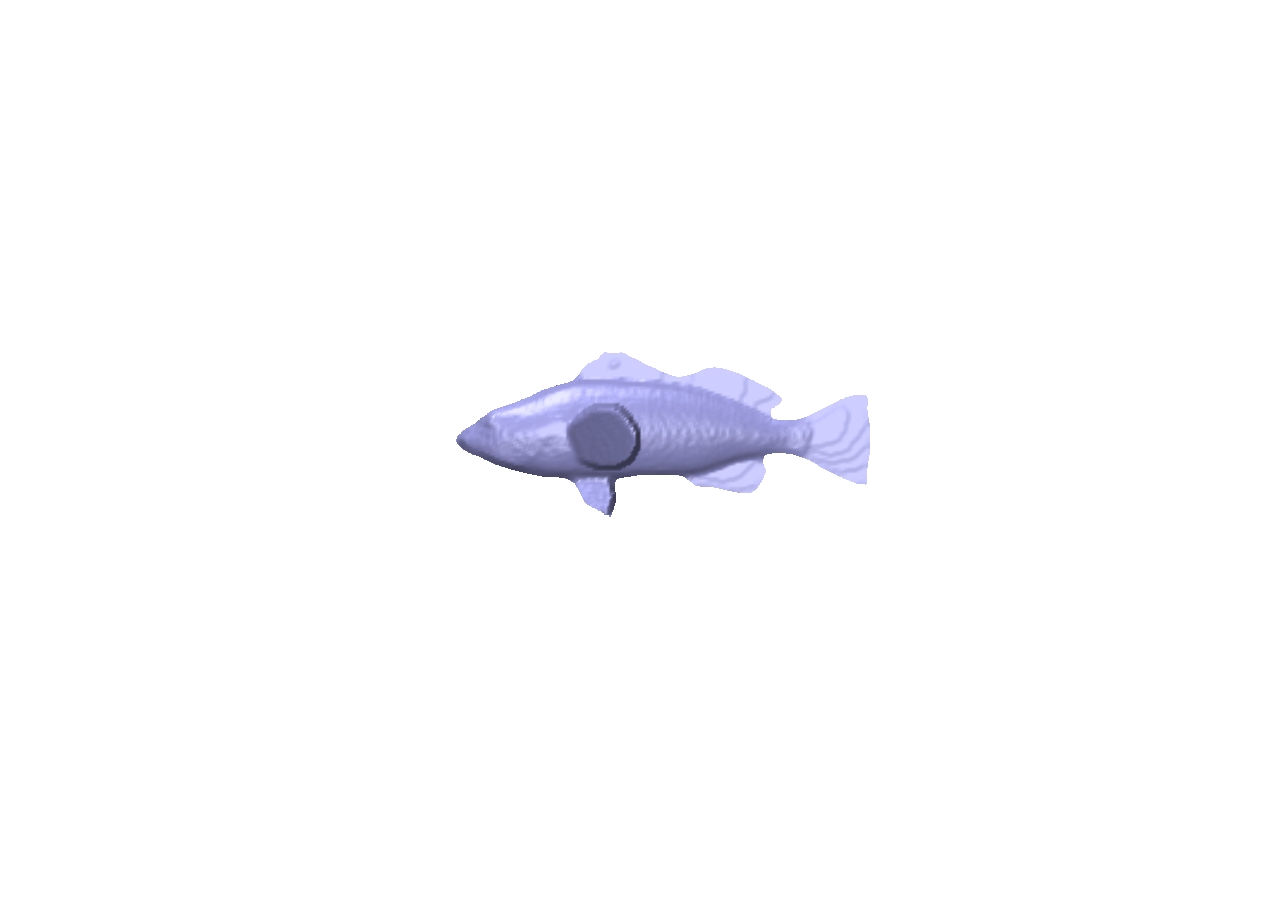}&%
\includegraphics[height=1.4cm, clip, trim=16cm 13cm 14.3cm 11cm] {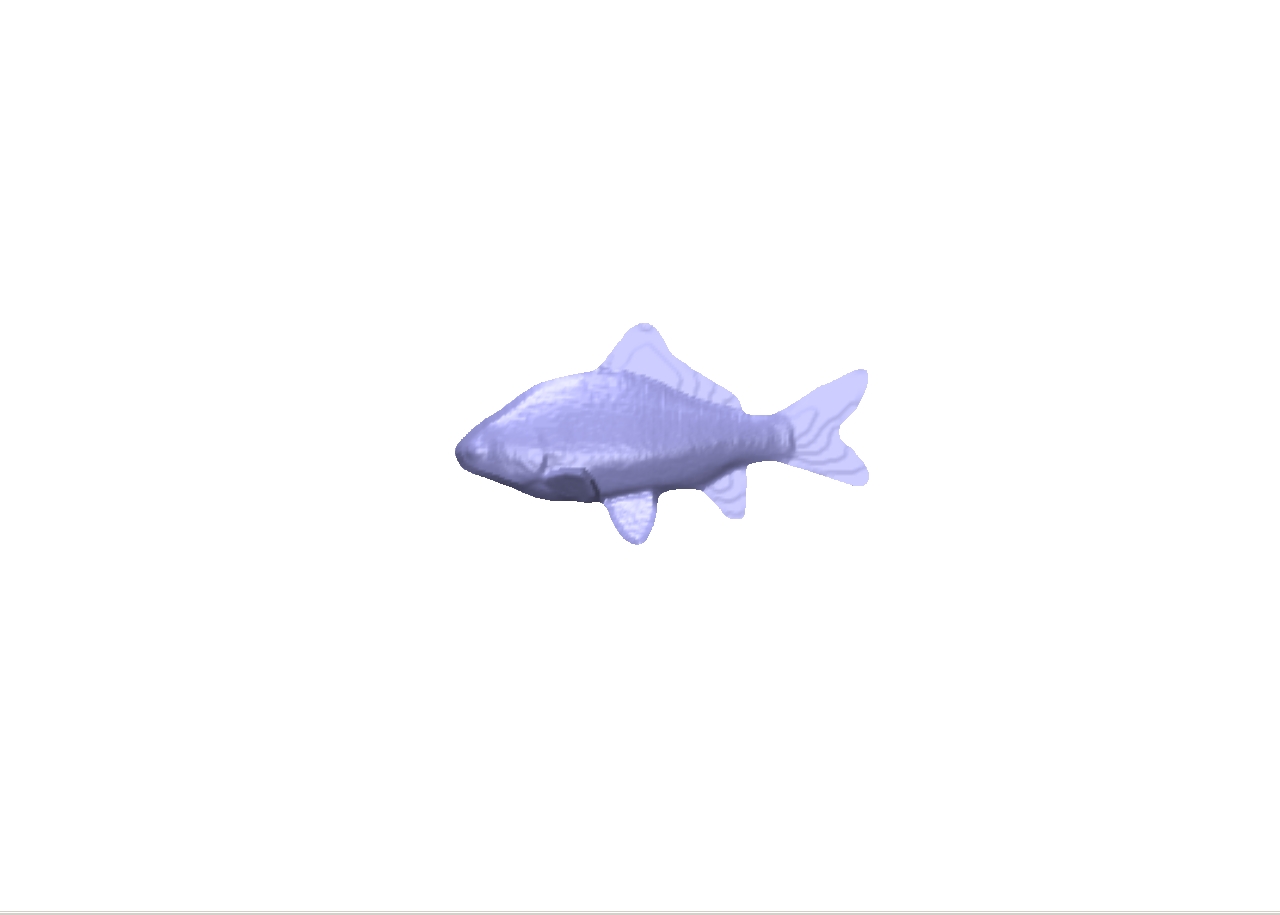}&%
\includegraphics[height=1.4cm, clip, trim=16cm 11cm 14.3cm 10cm] {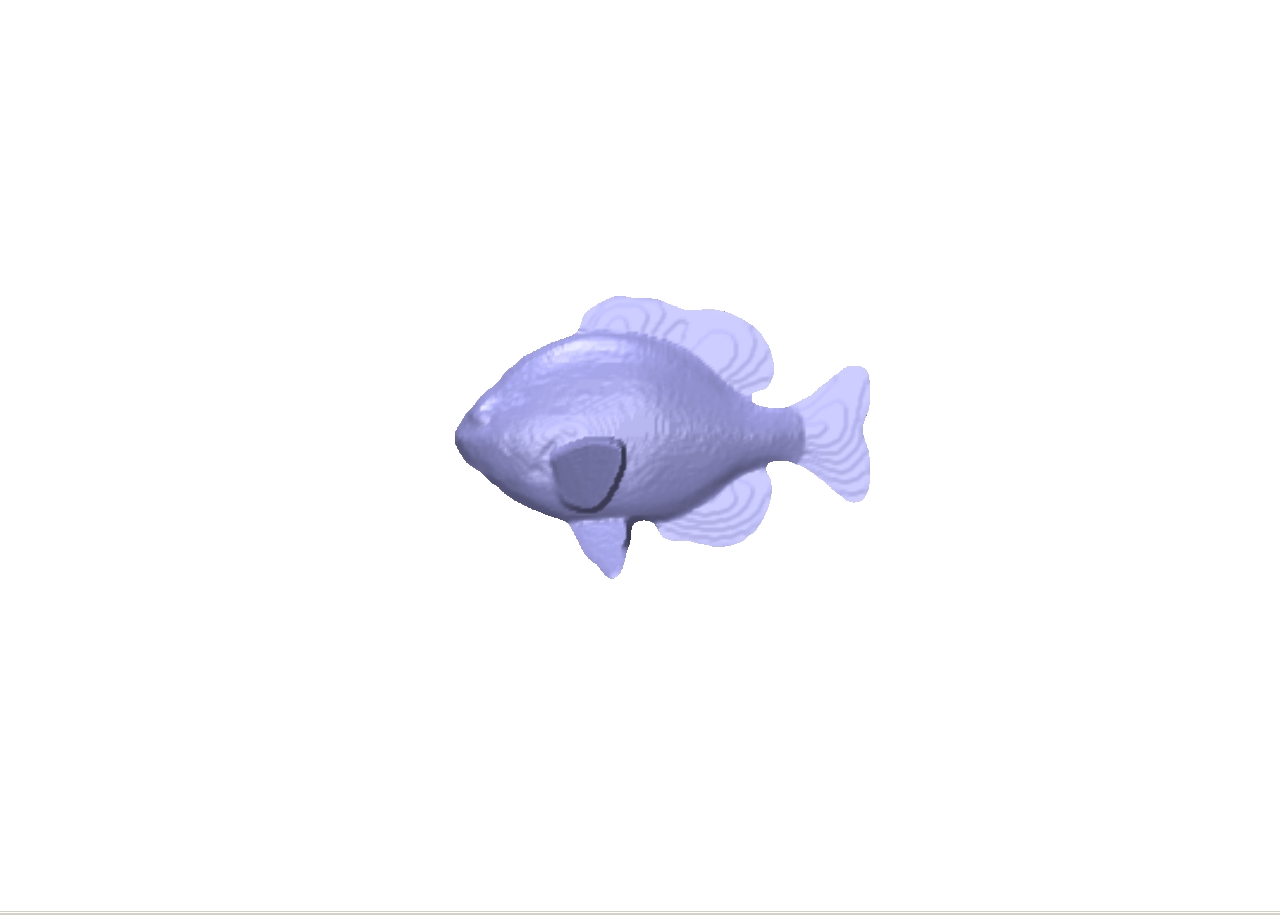}&%
\includegraphics[height=1.4cm, clip, trim=16cm 13cm 14.3cm 11cm] {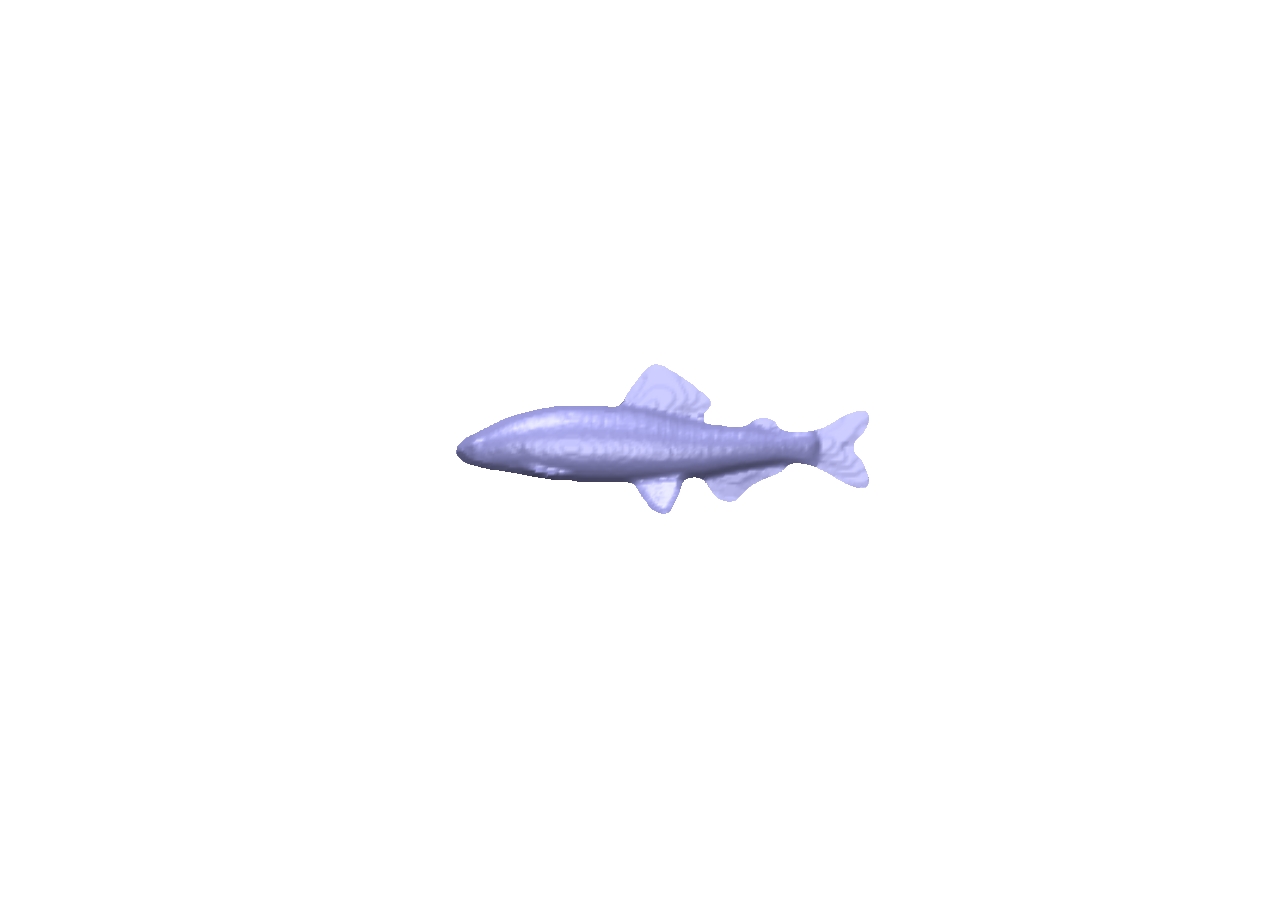}&%
\includegraphics[height=1.4cm, clip, trim=16cm 13cm 14.3cm 11cm] {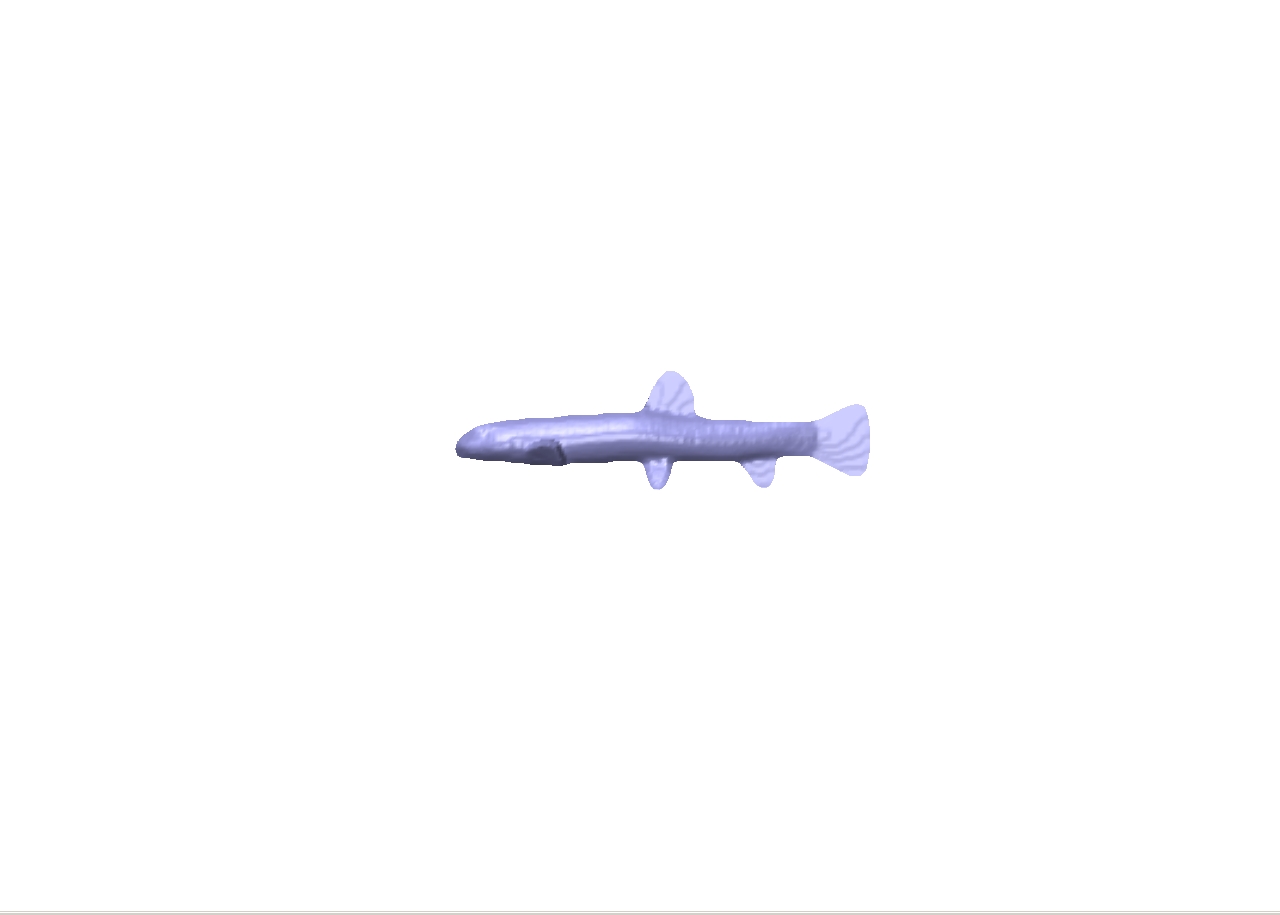}%
\\
\includegraphics[height=1.4cm, clip, trim=16cm 11cm 14.3cm 10cm] {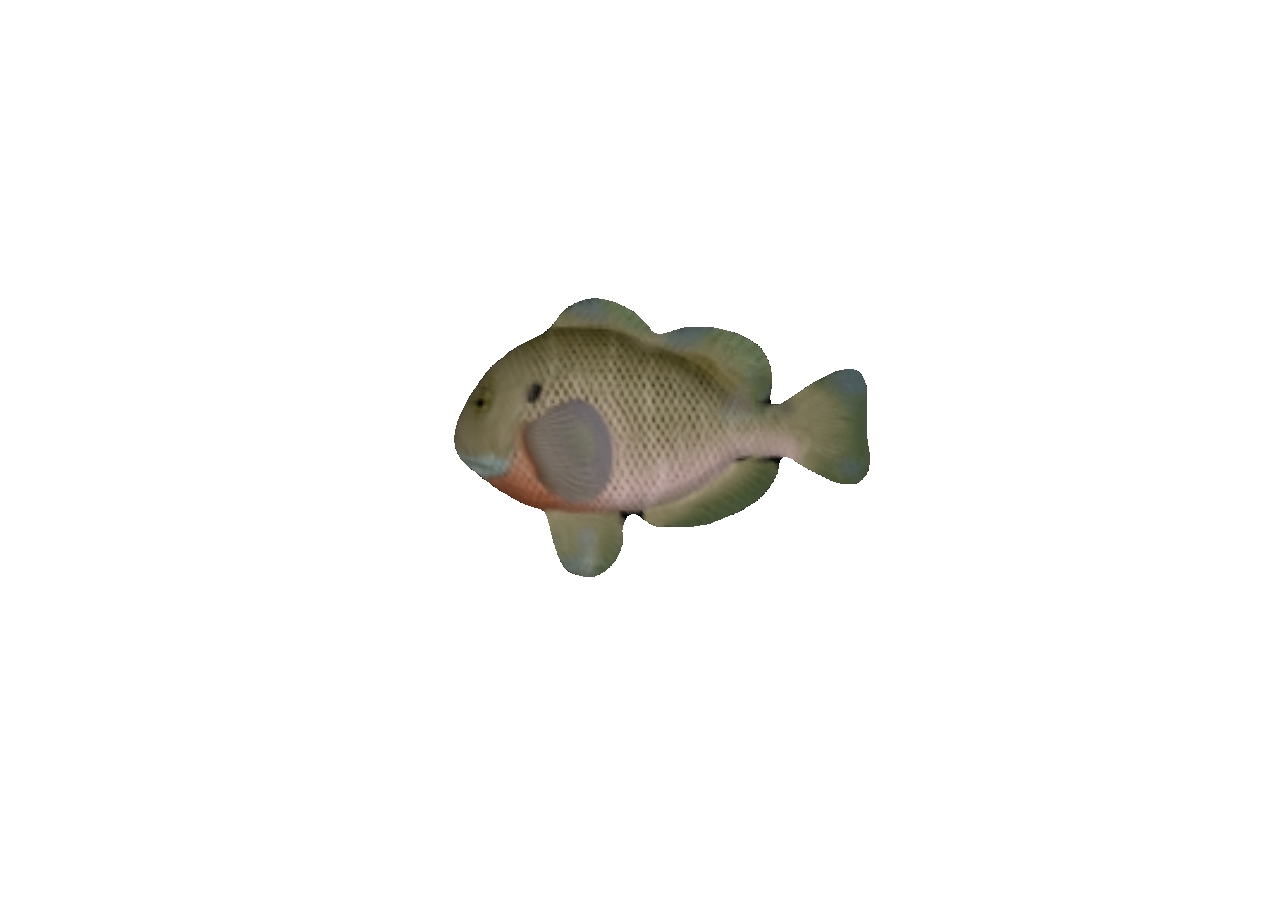}&%
\includegraphics[height=1.4cm, clip, trim=16cm 13cm 14.3cm 11cm] {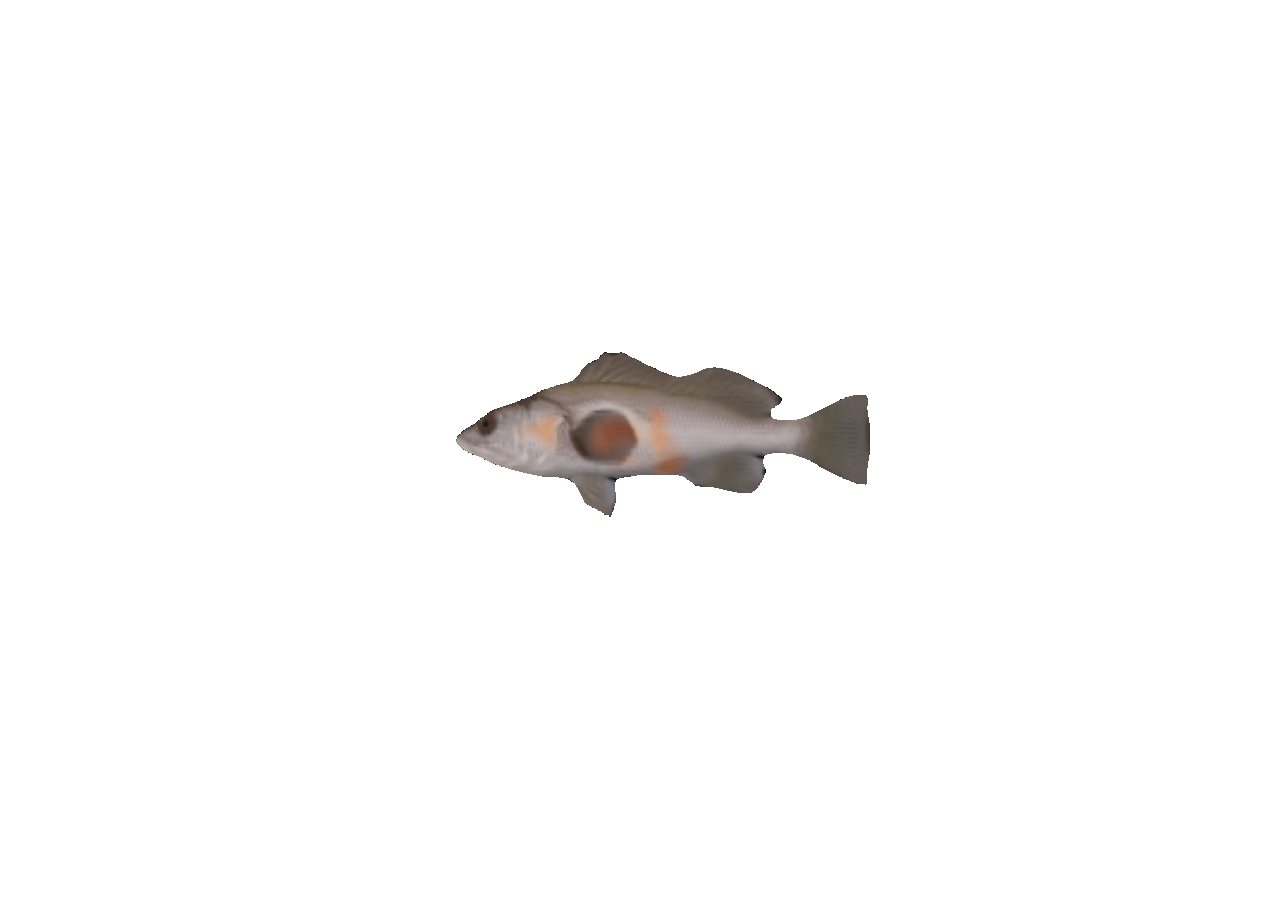}&%
\includegraphics[height=1.4cm, clip, trim=16cm 13cm 14.3cm 11cm] {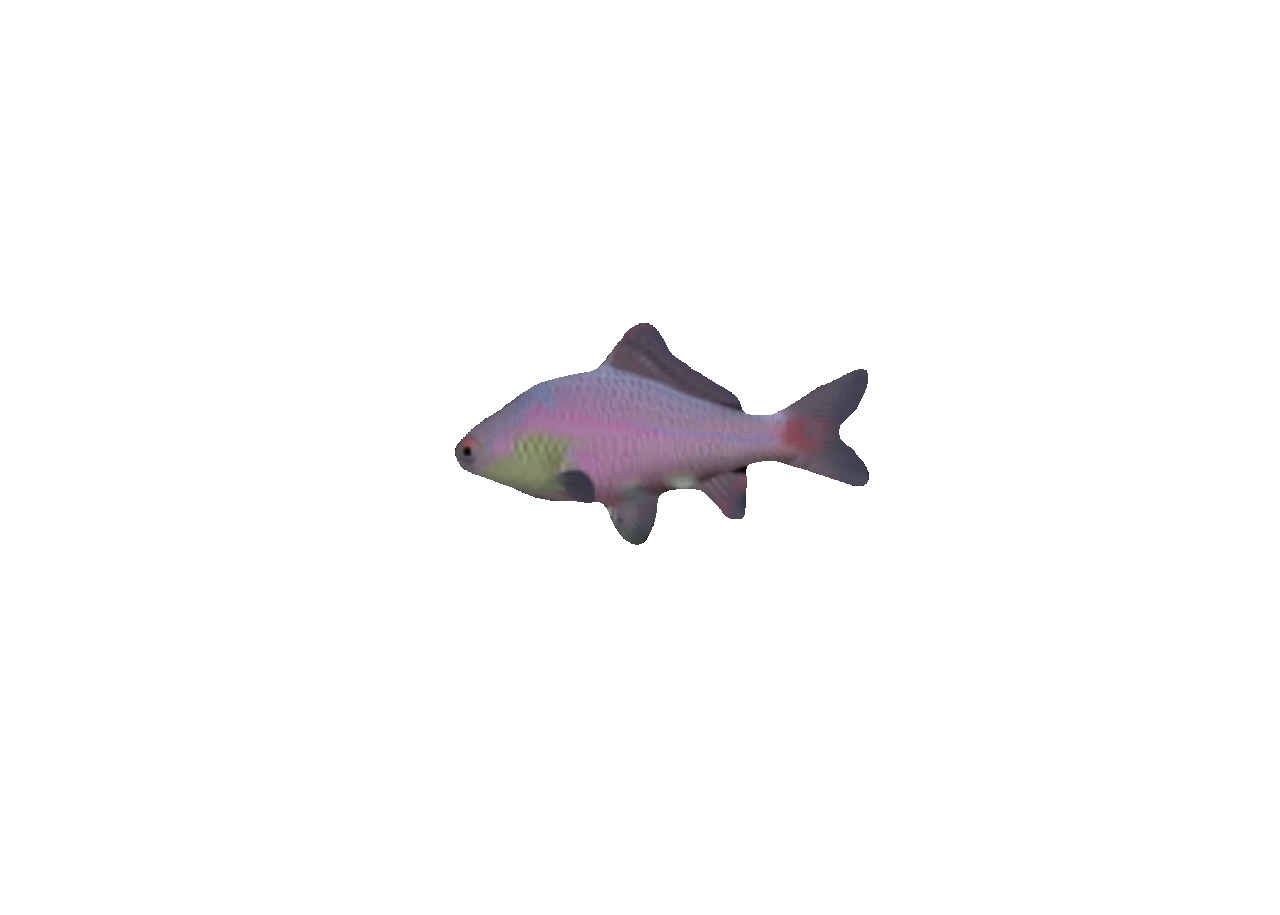}&%
\includegraphics[height=1.4cm, clip, trim=16cm 11cm 14.3cm 10cm] {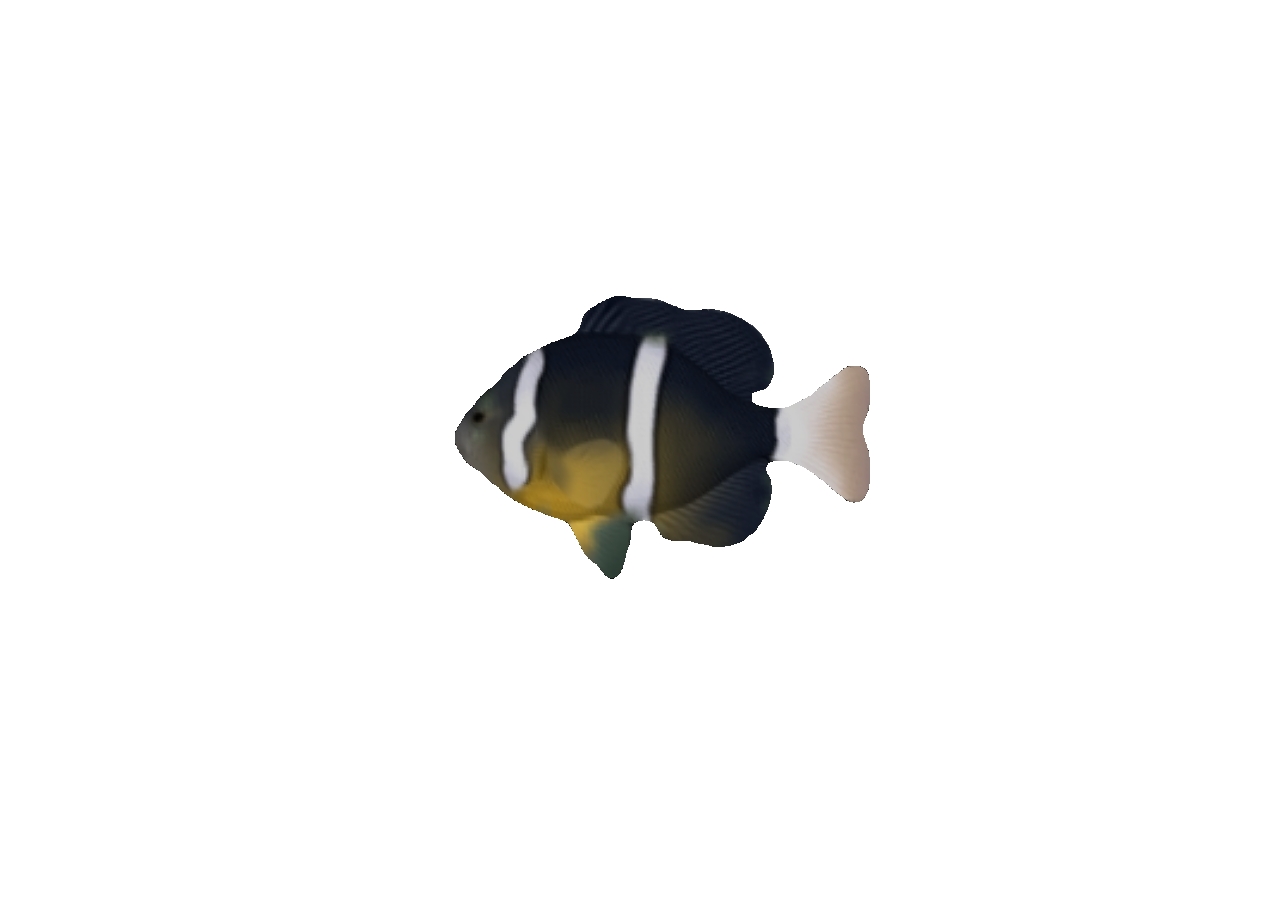}&%
\includegraphics[height=1.4cm, clip, trim=16cm 13cm 14.3cm 11cm] {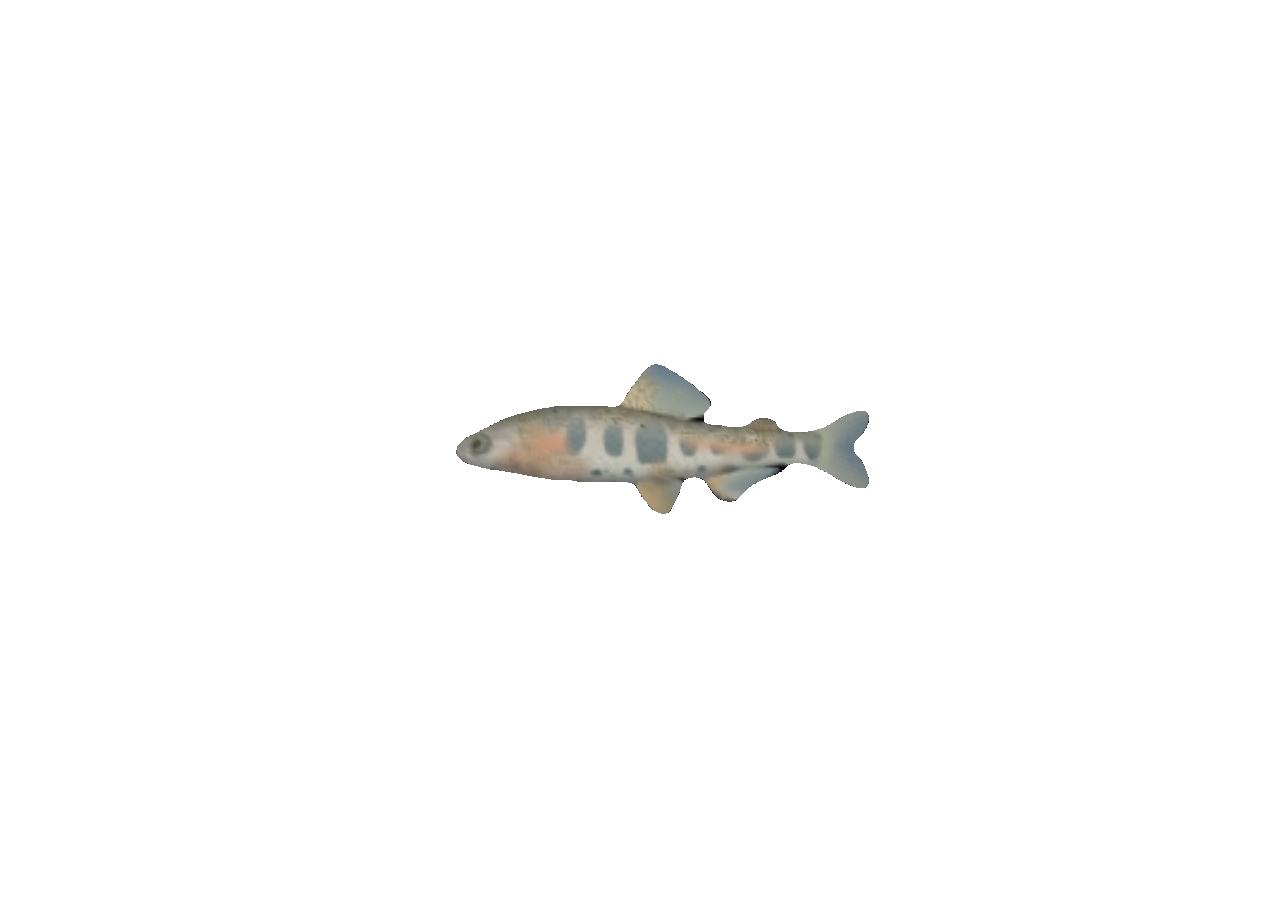}&%
\includegraphics[height=1.4cm, clip, trim=16cm 13cm 14.3cm 11cm] {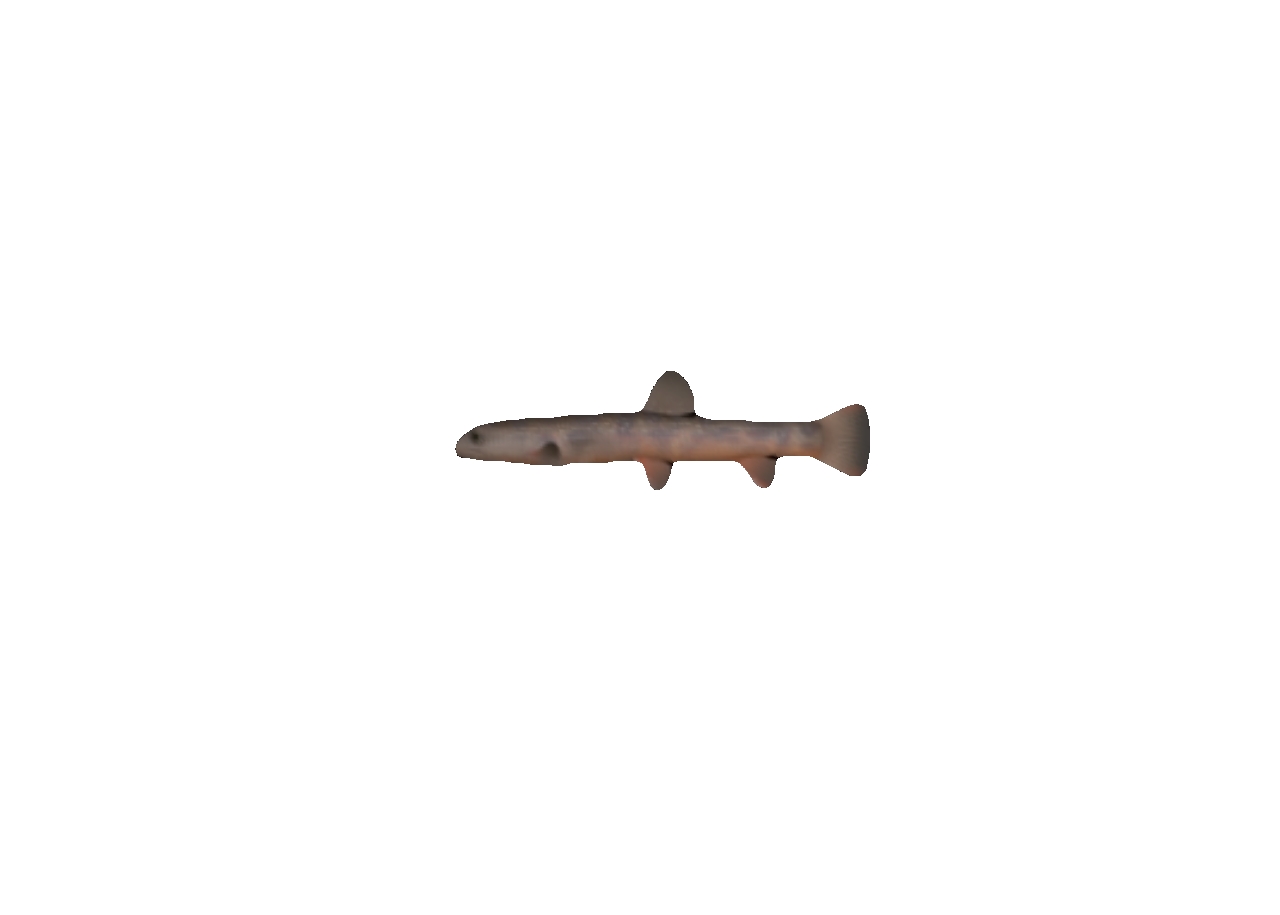}%
\end{tabular}
\end{center}
\caption{{\bf Fish image-maps.} Top row, input depth-maps; bottom
row, our output image-maps.} \label{fig:col_fish} 
\end{figure*}

\begin{figure*}[!th]
\begin{center}
\begin{tabular}[b]{c@{}c@{}cc@{}c@{}cc@{}c@{}c}
\hspace{-0.6cm}
\includegraphics[height=3.0cm, clip, trim=15.9cm 1.8cm 15.1cm 1.5cm] {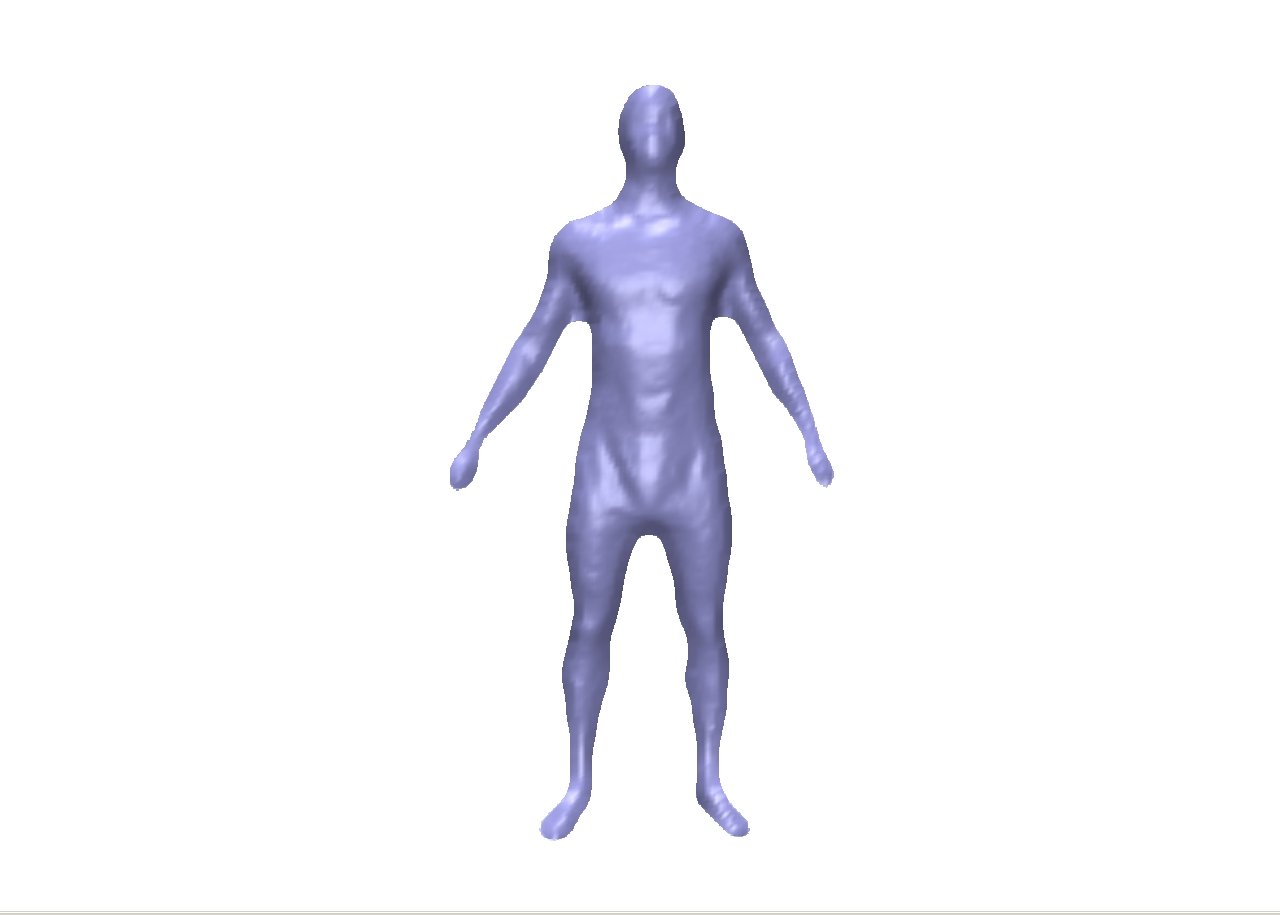}&%
\includegraphics[height=3.0cm, clip, trim=15.9cm 1.8cm 15.1cm 1.5cm] {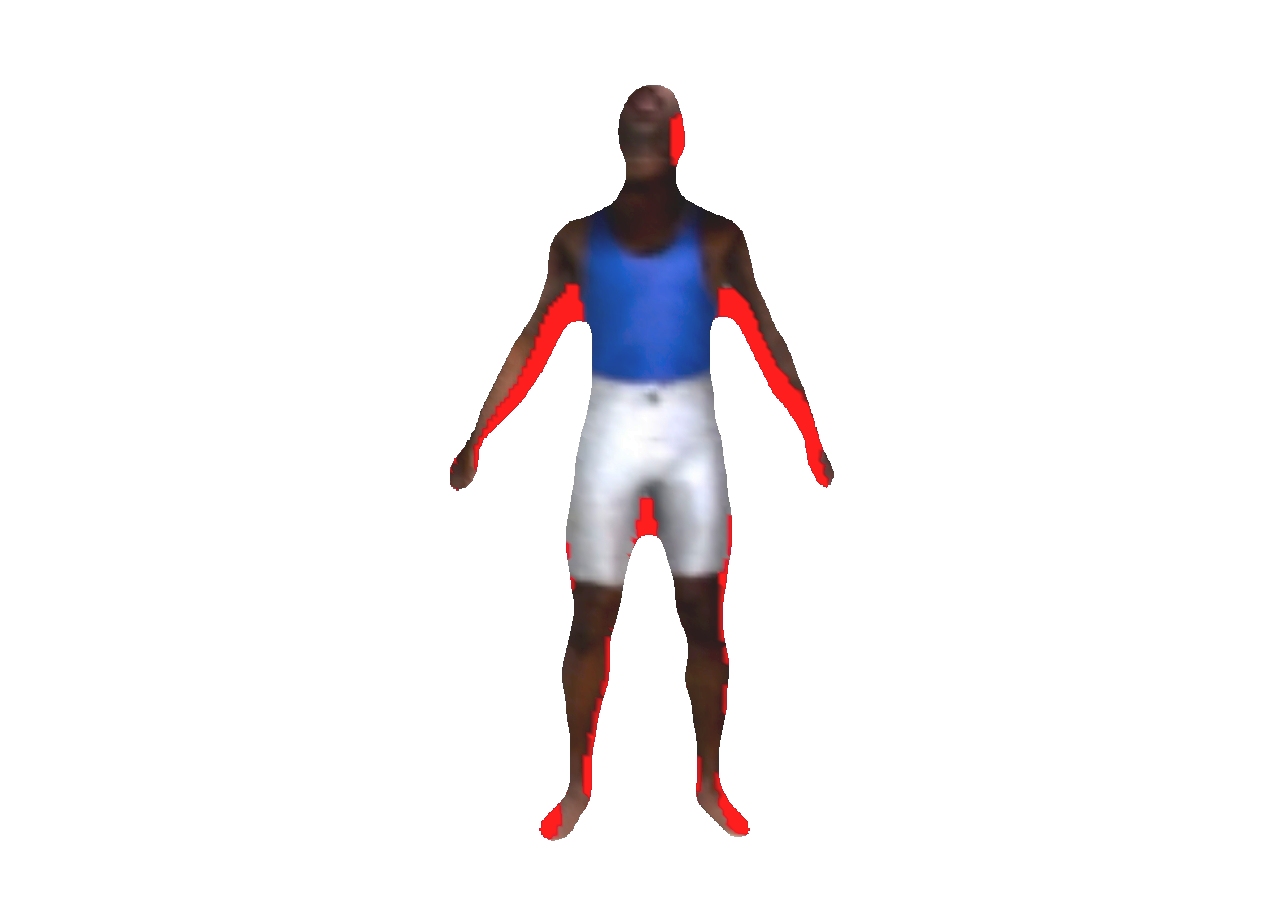}&%
\includegraphics[height=3.0cm, clip, trim=15.9cm 1.8cm 15.1cm 1.5cm] {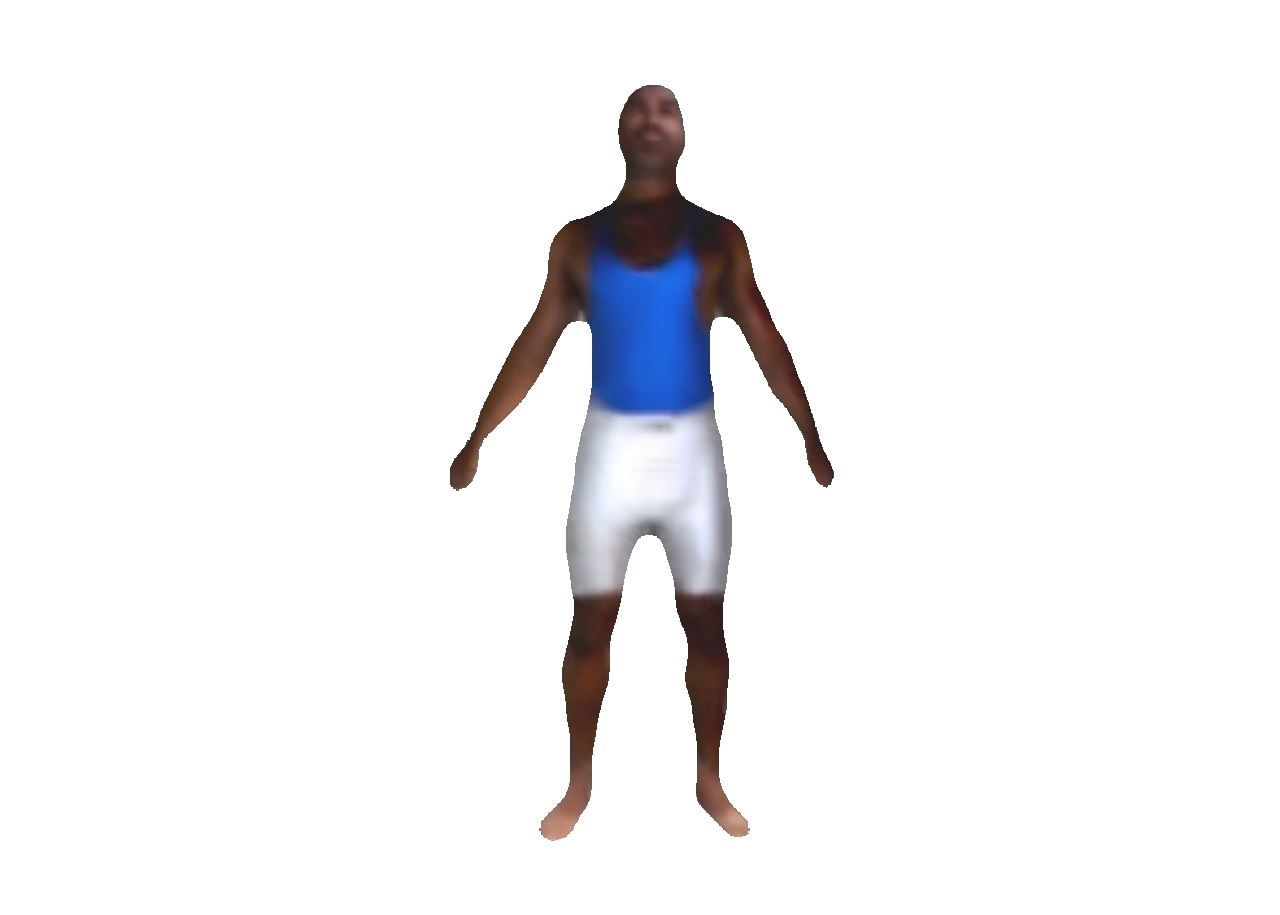}&%
\includegraphics[height=3.0cm, clip, trim=15.7cm 1.8cm 15.1cm 1.3cm] {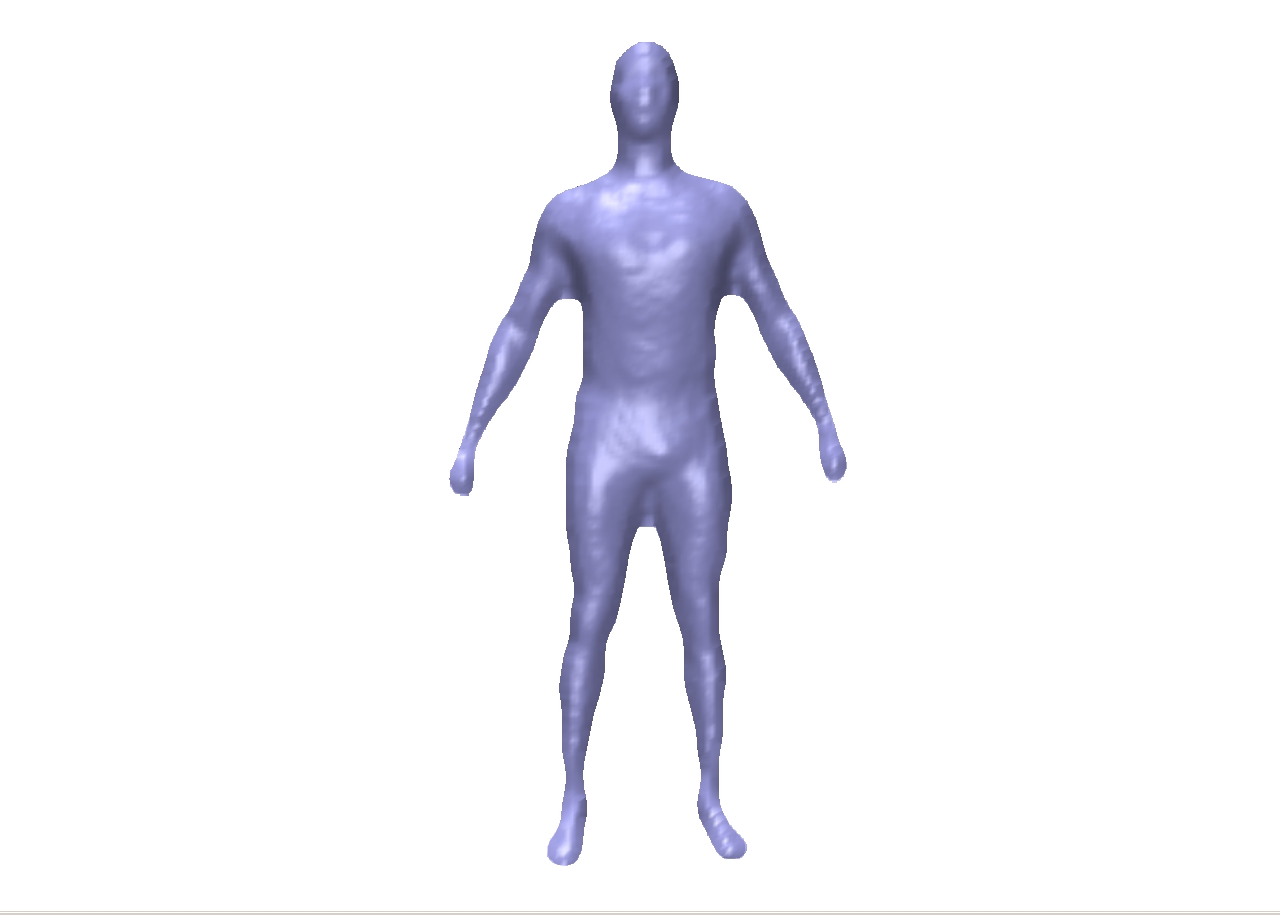}&%
\includegraphics[height=3.0cm, clip, trim=15.7cm 1.8cm 15.1cm 1.3cm] {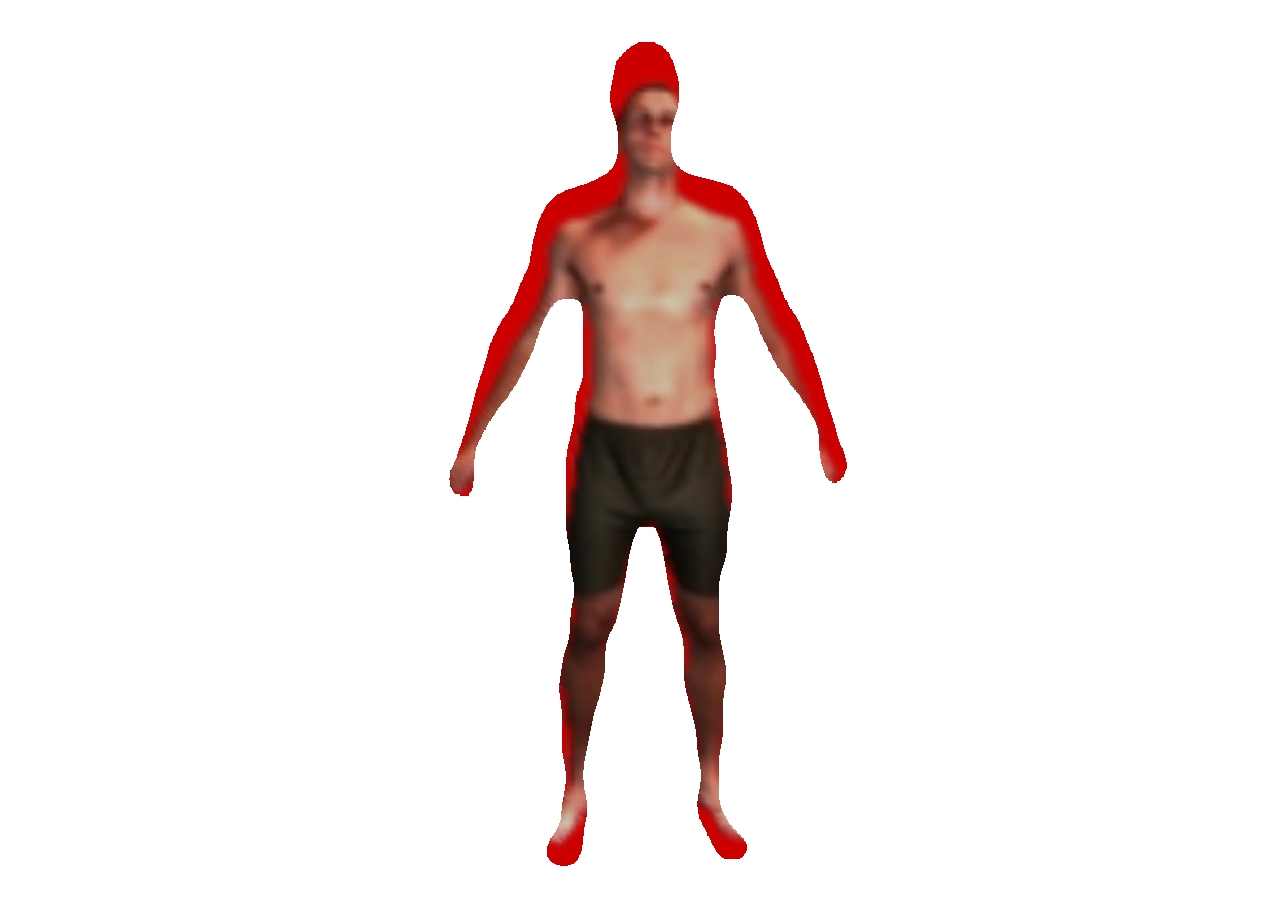}&%
\includegraphics[height=3.0cm, clip, trim=15.7cm 1.8cm 15.1cm 1.3cm] {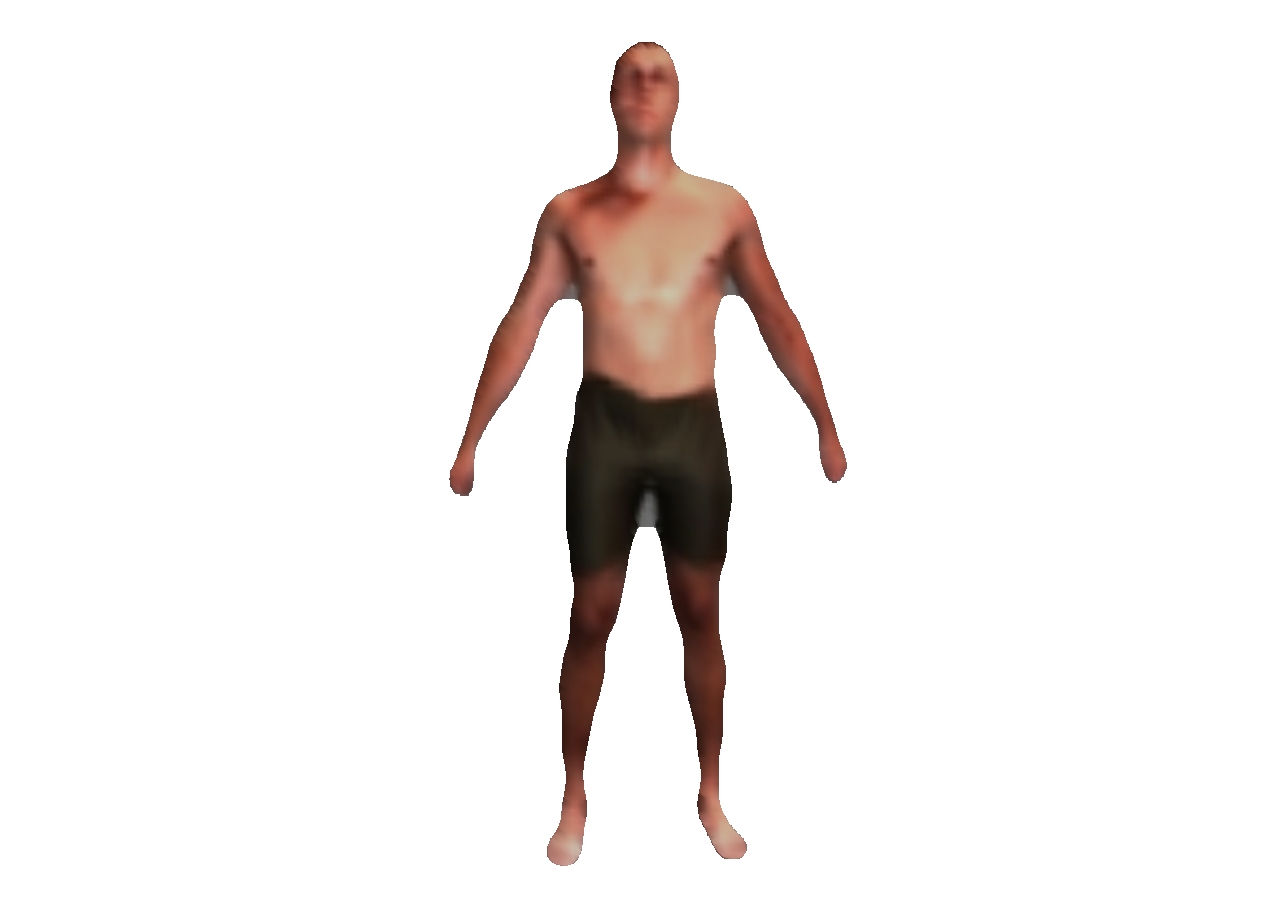}&%
\includegraphics[height=3.0cm, clip, trim=15cm 1.8cm 15cm 1.3cm] {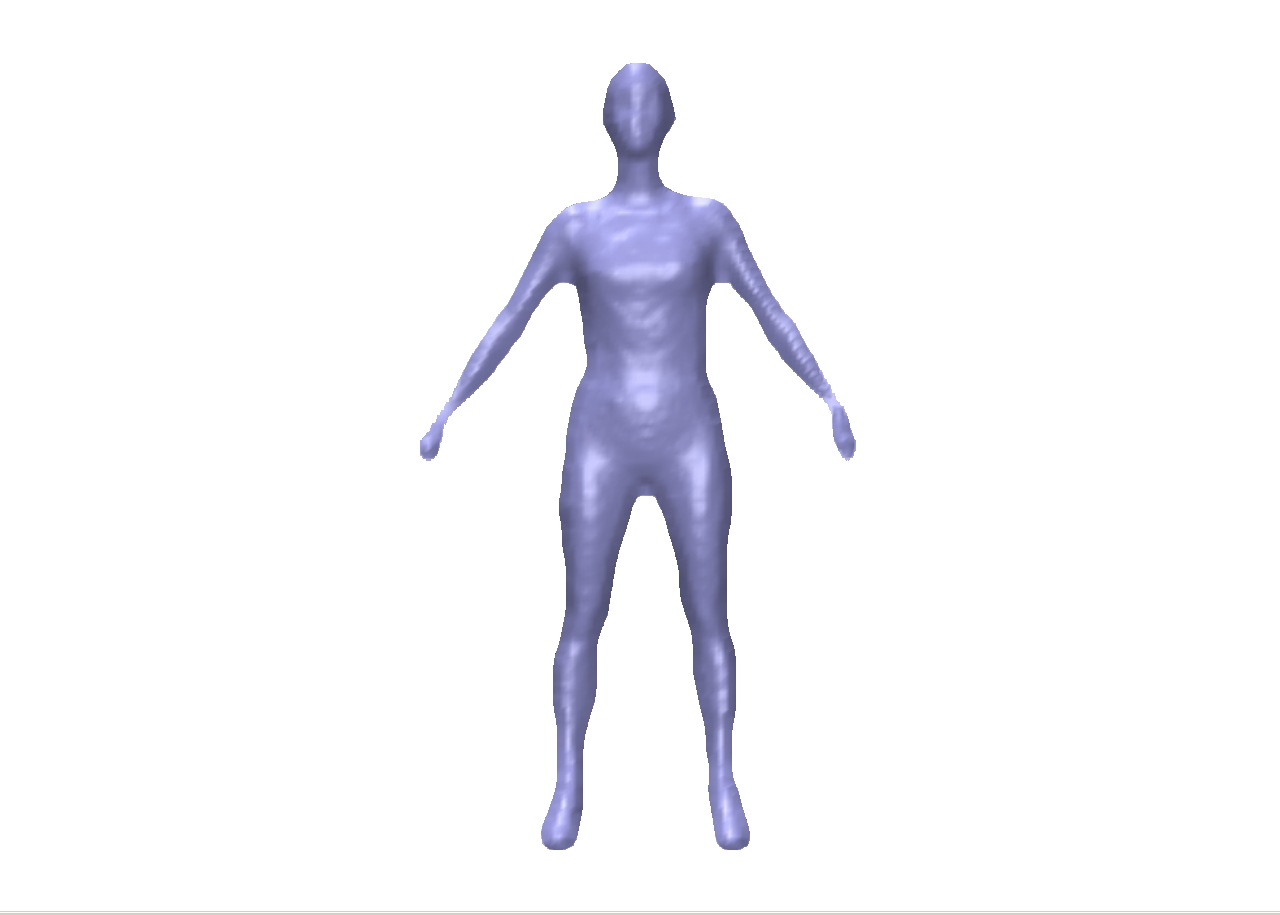}&%
\includegraphics[height=3.0cm, clip, trim=15cm 1.8cm 15cm 1.3cm] {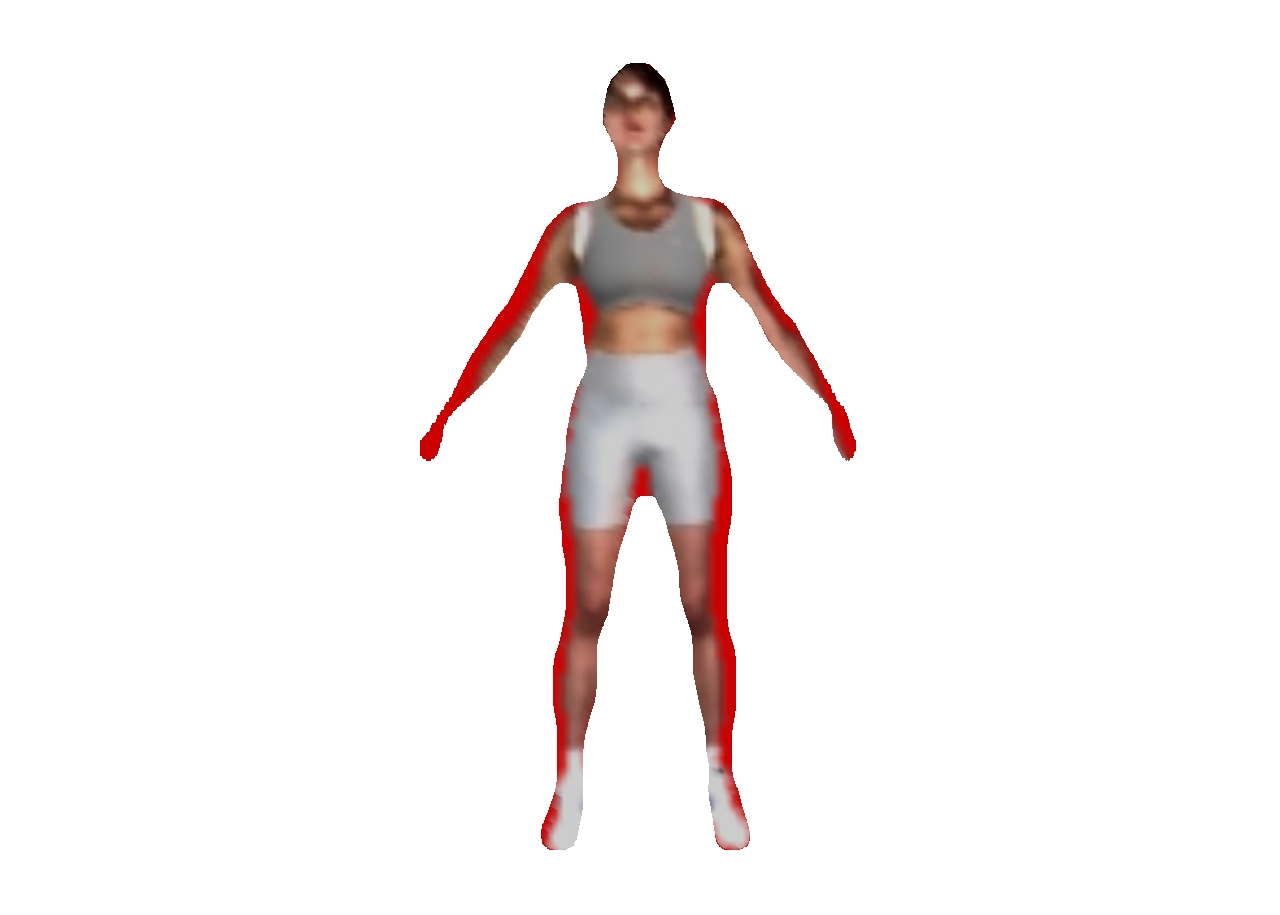}&%
\includegraphics[height=3.0cm, clip, trim=15cm 1.8cm 15cm 1.3cm] {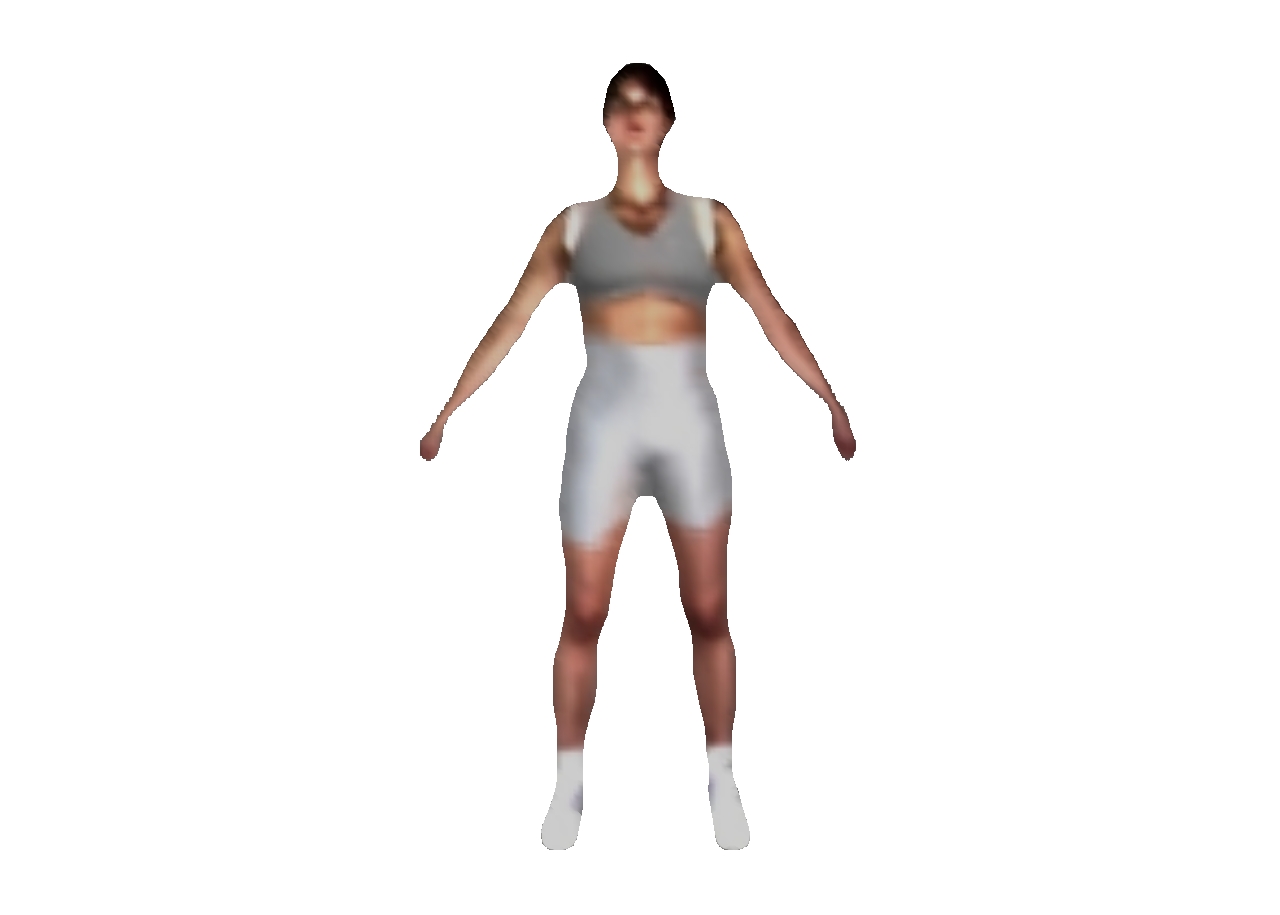}%
%
%
\end{tabular}
\end{center}
\caption{{\bf Human image-maps.} Three human figure results. Using
a single database object, our method effectively morphs the database image,
automatically fitting it to the input depth's 3D features. For
each result, displayed from left to right, are the input depth,
depth textured with automatically selected database image-map (in red,
depth areas not covered by the database map,) and our result.}
\label{fig:human}
\end{figure*}

\begin{figure*}
\begin{center}
\begin{tabular}[b]{c@{}c@{}cc@{}c@{}cc@{}c@{}c}
\hspace{-0.4cm}
\includegraphics[height=2.0cm, clip, trim=15cm 7.5cm 15cm 6cm] {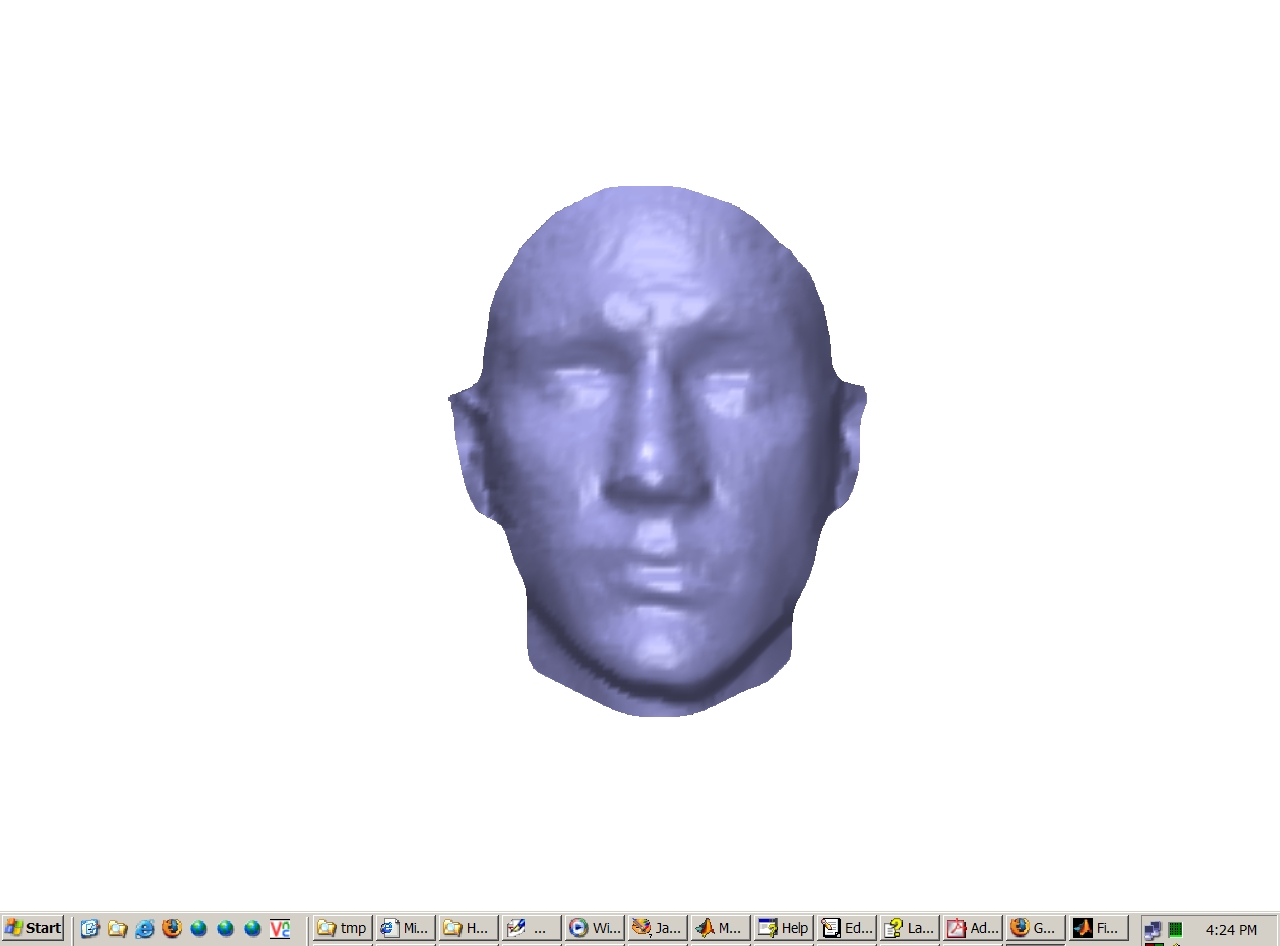}&%
\includegraphics[height=2.0cm, clip, trim=15cm 7.5cm 14.5cm 6cm] {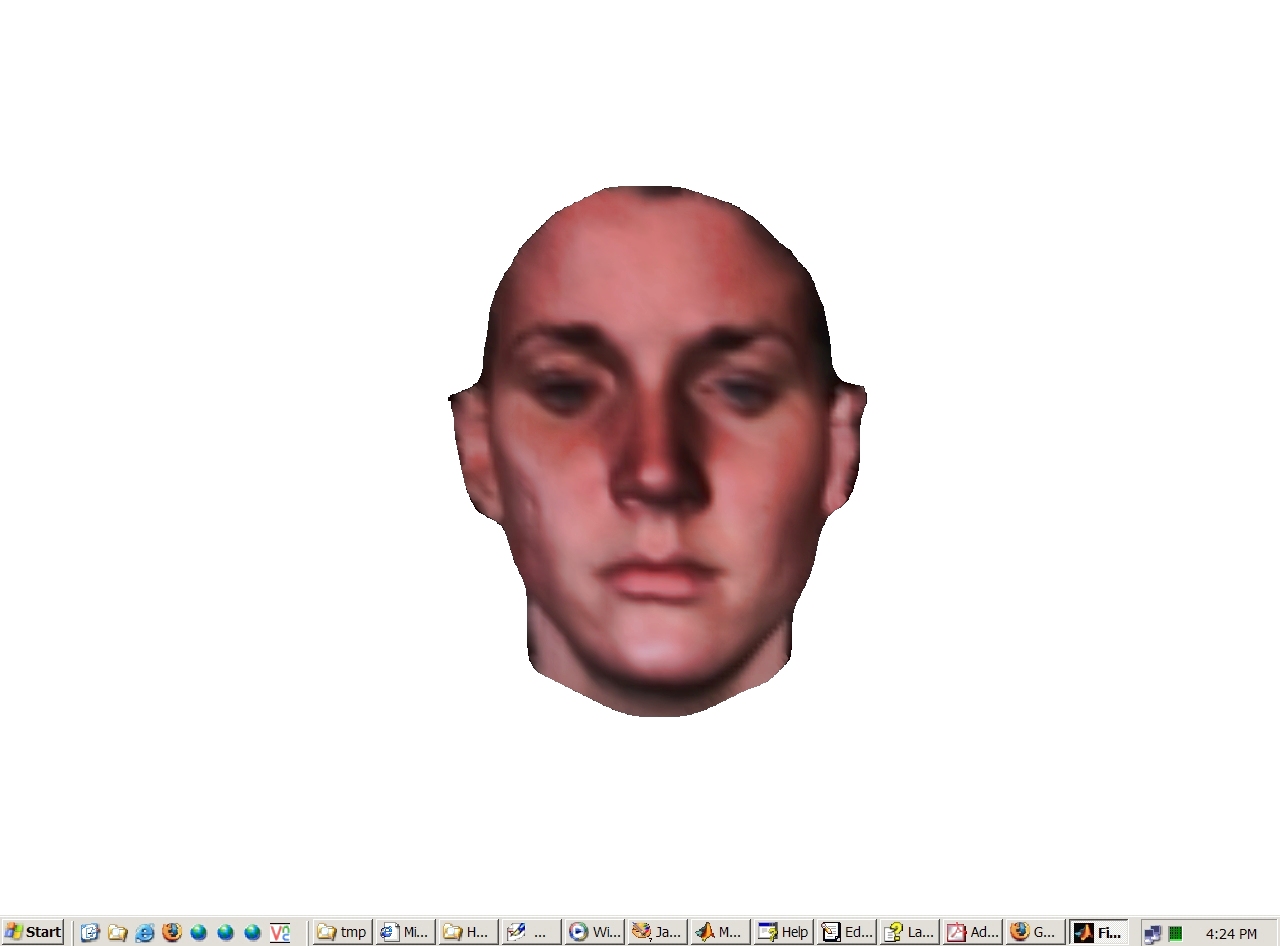}&%
\begin{tabular}[b]{c}
\includegraphics[height=1.07cm, clip, trim=1.5cm 0.2cm 1.5cm 0.2cm] {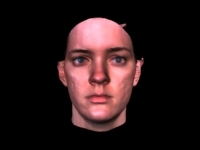}\\%
\includegraphics[height=1.07cm, clip, trim=1.5cm 0.2cm 1.5cm 0.2cm] {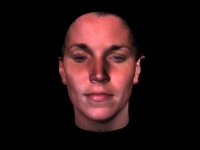}%
\end{tabular}&
\includegraphics[height=2.0cm, clip, trim=15cm 6.8cm 15cm 6cm] {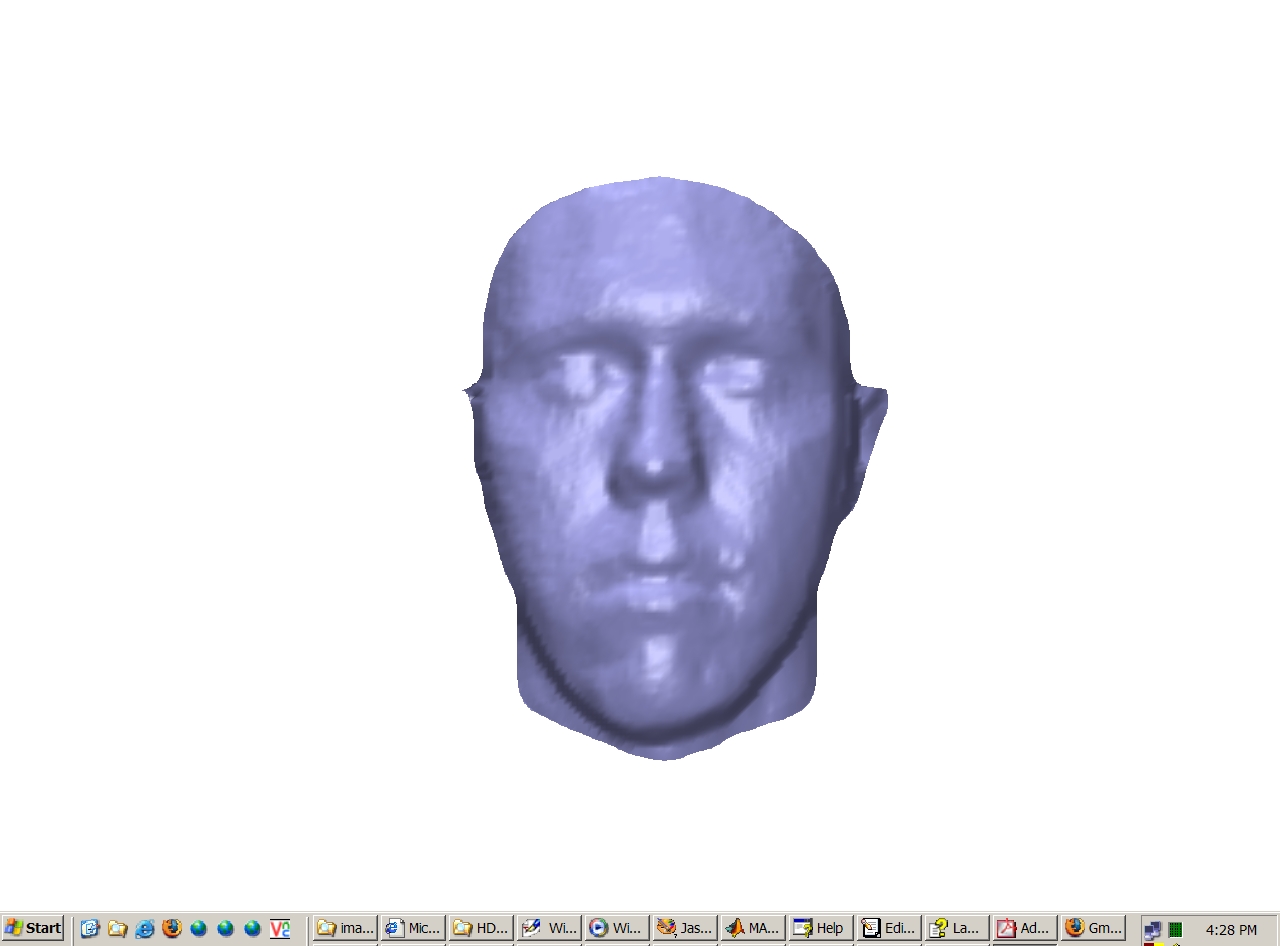}&%
\includegraphics[height=2.0cm, clip, trim=15cm 6.8cm 14.5cm 6cm] {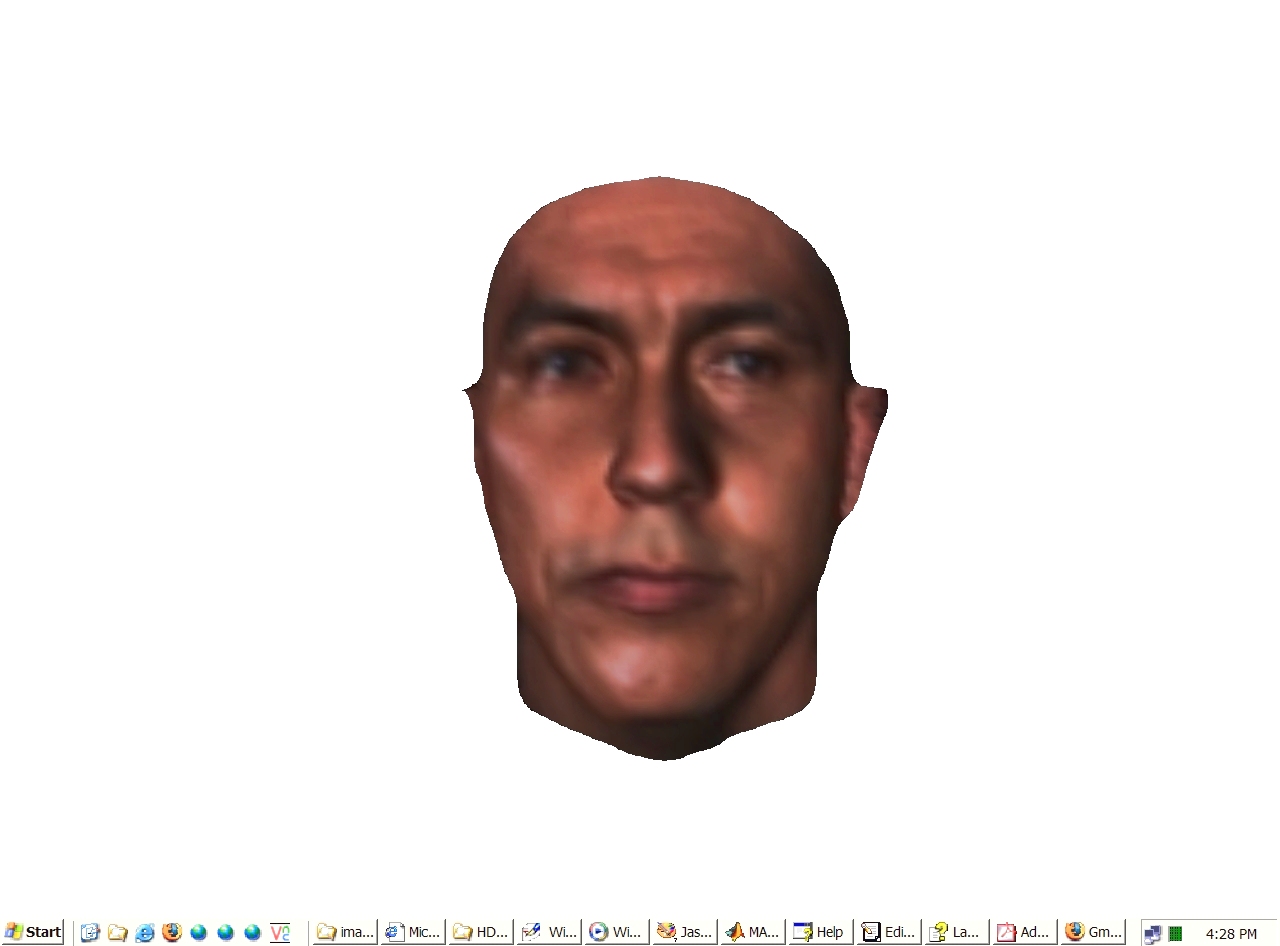}&%
\begin{tabular}[b]{c}
\includegraphics[height=1.07cm, clip, trim=1.5cm 0.2cm 1.5cm 0.2cm] {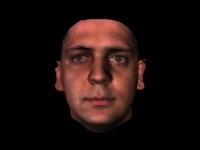}\\%
\includegraphics[height=1.07cm, clip, trim=1.5cm 0.2cm 1.5cm 0.2cm] {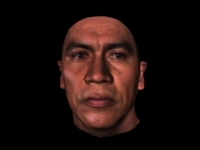}%
\end{tabular}&
\includegraphics[height=2.0cm, clip, trim=15cm 7.5cm 15cm 6cm] {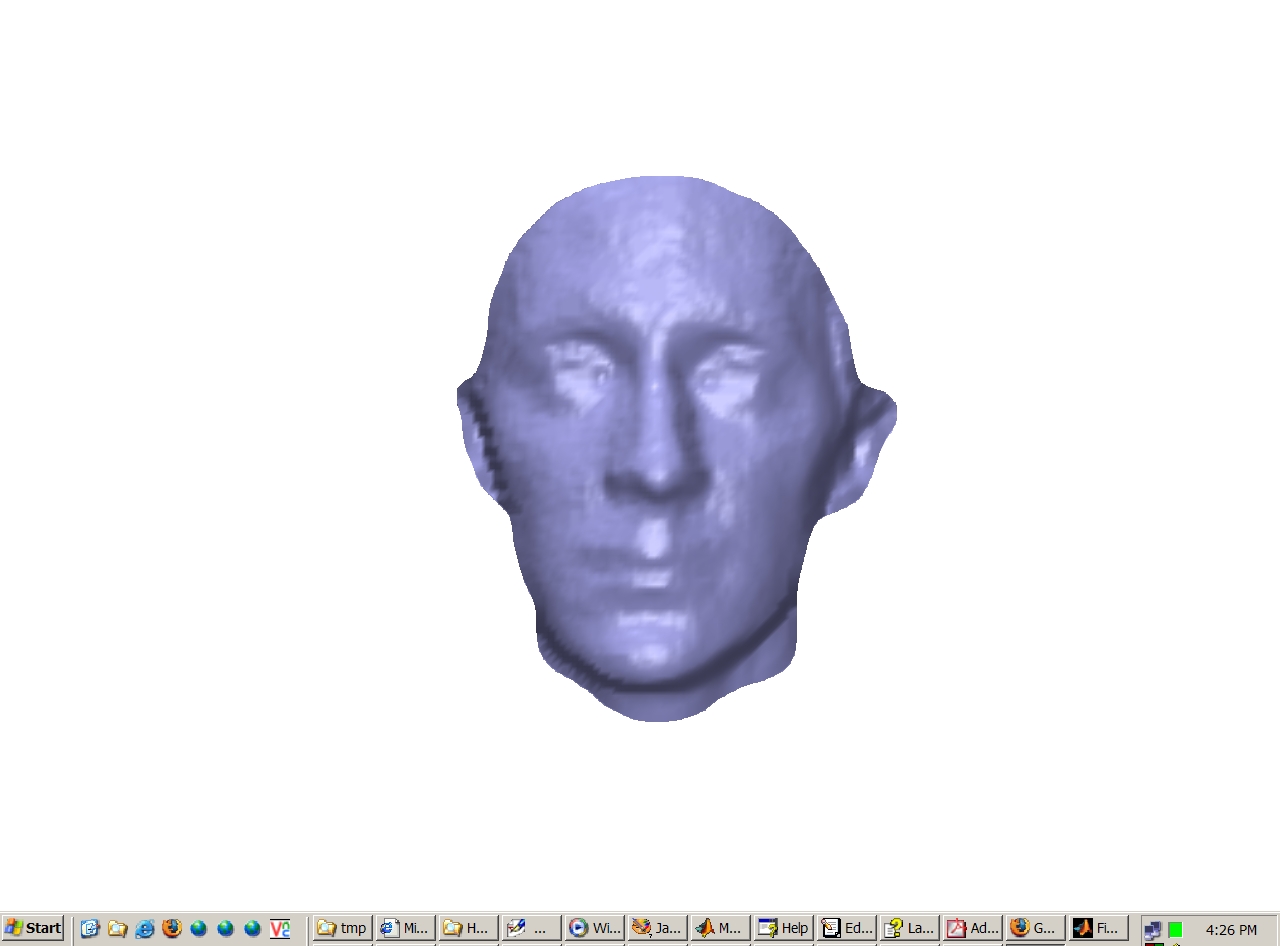}&%
\includegraphics[height=2.0cm, clip, trim=15cm 7.5cm 14.5cm 6cm] {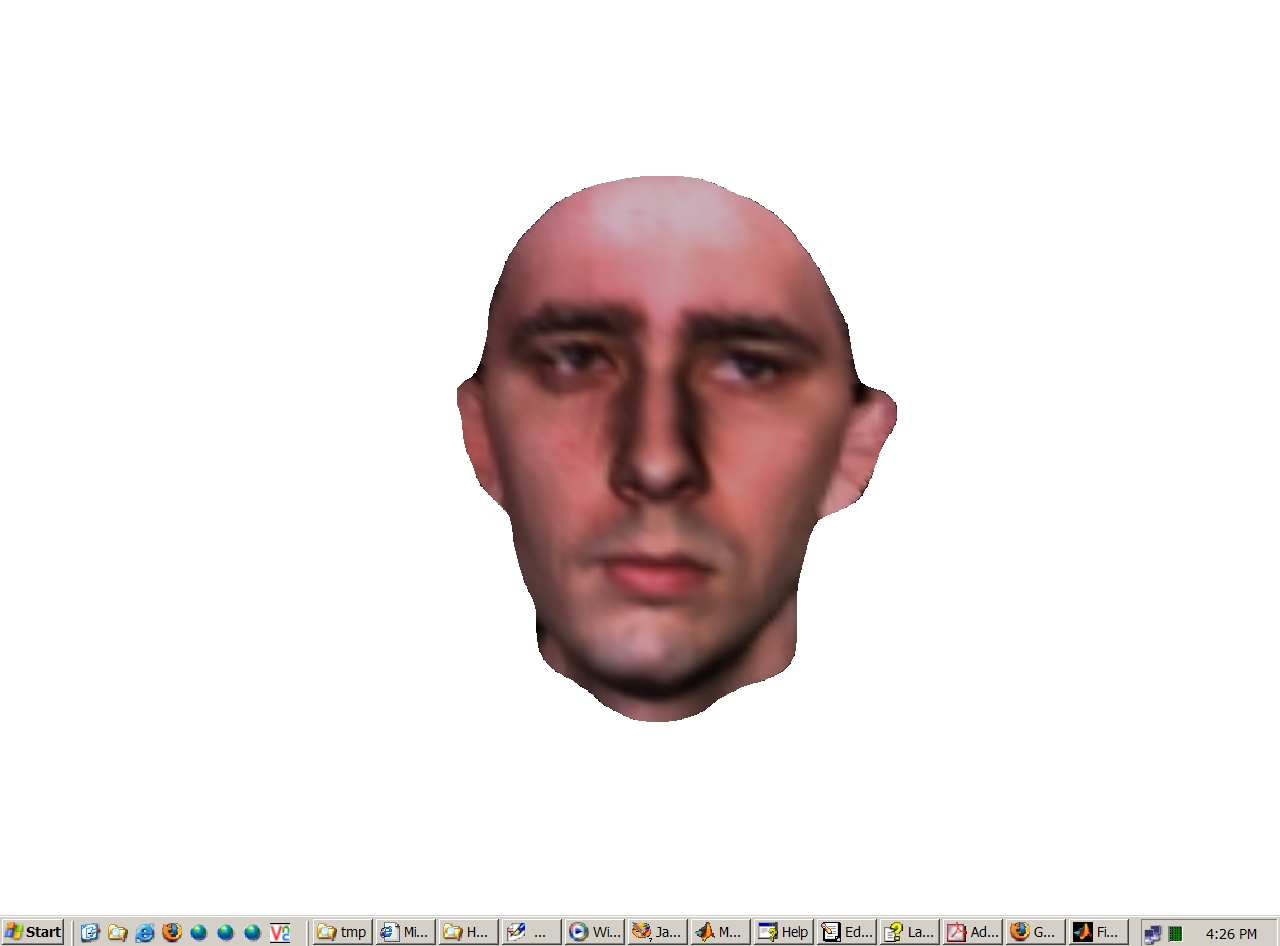}&%
\begin{tabular}[b]{c}
\includegraphics[height=1.07cm, clip, trim=1.5cm 0.2cm 1.5cm 0.2cm] {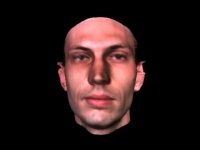}\\%
\includegraphics[height=1.07cm, clip, trim=1.5cm 0.2cm 1.5cm 0.2cm] {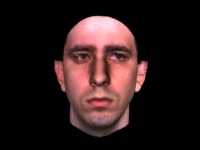}%
\end{tabular}
%
%
\end{tabular}
\end{center}
\caption{{\bf Bust image-maps.} Three bust results. For each
result, displayed from left to right, are the input depth, our
result and the two database objects automatically selected to produce
it.} \label{fig:busts}
\end{figure*}

\begin{figure*}[!ht]
\begin{center}
\begin{tabular}[]{c@{}cc@{}c}
\includegraphics[height=2.6cm, clip, trim=15cm 7.5cm 15cm 6cm] {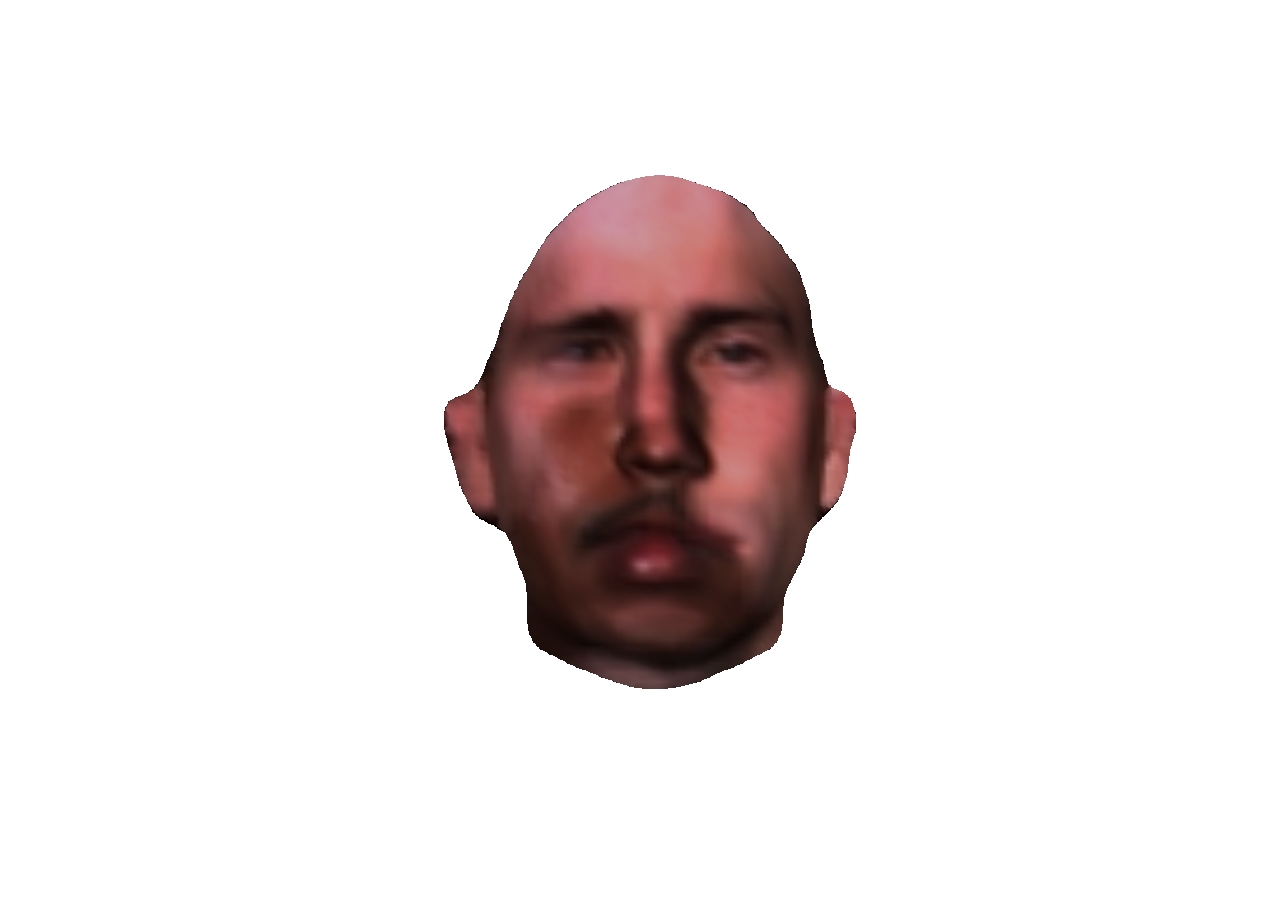}&%
\includegraphics[height=2.6cm, clip, trim=15cm 7.5cm 15cm 6cm] {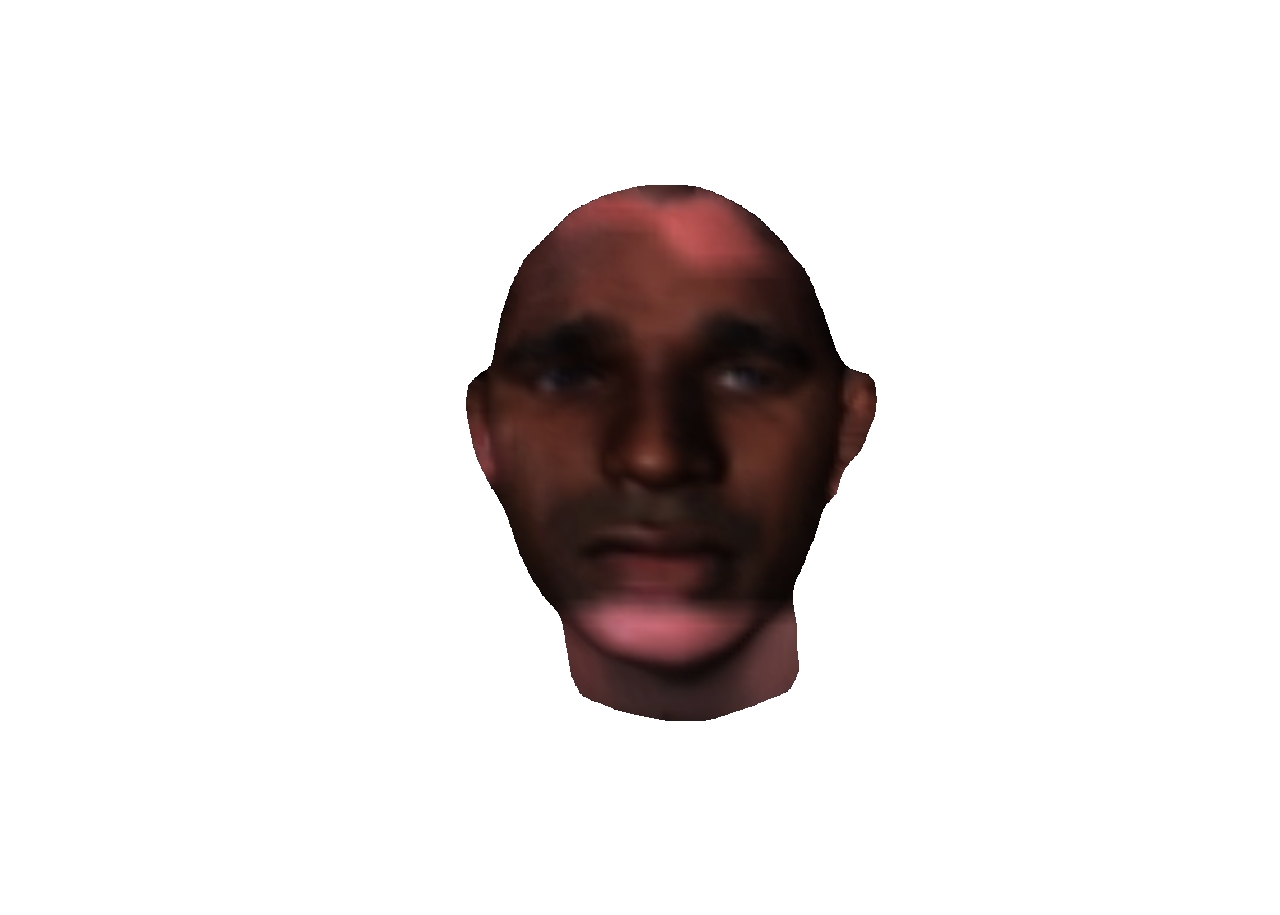}&%
\includegraphics[height=2.6cm, clip, trim=12cm 7cm 12cm 6cm] {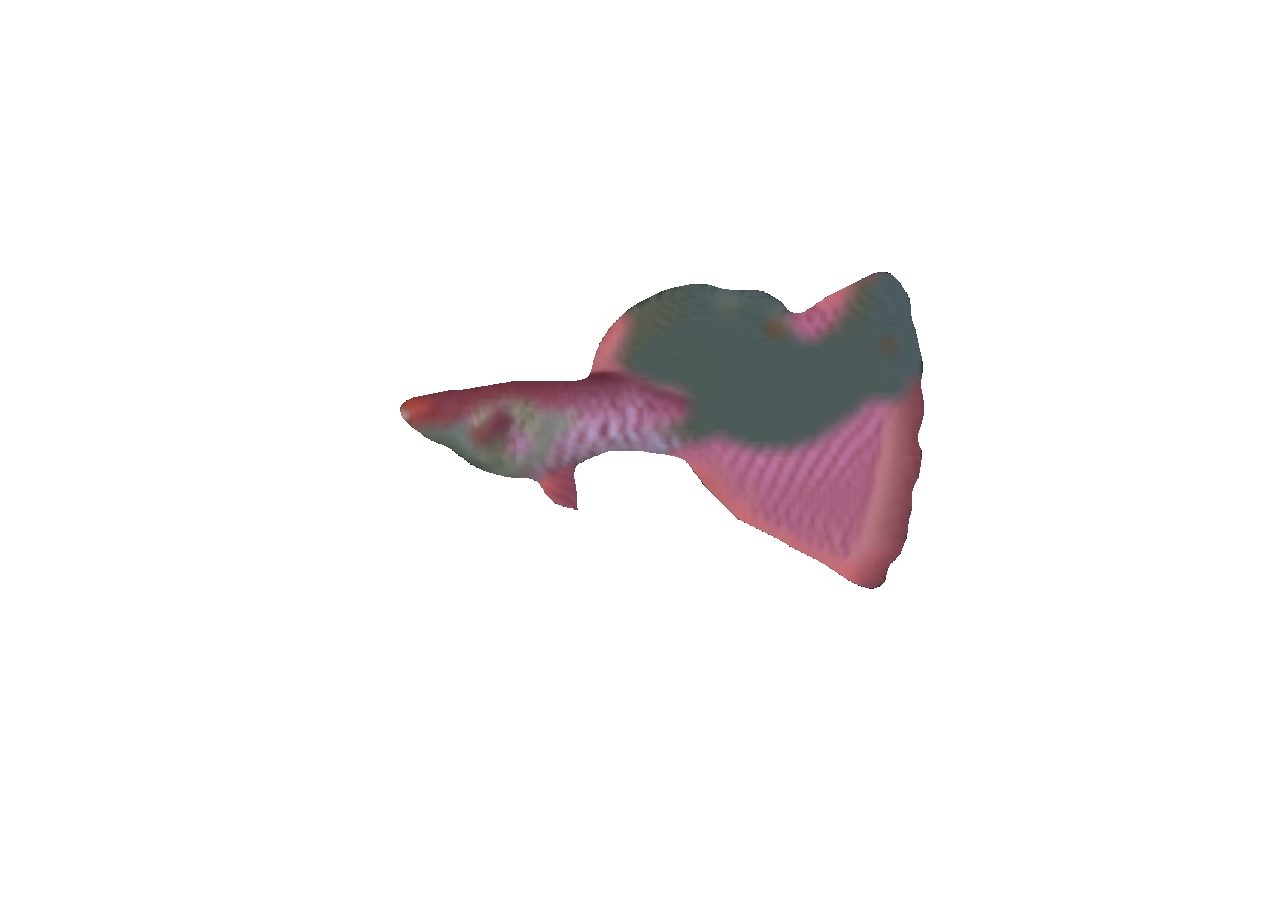}&%
%
\includegraphics[height=2.6cm, clip, trim=13.5cm 7cm 9cm 6cm] {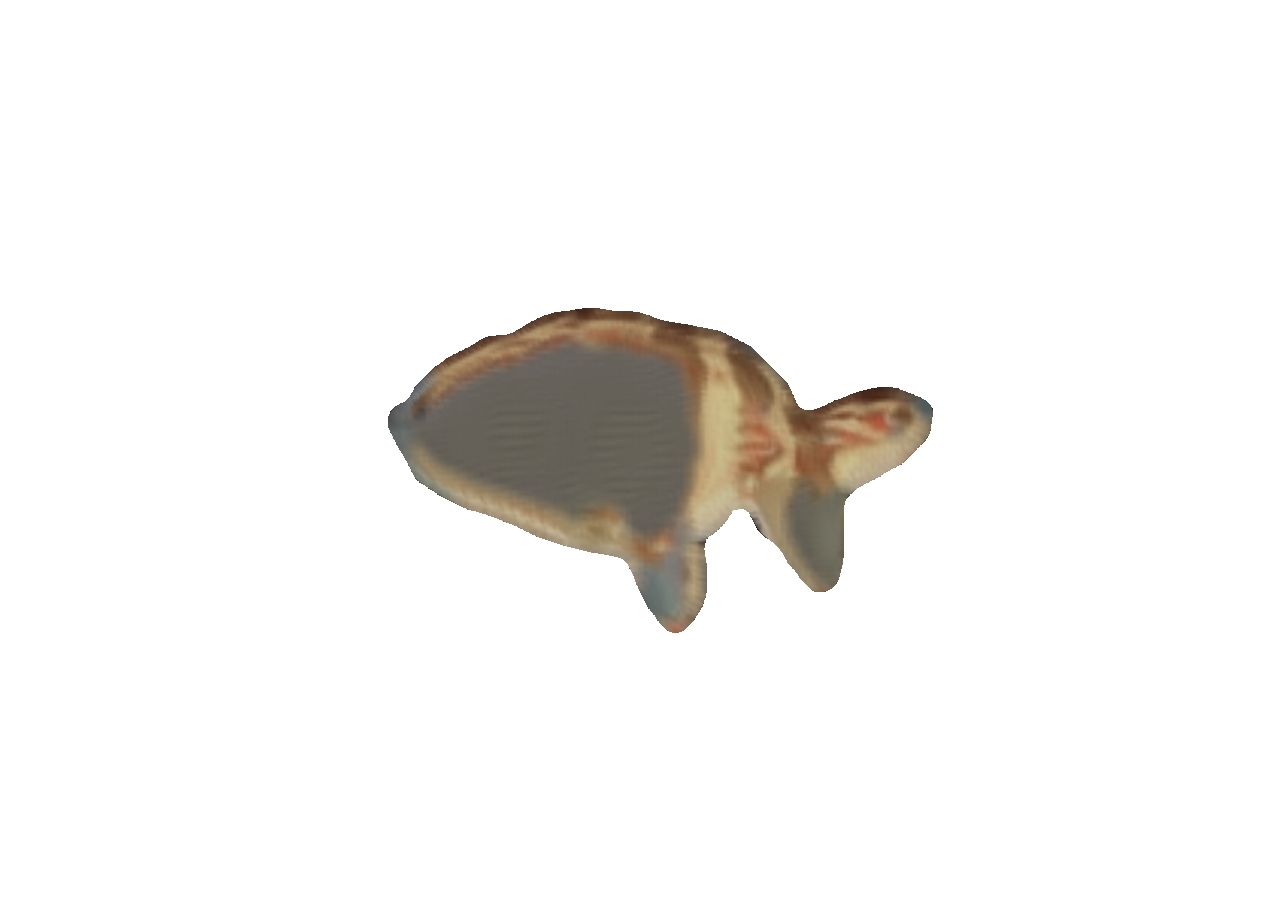}%
%
\end{tabular}
\end{center}
\caption{{\bf Failures.} Bust failures caused by differently
colored database image-maps. Fish failures are due to anomalous input
depths.} \label{fig:failures}
\end{figure*}

\noindent{\bf Automatic Colorization.} Some colorization results are presented in Fig.~\ref{fig:fish_detailed}--\ref{fig:busts}. Note in particular how pairs of database image-maps seamlessly mesh to produce the output image-map in~\ref{fig:busts}, where the database objects are presented alongside the output result. Some failed results are reported in Fig.~\ref{fig:failures}. We believe that the failed faces were due to the automatic example selection disregarding the different colors of the selected database examples. In the case of the fish, the failures were due to the anomalous shapes of the input depths.

To obtain quantitative results for our colorization scheme we again ran leave-one-out tests on the face and fish data sets. Here, since the quality of our colorization results is subjective, we poled 10 subjects, asking ``how many image-maps are faulty or otherwise appear inferior to those in the database''. Out of the 57 fish results on average 28$\%$ were found to be faulty. Similarly, 28$\%$ of our face results were found faulty out of the 76 faces in the database. The parameters used in these tests are reported in Table~\ref{tbl:example}.\\

\begin{table}[!t]\centering
\caption{{\bf Colorization database parameters.} m - Number of mappings (objects) $M_i$ used for synthesis. k - Patch width and height, from fine to coarse scale of three pyramid levels. Weights for depth, depth high-frequencies, Y, Cb, Cr, and relative position components. Note that relative position is amplified for the structured face and humans data-sets. Also, as our eyes are sensitive to 
intensities, we amplify Y as well.}\label{tbl:example}
\begin{tabular}{|l|c|c|c|} \hline
\emph{DB Name} & \emph{m} & \emph{k} & \emph{Weights}\\
\hline \hline
Humans& 1 & 7, 9, 9 & 0.08, 0.06, 8, 1.1, 1.1, 10\\
Busts& 2 & 7, 11, 9 & 0.08, 0.06, 8, 1.1, 1.1, 10\\
Fish& 2 & 7, 11, 9 & 0.08, 0.06, 8, 1.1, 1.1, 0.1\\
\hline
\end{tabular}
\end{table}

\noindent{\bf Run-time.} Our running time was approximately $7.5$ minutes for a $200\times 150$ pixel image using 12 example images at any one time, on a Pentium 4, 2.8GHz computer with 2GB of RAM. We used three pyramid levels, each scaled to half the size of the previous level. Patch sizes, unless otherwise noted, were taken to be $5\times 5$ at the coarsest scale, $7\times 7$ at the second level, and $9\times 9$ for the finest scale.

\section{Conclusions and future work}\label{sec:future}
Clearly, having prior knowledge about the shapes of objects in the world is beneficial for determining the shapes of novel objects. This idea is particularly useful when only a single image of the world is available. Motivated by this basic understanding we formulate an algorithm which produces depth estimates from single images, given examples of typical related objects. The ultimate goal of our algorithm is to produce a depth estimate which is consistent with both the appearance of the input image and the known depths in our example set. We show how this goal may be formally stated and achieved by way of a strong optimization technique.

At the heart of our method is the realization that the problem of depth estimation may be stated as using known mappings from appearances to depths to produce a new, plausible mapping given a novel appearance (image). This observation is coupled with the idea of storing 3D geometries explicitly and using them to render example appearance-depth, mappings on-the-fly. We can thus produce example mappings capturing an essentially infinite range of viewing conditions without limiting the example set a-priori.

As a consequence, we obtain an algorithm which is versatile in the objects and viewing conditions it can be applied to. Moreover, we show that the algorithm is versatile in the problems it may be used to solve: The general formulation of our mappings allows us to estimate additional properties of the objects in the scene, in particular, the shape of the occluded back of the object.\\

\noindent{\bf Future work.} It seems natural to explore how additional information may be estimated using the same framework. For example, can foreground-background segmentation be estimated alongside the depth estimation? There are additional directions which we feel require further study. Chiefly, we would like to explore how the method may be improved, both in accuracy and speed. Here we would like to capitalize on recent advances in image representation and matching, mainly in dense and invariant image representations such as~\cite{basri2011approximate,hassner2012sifts}. 

We believe it would also be interesting to explore how explicit 3D representations may be further exploited. In particular, can an accurate camera viewing position (or illumination, or posture etc.) be estimated by a similar means of producing novel examples on-the-fly and comparing them to the input image? Also, can we learn more about the world occluded from view, by using both known and example information?

\bibliographystyle{elsarticle-num}

\end{document}